\documentclass{article}

\PassOptionsToPackage{numbers, compress}{natbib}

\usepackage[preprint]{neurips_2019}

\usepackage[utf8]{inputenc} 
\usepackage[T1]{fontenc}    
\usepackage{hyperref}
\usepackage{url}            
\usepackage{booktabs}       
\usepackage{amsfonts}       
\usepackage{nicefrac}       
\usepackage{microtype}      
\usepackage{graphicx}
\usepackage{amsmath}
\usepackage{xcolor}
\usepackage{mathtools}
\usepackage{caption}
\usepackage{subcaption}
\usepackage{wrapfig}
\usepackage{verbatim}
\usepackage[verbose]{placeins}
\usepackage{authblk}
\usepackage{textcomp}
\usepackage{makecell}
\usepackage{longtable}
\usepackage[titletoc]{appendix}

\newcommand{\relu}{\text{relu}}

\newtheorem{theorem}{Theorem}

\DeclareMathOperator*{\argmax}{arg\,max}

\DeclareMathOperator*{\R}{\mathbb{R}}
\newcommand{\norm}[1]{\left\lVert#1\right\rVert}
\newcommand{\abs}[1]{\left\lvert#1\right\rvert}
\newcommand{\sgn}{\mathrm{sgn}}
\renewcommand{\d}{\mathrm{d}}
\newcounter{oldtocdepth}
\newcommand{\hidefromtoc}{%
  \setcounter{oldtocdepth}{\value{tocdepth}}%
  \addtocontents{toc}{\protect\setcounter{tocdepth}{-10}}%
}
\newcommand{\unhidefromtoc}{%
  \addtocontents{toc}{\protect\setcounter{tocdepth}{\value{oldtocdepth}}}%
}

\title{Towards Robust Explanations for \\Deep Neural Networks}

\author[a]{Ann-Kathrin Dombrowski}

\author[a,d]{Christopher J. Anders}

\author[a,b,c,d,e]{\\Klaus-Robert M\"uller}

\author[a,d]{Pan Kessel}

\affil[a]{Machine Learning Group, Technische Universität Berlin, Germany}
\affil[b]{Department of Artificial Intelligence, Korea University, Seoul, Korea}
\affil[c]{Max Planck Institute for Informatics, Saarbrücken, Germany}
\affil[d]{BIFOLD -  Berlin Institute for the Foundations of Learning and Data, Berlin, Germany}
\affil[e]{Google Research, Brain team, Berlin, Germany}
\affil[ ]{\texttt {\{klaus-robert.mueller, pan.kessel\}@tu-berlin.de}}

\begin{document}

\maketitle

\begin{abstract}
Explanation methods shed light on the decision process of black-box classifiers such as deep neural networks. But their usefulness can be compromised because they are susceptible to manipulations. With this work, we aim to enhance the resilience of explanations. We develop a unified theoretical framework for deriving bounds on the maximal manipulability of a model. Based on these theoretical insights, we present three different techniques to boost robustness against manipulation: training with weight decay, smoothing activation functions, and minimizing the Hessian of the network. Our experimental results confirm the effectiveness of these approaches.
\end{abstract}



\hidefromtoc

\section{Introduction}
In recent years, deep neural networks have revolutionized many different areas. Despite their impressive performance, the reasoning behind their decision processes remains difficult to grasp for humans. This can limit their usefulness in applications that require transparency. Explanation methods promise to make neural networks interpretable. In this work we consider explanations of individual predictions that can be given in terms of explanation maps~\cite{Baehrens2010ExplainIndividual, Simonyan2014DeepInsideCNN, Zeiler2014Deconvnet, gbp, Bach2015OnPixelwiseExplanations, Selvaraju2017GradCAM, Ribeiro2016LIME, Zintgraf2017VisualizingDeepNNDecisions, Shrikumar2017LearningImportantFeatures, Lundberg2017SHAP, Dabkowski2017RealTimeImageSaliency, Smilkov2017SmoothGrad, Sundararajan2017AxiomaticAttribution_IntGrad, fong2017interpretable, Montavon2017DeepTaylorDecomposition, Kindermans2018PatternAttrNet, kim2017interpretability,montavon2018methods,samek2019ExplainableAI, samek2020toward} which visualize the importance of each input feature for the network's prediction.
They give valuable information about relevant features~\cite{sturm2016InterpretableEEG, arbabzadah2016IdentifyingFacialExpressions, schutt2017QuantumInsights, hagele2020resolving}, help us understand what a model has learned~\cite{Zeiler2014Deconvnet, zahavy2016Graying, arras2017relevant, greydanus2018VisualizingAtari}, and identify unwanted behavior or biases in the data~\cite{Ribeiro2016LIME, lapuschkin2016analyzing,lapuschkin2019unmasking, anders2019SPRAY}. 

While explanation methods show promising results in many areas, concerns regarding their reliability exist. Recent work has shown that explanations are sensitive to small perturbations of the input that do not change the classification result \cite{fragile}. Furthermore, these perturbations can be constructed such that an arbitrary target explanation is closely reproduced and all class scores are approximately unchanged (as opposed to only the classification result), see \cite{blamingGeometry}. An alternative approach leaves the input unchanged but manipulates the model such that it has the same output on the entire data manifold but reproduces an arbitrary target explanation map~\cite{heo2019fooling, fairwashing}. The former class of methods is often refered to as input manipulations and the latter as model manipulations.

Untrustworthy explanations are evidently problematic for various reasons. For a large number of applications, one is interested in the prediction as well as in the explanation of a phenomenon. Examples include medical and natural science applications. As some explanations are susceptible even to random input perturbations, it seems questionable if much insight can be derived from inspecting such explanations. In a setting where explanations are legally required~\cite{rightToExplanation}, explanation manipulability obviously raises serious concerns as they cannot be considered trustworthy evidence. An example for this is credit risk assessment: The supplier can obfuscate that a decision was made based on racist, sexist or other discriminating features by manipulating the model~\cite{fairwashing}. Similarly, attacks from the user side are possible by manipulating the input as they can create the impression that the decision was based on unaccepted features and thus subvert the result.

In this paper, we develop methods to make explanations provably more robust against attacks that manipulate the {\em input}.
To this end, we provide the following key contributions:

\begin{itemize}
\item We analyze the difference between the original and the manipulated explanation maps theoretically and provide a unified theoretical framework which allows us to derive bounds on the maximal change.
\item Based on this theoretical framework, we derive several techniques to make neural networks more resilient against attacks on the explanation, namely:
\begin{itemize}
\item regularizing: here, training with weight decay,
\item training with smoothed activation functions,
\item training while minimizing the Hessian of the network with respect to the input. 
\end{itemize}
\item We demonstrate the effectiveness of the above methods experimentally for several different explanation methods on the CIFAR-10 data set.
\end{itemize}

\begin{figure}[htp!]
  \centering
  \includegraphics[width=1.0\linewidth]{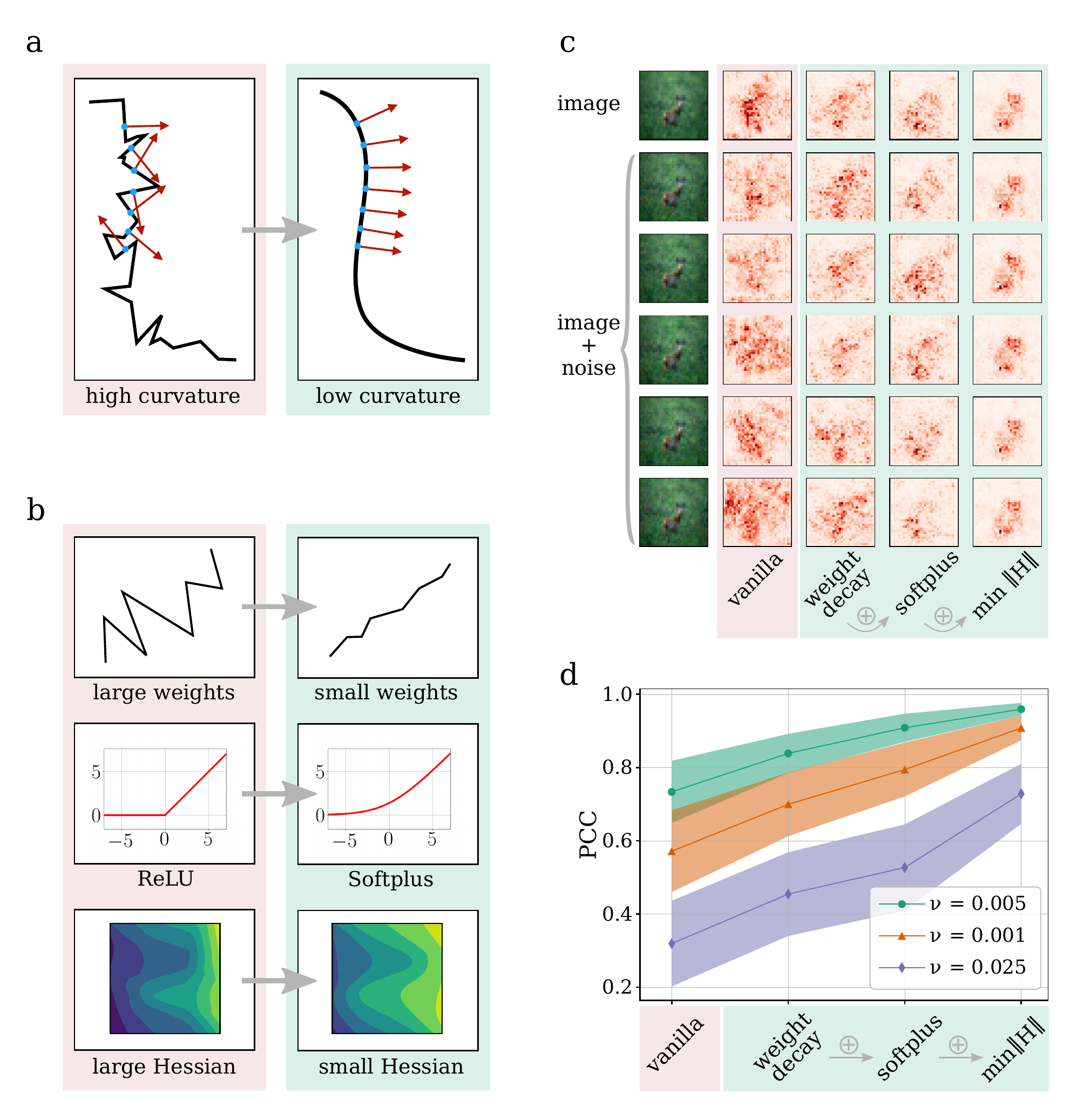}
  \caption{Intuition for our approach and results. \textbf{a} The gradient (red arrows) changes drastically when moving along a line with high curvature but changes only gradually when the curvature is low. We get a similar effect for networks with high and low curvature (see c). \textbf{b} We propose several techniques that reduce curvature when incorporated in the training procedure. Weight decay flattens the angles between piecewise linear functions, Softplus smooths out the kinks of the ReLU function, Hessian minimization reduces curvature locally at the data points. \textbf{c} We show how the Gradient explanation maps change when adding small perturbations to the input. For the vanilla net the explanation maps differ strongly. For networks trained with a combination of our proposed methods the explanation maps become robust to the input perturbations. For a network trained with weight decay, Softplus activations and Hessian minimization (last column) the explanation stays approximately constant. \textbf{d} A quantitative analysis on the complete test set confirms our theoretical findings. The similarity\textemdash measured as Pearson correlation coefficient (PCC)\textemdash between original explanation and explanation of a perturbed input is significantly higher for networks trained with our methods. We get most benefits when combining our methods (last network). We show results for three different noise levels $\nu$.}
  \label{fig:graphical_overview}
\end{figure}

Figure~\ref{fig:graphical_overview} provides an intuition for why explanations are susceptible to manipulation and how our methods lead to more robust explanations.

\subsection{Related Work}
In some works, changing the input is part of the explanation process itself:~\cite{Smilkov2017SmoothGrad} averages gradients over noisy inputs,~\cite{Ribeiro2016LIME} trains an interpretable (linear) classifier on perturbed samples,~\cite{Sundararajan2017AxiomaticAttribution_IntGrad} and~\cite{Montavon2017DeepTaylorDecomposition} use gradual interpolation between an input and a rootpoint to create the explanation and~\cite{Zeiler2014Deconvnet} occludes part of the input with a gray square and then tracks the correct class probability, as a function of the position of the occluded area.
Other studies have used input or model manipulation to compare different explanation methods: In~\cite{Bach2015OnPixelwiseExplanations, evaluating} the pixel flipping method is introduced. Based on the relevance score of the explanation, pixel values are sequentially replaced and the effect on the network output is measured. The idea is to verify that the explanation method accurately identified the relevant pixels\textemdash those that the network needs to correctly classify an image. Variants of this method are also used in~\cite{arras2017ExplainingRNN} and~\cite{sanityChecksMetrics}.~\cite{roar} takes a similar approach but retrains on the modified data to avoid effects from artifacts outside of the original data distribution. Reference~\cite{sanity} proposes ``sanity checks'' for explanations by either perturbing the model weights or the labels of the data set and measuring the resulting change in the explanation. Some explanations turn out to be visually appealing but insensitive to the model or the data generating process. In~\cite{Sixt2020WhenExplLie}, a more theoretical analysis of why many propagation-based explanations fail the sanity checks introduced in~\cite{sanity} is given. Reference~\cite{cat} modifies the input by adding a constant shift, which is then subtracted in the first layer by changing the bias. The modified network and data set produce the exact same outputs as the original but some explanation methods attribute relevance to the shifted pixels.

On the other hand, attention towards more malignant manipulation of explanations has developed relatively recently.
The authors in~\cite{heo2019fooling} explicitly change model weights to manipulate the explanations while keeping the model output approximately unchanged and~\cite{fairwashing} expands on this, by analyzing these experimental results with theoretical insights and providing a method to make explanation more robust against model manipulations. In~\cite{wang2020GradientNLPManipulable}, a facade model is added to the original network in the context of Natural Language Processing. The combined network has similar predictions but the Gradient Explanation is dominated by the customized facade model. Furthermore,~\cite{FoolingLimeAndShap} proposes a similar approach treating the classifier as a blackbox. Their scaffolding technique can change an explanation of a biased classifiers to look completely innocuous.

In~\cite{fragile}, an approach similar to conventional adversarial attacks on the model output is presented. The difference to adversarial attacks on the model output is that attacks on the explanation aim to keep the classification unchanged while the explanation shows major modifications. The work~\cite{blamingGeometry} expands on this and shows that explanations can be changed to arbitrary target maps by adding imperceptible perturbations to the input. The authors explain this susceptibility to manipulations with tools from differential geometry. Our theoretical analysis extends these results significantly in that it also holds for a small (but non-vanishing) change in output of the network, for more general network architectures and various attack methods.

Explanations that include averaging over multiple inputs~\cite{Smilkov2017SmoothGrad} are naturally more resilient to input perturbations, but are not completely save from manipulation~\cite{blamingGeometry}. Several works propose to counteract susceptibility of explanations by changing the explanation process. In~\cite{blamingGeometry} ReLU activations are changed to Softplus activations. This is done for the explanation process only and not part of the training process, in contrast to the approach taken in the present work. Reference~\cite{Wang2020SmoothedGeometry} proposes to include a penalty on the largest principle curvature in the loss function to train networks that are more resilient to attacks on the explanation. This is different to our Hessian norm training which can be roughly understood as a penalty on all principle curvatures. Furthermore,~\cite{fairwashing} proposes a projection of the explanation onto the previously estimated data manifold,~\cite{rieger2020simple} shows that combining several explanation methods can often improve robustness to manipulated inputs,~\cite{Wang2020SmoothedGeometry} averages over several examples from a uniform distribution around the input, and~\cite{lakkaraju2020RobustBlackBoxExpl} proposes adversarial training to construct black box explanations like~\cite{Ribeiro2016LIME, Lundberg2017SHAP} that are robust to input perturbations and distribution shifts.

\section{Theoretical considerations}
In the following, we formally introduce the basic  underlying idea of a theoretical analysis of explanation manipulability. Let us consider gradient explanations for concreteness. We restrict to the output of the winning class, i.e. $g(x):=g(x)_k$ with $k=\argmax_{i} g(x)_i$, since the gradient method only depends on this component of the output. To manipulate the explanation of an input $x \in \mathbb{R}^N$ of a classifier $g:\mathbb{R}^N \to \mathbb{R}$, we construct an adversarially perturbed input $x_{\textrm{adv}}=x+\delta x$ such that the output of the network is (approximately) unchanged, i.e.
\begin{align}
g(x) \approx g(x_{\textrm{adv}})
\end{align}
but the corresponding (gradient) explanations $h = \nabla g$ are drastically different, i.e. 
\begin{align}
    || h(x) - h(x_{\textrm{adv}}) || \gg 1 \,.
\end{align}
Typically, the perturbation is assumed to be small, $|| \delta x || \ll 1$, such that it is imperceptible. For theoretical analysis, one would like to derive upper bounds on the change of saliency map $||h(x)-h(x+\delta x)||$ by any such perturbation $\delta x$. To this end, one considers a curve $\gamma:\mathbb{R} \to \mathbb{R}^N$ with affine parameter $t$ connecting the unperturbed data point $x$ with its adversarially perturbed counterpart $x_{\textrm{adv}}$, i.e.
\begin{align}
    \gamma(t=-\infty) = x \,, && \gamma(t=+\infty) = x_{\textrm{adv}} \,.
\end{align}
In practice, intermediate points on the curve may correspond to iterations of an optimization procedure which adversarially perturbs the input in an iterative manner (although this interpretation is not needed for any of the theoretical considerations). One can then use the gradient theorem to rewrite the change in $j$-th component of the explanation $h$ as\footnote{Here we assume that the classifier $g$ is twice differentiable. However, this assumption can, under certain circumstances, be relaxed as  discussed in Section~\ref{sec:smoothingact}.}
\begin{align}
h_j(x) - h_j(x_{\textrm{adv}}) &= \partial_j g(x) - \partial_j g(x_{\textrm{adv}}) = \int_\gamma \sum_i \partial_i \partial_j g(x) \, \d x_i \nonumber\\
&= \int_{-\infty}^\infty \sum_i\partial_i\partial_j g(\gamma(t)) \, \dot{\gamma}_i(t) \, \d t\,, \label{eq:gradienttrick}
\end{align}
Let the Frobenius norm of the Hessian $H_{ij}(g) = \partial_i \partial_j g$ be bounded, i.e.
\begin{align*}
    || H(g)(x) || \le H^* \in \mathbb{R}_+ \,, && \forall x \in \mathbb{R}^N \,. 
\end{align*}
It then follows immediately that the maximal change in explanation is also bounded: 
\begin{align}
    || h(x) - h(x_{\textrm{adv}}) || &\le \int^{+\infty}_{-\infty} || H(g) \, \gamma(t) || \, \d t \nonumber \\
    & \le H^* \int^{+\infty}_{-\infty} ||\gamma(t)|| \, \d t =  H^* \, L(\gamma) \,, \label{eq:explanationbound}
\end{align}
where $L(\gamma)=\int^{+\infty}_{-\infty} ||\gamma(t)|| \, \d t$ is the length of the curve $\gamma$. We have therefore deduced that bounding the Frobenius norm of the Hessian implies a bound on the maximal possible change in explanation by input manipulation.

\section{Methods for robuster explanations}
Based on the theoretical analysis in the last section, we propose three approaches to reduce the Frobenius norm of the Hessian and thereby increase the robustness with respect to explanation manipulation. 

\subsection{Curvature minimization}\label{sec:CurvatureMinimization}
As a first approach, we propose to modify the training procedure such that a small value of the Frobenius norm of the Hessian is part of the objective. To this end, we add an additional term to the loss function which penalizes the Frobenius norm, i.e.
\begin{align}
    \mathcal{L} = \mathcal{L}_0 + \zeta  \sum_{x \in \mathcal{T}}\norm{H}_F^2(x) \,,
    \label{eq:loss}
\end{align}
where $\zeta$ is a hyperparameter regulating how strongly the Hessian norm is minimized. Futhermore, $\mathcal{T}$ denotes the training set and $\mathcal{L}_0$ is the unregularized loss function. A related approach has been previously proposed in~\cite{RobustnessViaCurvReg} in the context of conventional adversarial attacks. 

Calculating the Frobenius norm of the Hessian is expensive, i.e. to obtain the second derivative we would have to backpropagate through the network once per input pixel. For larger images, this becomes unfeasible especially when we want to include the norm minimization in the training procedure. 

We therefore propose to estimate the Frobenius norm stochastically. Let $v \sim \mathcal{N}(0,1)$, which implies that $\mathbb{E}[v_i]=0$ and $\mathbb{E}[v_i v_j]=\delta_{ij}$.\footnote{Here, we use the Kronecker delta symbol with $\delta_{ij} = \begin{cases} 
      0 & i\neq j \,,\\
      1 & i=j \,. 
   \end{cases}$}
We can then rewrite the Frobenius norm of the Hessian as follows
\begin{align*}
||H||_F^2 &=
\sum_i \left(\frac{\partial^2 g}{\partial x_i\partial x_i}\right)^2 \\
&= \sum_{i,j} \mathbb{E}\left[v_i v_j\right]\left(\frac{\partial^2 g}{\partial x_i\partial x_j}\right)^2 \\
&=\mathbb{E}\left[\sum_i \left(\frac{\partial}{\partial x_i}\sum_j v_j\frac{\partial g}{\partial x_j}\right)^2\right] \,.
\end{align*}
We can estimate the final expectation value by Monte-Carlo, i.e. we draw a random vector $v$, and compute $v^T \nabla g(x)$ at the usual cost of a single backward pass. Since the resulting expression is a scalar, we can calculate its derivative at the cost of another single backward pass \cite{Pearlmutter1994FastExactHessian}. Multiple samples can be combined in mini-batches. The average over the mini-batch is then an unbiased estimator for the expectation value.

\subsection{Weight decay}
The second approach starts from the observation that the Frobenius norm of the Hessian depends on the weights of the neural network. More precisely, in Appendix~\ref{appndx:proofTheorem1} we show that
\begin{theorem}
Let $g:\mathbb{R}^N \to \mathbb{R}$ be a fully-connected neural network with $L$ layers. The weights of the $l$-th layer are denoted by $W^{(l)}$ and its activation functions $\sigma$ are twice-differentiable and bounded
\begin{align}
    |\sigma'(x)| \le \Sigma_1 \,, && |\sigma''(x)| \le \Sigma_2 \,. \label{eq:actbound}
\end{align}
The Hessian of the network is then bounded by
\begin{align}
    ||H(g)||_F \le \sum_{m=1}^L \left( \prod_{l=1}^m ||W^{(l)}||_F^2 \prod_{l=m+1}^L ||W^{(l)}||_F \right) \, \Sigma_1^{L+m-2} \, \Sigma_2 \,. \label{eq:masterbound}
\end{align}
\end{theorem}
As a practical consequence of the theorem, we can reduce the maximal possible change in explanation by decreasing the Frobenius norms of the weights. Motivated by this theoretical insight, we propose to use weight decay for training neural networks such that their explanations are more robust to manipulation. Note while it is well-known that weight decay can improve generalization of neural networks~\cite{WeightDecay_krogh, WeightDecay_hanson, WeightDecay_weigend}, its effect on the manipulability of explanations has not previously been established. 
Other regularizations that reduce the weight norms ($L^1$-regularization, variants of $L^2$-regularization, etc ~\cite{GoodfellowBook}) may have a similar effect.

\subsection{Smoothing activation functions}\label{sec:smoothingact}
As a third approach, we note that the bound of the network's Hessian~\eqref{eq:masterbound} also depends on the maximal values of the activation function's first and second derivatives~\eqref{eq:actbound}. Choosing activations with smaller values for these maximal values therefore will lead to robuster explanations. 

As a concrete example, consider the Softplus activation function
\begin{align}
    \sigma(x) = \frac{1}{\beta} \ln( 1 + e^{\beta x}) \,, \label{eq:Softplus}
\end{align}
where $\beta \in \mathbb{R}_+$ is a hyperparameter. Its first and second derivative are bounded by
\begin{align}
    |\sigma'(x)| \le 1 \,, && |\sigma''(x)| \le \frac14 \beta \,, \label{eq:Softplusbound}
\end{align}
and thus $\Sigma_1 =1$ and $\Sigma_2 = \frac14 \beta$, see \eqref{eq:actbound}. From the bound \eqref{eq:masterbound}, it then follows that networks with Softplus non-linearities with smaller $\beta$ value have robuster explanations compared to networks with larger values of $\beta$ (provided that the Frobenius norms of the weights is the same). 

We therefore propose to use smoother non-linearities, i.e. functions with small $\Sigma_1$ and $\Sigma_2$, to make explanations more robust. 

\paragraph{Note on ReLU non-linearites} The popular ReLU non-linearity can be recovered from Softplus in the limit $\beta \to \infty$. Note however that the bound \eqref{eq:masterbound} diverges in this limit since $\Sigma_2 \to \infty$, see \eqref{eq:Softplusbound}. The fundamental underlying difficulty is that the second derivative $\relu''(x)$ is ill-defined at $x=0$. In Appendix~\ref{app:relu}, we, however, generalize the bound \eqref{eq:masterbound} to the case of ReLU non-linearities. For this, we use the fact that a distributional generalization of the second derivative of the ReLU non-linearity can be defined, i.e. $\relu''(x)=\delta(x)$ where $\delta$ denotes the Dirac distribution. The corresponding right-hand-side of this generalized bound only depends on the weights of the neural network. Thus, this result establishes theoretically that weight decay also certifiably improves robustness for ReLU non-linearities.

\section{Experimental Analysis}\label{sec:ExperimentalAnalysis}

\subsection{Overview}
In this section, we compare the performance of the proposed methods experimentally. 

Briefly summarized, we measure the degree of robustness as follows: we perturb an input sample $x$ by Gaussian noise $\delta x \sim \mathcal{N}(0, \sigma^2)$. For the resulting adversarially perturbed input $x_{\textrm{adv}}=x+\delta x$, we then calculate the explanation $h(x_{\textrm{adv}})$ and measure its similarity to the original explanation $h(x)$. The standard deviation $\sigma$ is chosen such that the output of the neural network is approximately unchanged, i.e. $g(x) \approx g(x_{\textrm{adv}})$. We repeat this analysis for various explanation methods.

In more detail, our experiments use the following setup:

\paragraph{Similarity Scores for Explanations}
In order to quantify the visual similarity of the explanations, we use three different measures following~\cite{sanity}: Pearson correlation coefficient (PCC), structural similarity index measure (SSIM) and mean squared error (MSE). PCC and SSIM are relative error measures where values close to 1 indicate high similarity and small values indicate low similarity. MSE is an absolute error measure where values close to 0 indicate high similarity and large values indicate low similarity.

\paragraph{Model and Dataset}
To demonstrate the proposed robustness effects generically, we use the same convolutional neural network (CNN) architecture for all our models and train on the CIFAR10 dataset~\cite{cifar}. The models achieve up to $88\%$ test set accuracy. For more details on the network architecture and training, we refer to Appendix~\ref{appndx:network}.

\paragraph{Noise Level}
We choose the level of noise such that it does not significantly change the network's output. To this end, we perturb all 10k images of the test set of CIFAR10 with Gaussian noise of a given standard deviation $\sigma$. It is convenient to express the standard deviation $\sigma$ in terms of the noise level $\nu$ by
\begin{align}
 \sigma = (x_{max} - x_{min}) \nu   \,,
\end{align}
where $x_{max}$ and $x_{min}$ denote the maximum and the minimum values of the input domain. For other types of noise we refer to Appendix~\ref{app:othernoise}.

\begin{figure}[htp!]
  \centering
  \includegraphics[width=.6\linewidth]{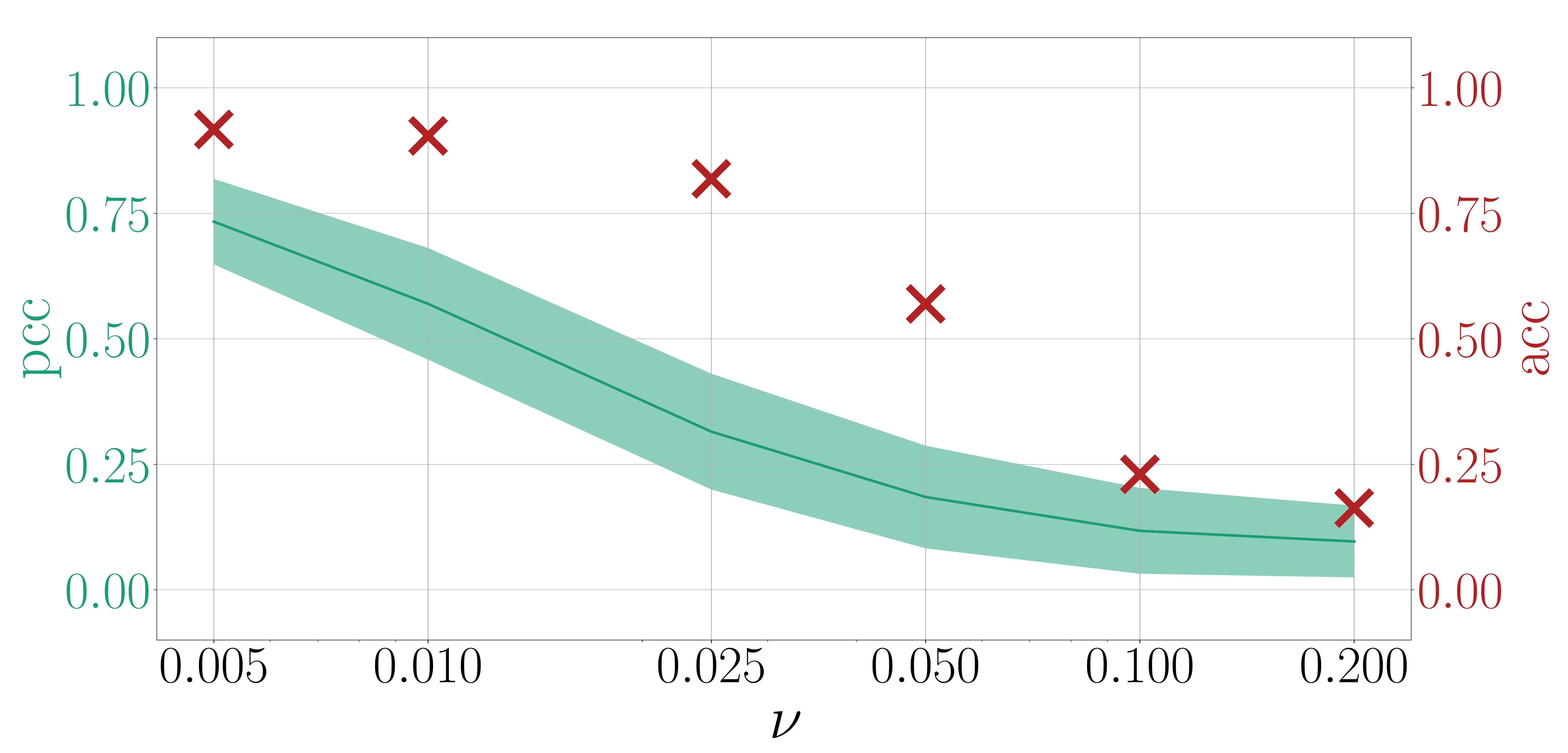}
  \caption{PCC between explanations drops more rapidly than accuracy when adding noise with small $\nu$ to the original image. We show mean +/- std for PCC.}
  \label{fig:pcc_vs_acc}
\end{figure}

Figure~\ref{fig:pcc_vs_acc} shows the classification accuracy and the PCC similarity score between the original and adversarially perturbed explanations for different noise levels $\nu$. Smaller noise levels (between 0.005 and 0.025) lead to a comparatively mild drop in accuracy but result in a significant reduction in the similarity of the explanations. We therefore restrict the noise levels to this interval for our experiments.

\paragraph{Explanation Methods}
We apply our approach to the explanation methods following~\cite{blamingGeometry} :
\begin{itemize}
    \item \textbf{Gradient}: The map $h(x)=\frac{\partial g}{\partial x}(x)$ quantifies change of the scores $g(x)$ due to infinitesimal perturbations in each pixel \cite{Baehrens2010ExplainIndividual, Simonyan2014DeepInsideCNN}.
    \item \textbf{Gradient$\times$Input}: This method uses the map $h(x)=x \odot \frac{\partial g}{\partial x}(x)$  \cite{Shrikumar2017LearningImportantFeatures} which, for linear models, gives the exact contribution of each pixel to the prediction.
    \item \textbf{Integrated Gradients}: This method defines the explanation map $h(x)= (x-\bar{x}) \odot \int_0^1 \frac{\partial g(\bar{x}+t(x-\bar{x}))}{\partial x}\text{d}t$, where $\bar{x}$ is a suitable baseline \cite{Sundararajan2017AxiomaticAttribution_IntGrad}.
    \item \textbf{Guided Backpropagation (GBP)}: This method is a modification of the Gradient explanation which blocks negative components of the gradient when backpropagating through the non-linearities \cite{gbp}.
    \item \textbf{Layer-wise Relevance Propagation (LRP)} is a framework~\cite{Bach2015OnPixelwiseExplanations, Montavon2017DeepTaylorDecomposition} that applies specific rules at different layers to propagate relevance backwards through the network, see~\cite{montavon2019LRPOverview} for a complete overview over the possible choices for the propagation rules. We adopt the following conventions: for the output layer, the relevance is given by
    \begin{align}
        R^L_i = \delta_{ik} \,, \label{eq:lrp_final_layer}
    \end{align}
    where $k$ is the index of the predicted class. This is then propagated backwards through all layers but the first using the $z^+$ rule
    \begin{align}
        R^l_i = \sum_{j} \frac{x_i^l (W^l)^+_{ji}}{\sum_i x_i^l (W^l)^+_{ji}} R^{l+1}_j \,, \label{eq:lrp_interm_layer}
    \end{align}
    where $(W^l)^+$ denotes the positive weights of the $l$-th layer and $x^l$ is the activation vector of the $l$-th layer.
    For the first layer, we use the $z^\mathcal{B}$ rule to account for the bounded input domain 
    \begin{align}
        R^0_i = \sum_{j} \frac{x_j^0 W^{0}_{ji}-l_j (W^{0})^+_{ji}-h_j (W^{0})^{-}_{ji}}{\sum_i ( x_j^0 W^{0}_{ji}-l_j (W^{0})^+_{ji}-h_j (W^{0})^{-}_{ji})} R^{1}_j \,, \label{eq:lrp_first_layer}
    \end{align}
    where $l_i$ and $h_i$ are the lower and upper bounds of the input domain respectively. 
\end{itemize}
For an extensive overview of these methods see~\cite{samek2019ExplainableAI,samek2020toward}. To obtain a pixelwise relevance score, we sum over absolute values of the three colour channels and normalize the explanation to have $\sum_i|h(x)_i| = 1$.

\subsection{Robustness from weight decay}
\begin{figure}[htp!]
\minipage{0.49\textwidth}
\includegraphics[width=.9\linewidth]{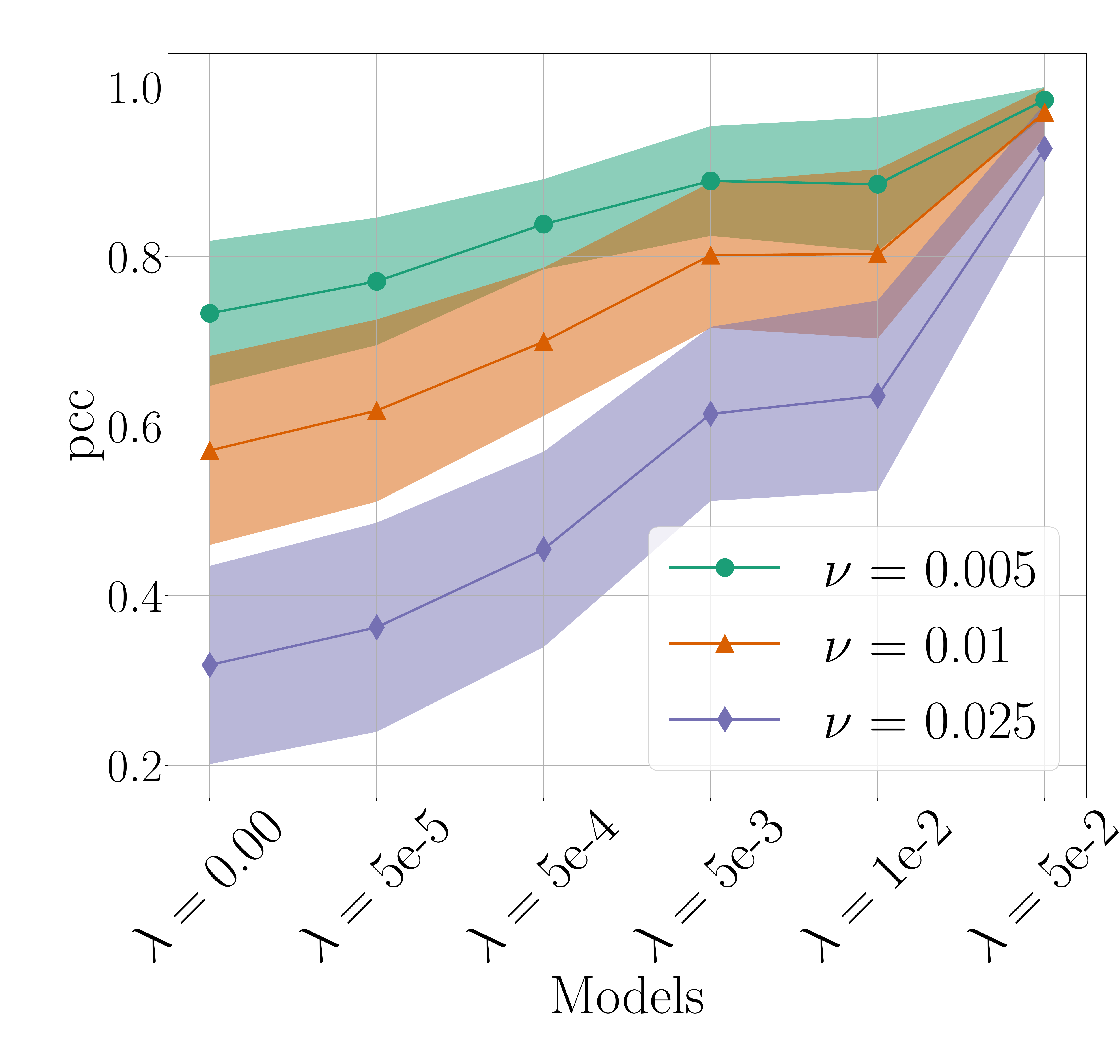}
\endminipage\hfill
\minipage{0.49\textwidth}
\includegraphics[width=.9\linewidth]{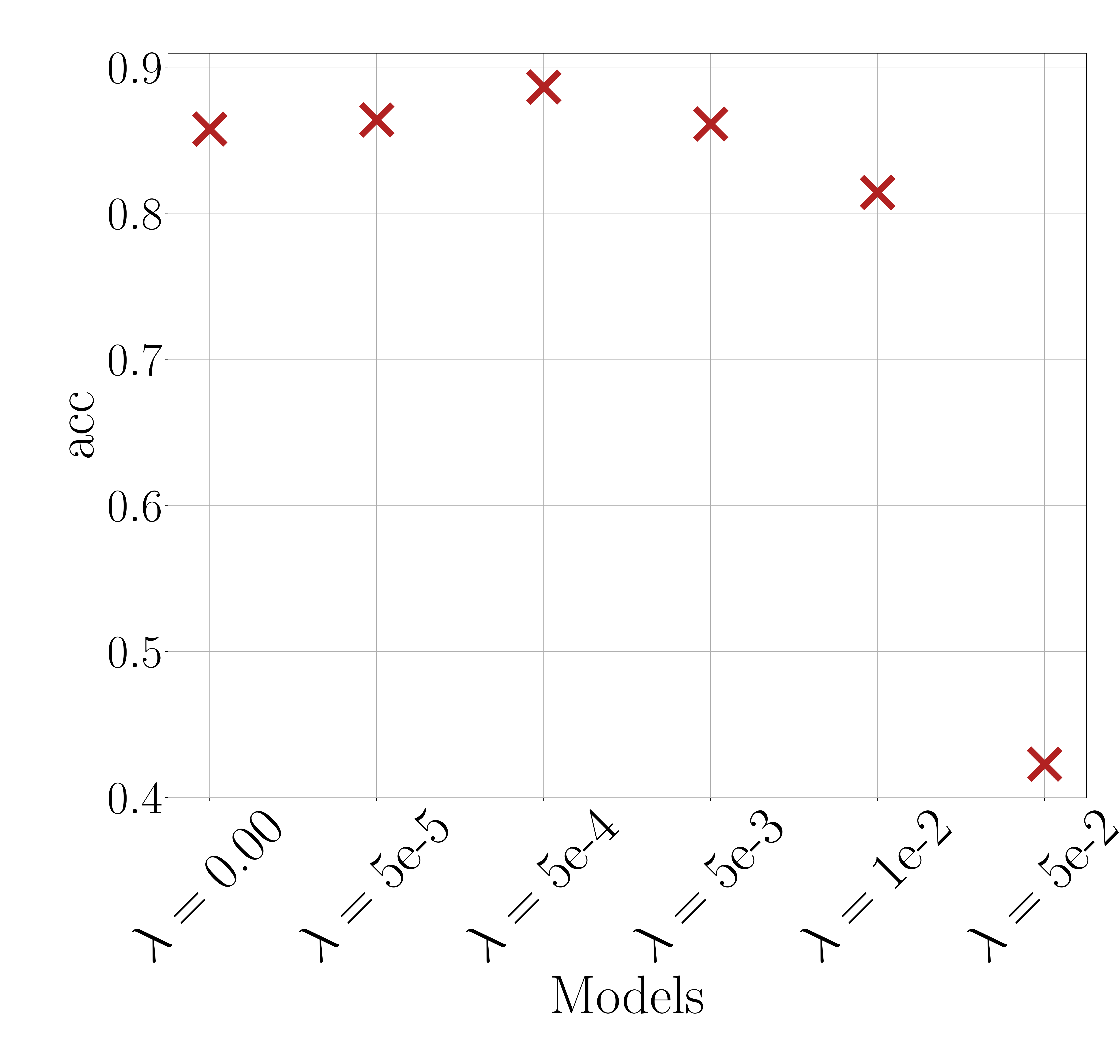}
\endminipage
\caption{Left: PCC increases with stronger weight decay (higher $\lambda$). Therefore, weight decay improves robustness of explanations. We show mean +/- std for three different noise levels $\nu$. Right: For moderate weight decay ($\lambda\approx$~5e-4) accuracy increases, while for strong weight decay ($\lambda\geq$~1e-2) accuracy drops.}
\label{fig:wd_gradient}
\end{figure}

Weight decay adds a regularizing term to the update rule of the network parameters $w_i$ so that large values are penalized.
The update is then given by
\begin{equation}
w_i \rightarrow w_i - \alpha (\frac{\partial\mathcal{L}_0}{\partial w_i} + \lambda w_i)
\end{equation}
where $\alpha$ is the learning rate and $\mathcal{L}_0$ is the unregularized loss. The hyperparameter $\lambda$ controls how strongly the network parameters are penalized. We choose five different values for $\lambda$ and train the CNN for each. 
Figure~\ref{fig:wd_gradient} shows higher PCC values for larger values of $\lambda$, i.e. weight decay increases the robustness of explanations with respect to input manipulation. As was to be expected, there is a trade-off between robustness and accuracy of the networks. For networks trained with strong weight decay ($\lambda>$~1e-2), the accuracy decreases drastically. On the other hand, networks trained with 5e-5~$\leq\lambda\leq$~5e-3 achieve comparable accuracy but are significantly more robust to manipulations than a network trained with $\lambda=0$. 

\subsection{Robustness from Softplus}
\begin{figure}[!htb]
\minipage{0.49\textwidth}
\includegraphics[width=.9\linewidth]{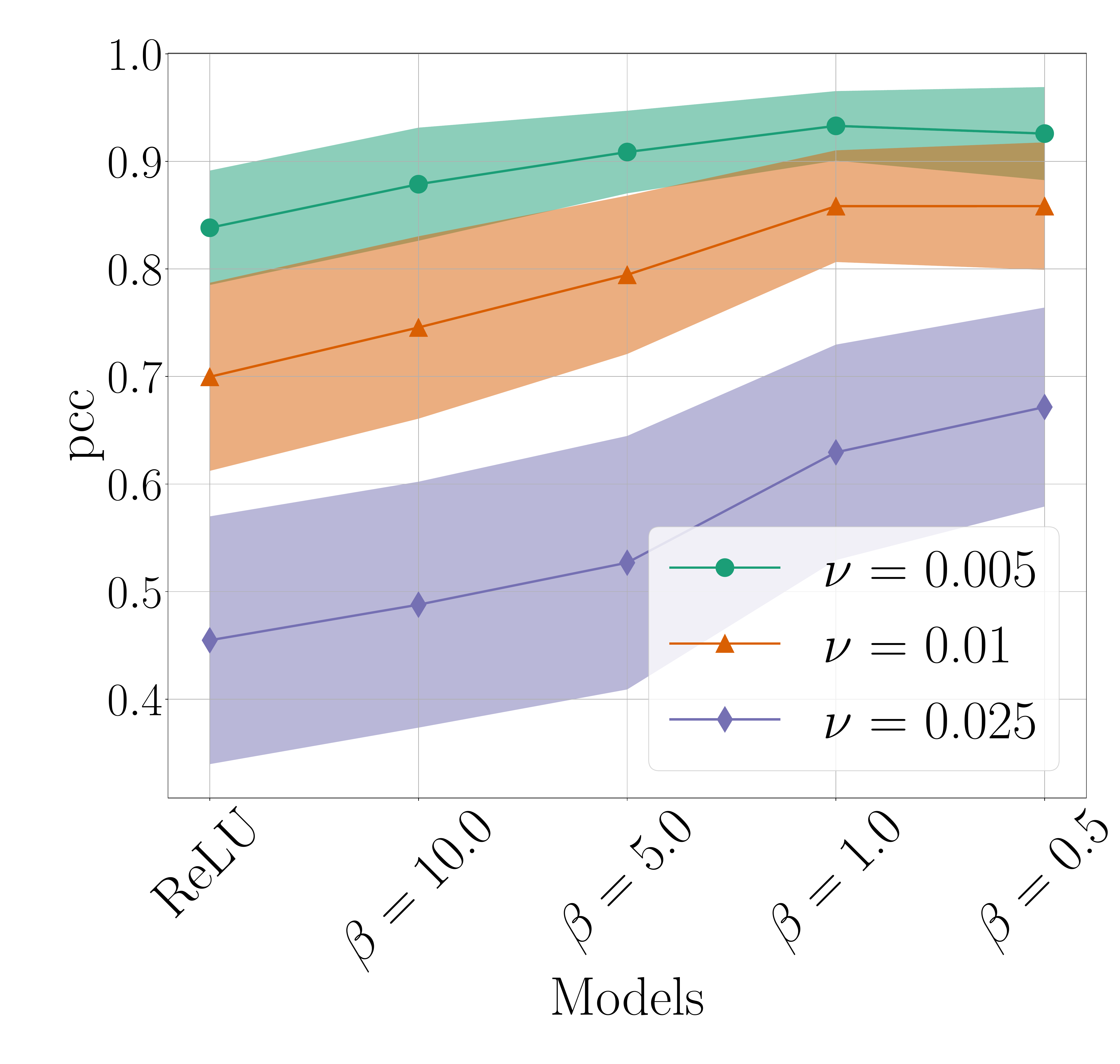}
\endminipage\hfill
\minipage{0.49\textwidth}
\includegraphics[width=.9\linewidth]{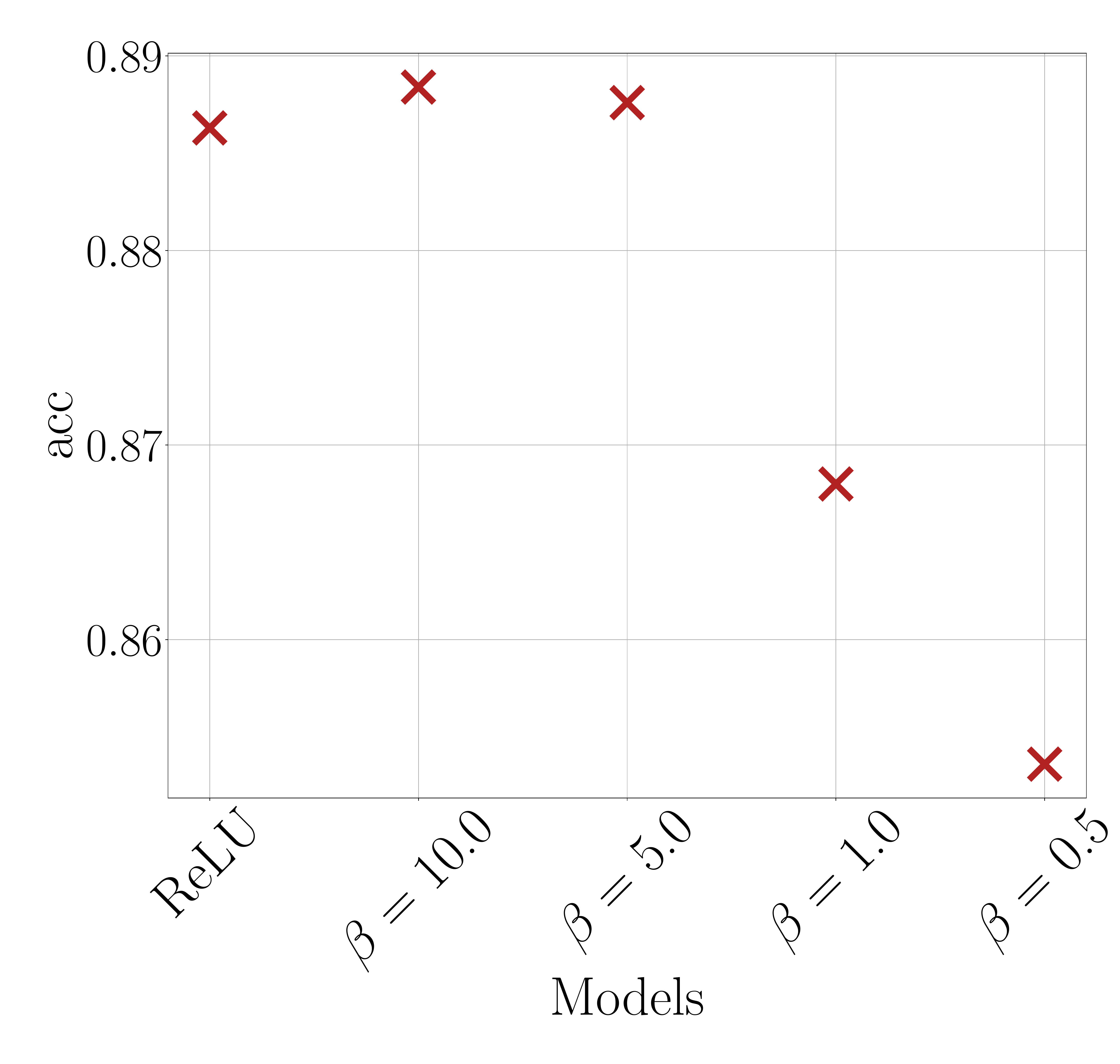}
\endminipage
  \caption{Left: PCC is higher for networks trained with Softplus activation that have small $\beta$ value. Replacing ReLU with Softplus activations improves robustness of explanations. We show mean +/- std for three different noise levels $\nu$. Right: Accuracy decreases if $\beta$ is very small. All networks were trained with weight decay ($\lambda=$~5e-4).}
\label{fig:sp_gradient_wd_5e-4}
\end{figure}

To see how the $\beta$ value of the Softplus activations \eqref{eq:Softplus} affects the robustness, we train networks with four different $\beta$ values. We do this for all but the largest value of the weight-decay hyperparameter $\lambda$ from the previous section; in total $4\cdot5=20$ networks.
With decreasing $\beta$ values, the explanations become less prone to input manipulations. Figure~\ref{fig:sp_gradient_wd_5e-4} shows the results for networks trained with $\lambda=$~5e-4 and different values for $\beta$. For $\beta$ values smaller than 5, the accuracy of the network decreases slightly. Crucially, comparable accuracy is achieved for $\beta$ values of 5 and 10. Results for other choices of the weight decay parameter $\lambda$ look qualitatively similar. We list results for all combinations in Appendix~\ref{appndx:network}. 

\subsection{Robustness from curvature minimization}
To evaluate the effectiveness of Hessian norm minimization, we train networks with different values of the hyperparameter $\zeta$ which controls the degree of regularization in the modified loss in Eq.~\eqref{eq:loss}.  

We approximate the Hessian norm only for Softplus networks since we need to calculate second derivatives and\footnote{More precisely, the second derivative $\relu''(x)$ is not defined for $x=0$ and the relation only holds up to such root points of the non-linearity.} 
\begin{align*}
\frac{\partial^2 g}{\partial x^2} \propto \relu'' = 0 \,
\end{align*}
for ReLU networks. We consider six different values for $\zeta$ for each of the networks from the previous section, i.e. we train $6\cdot 20=120$ networks in total.

\begin{figure}[!htp]
\minipage{0.49\textwidth}
\includegraphics[width=.9\linewidth]{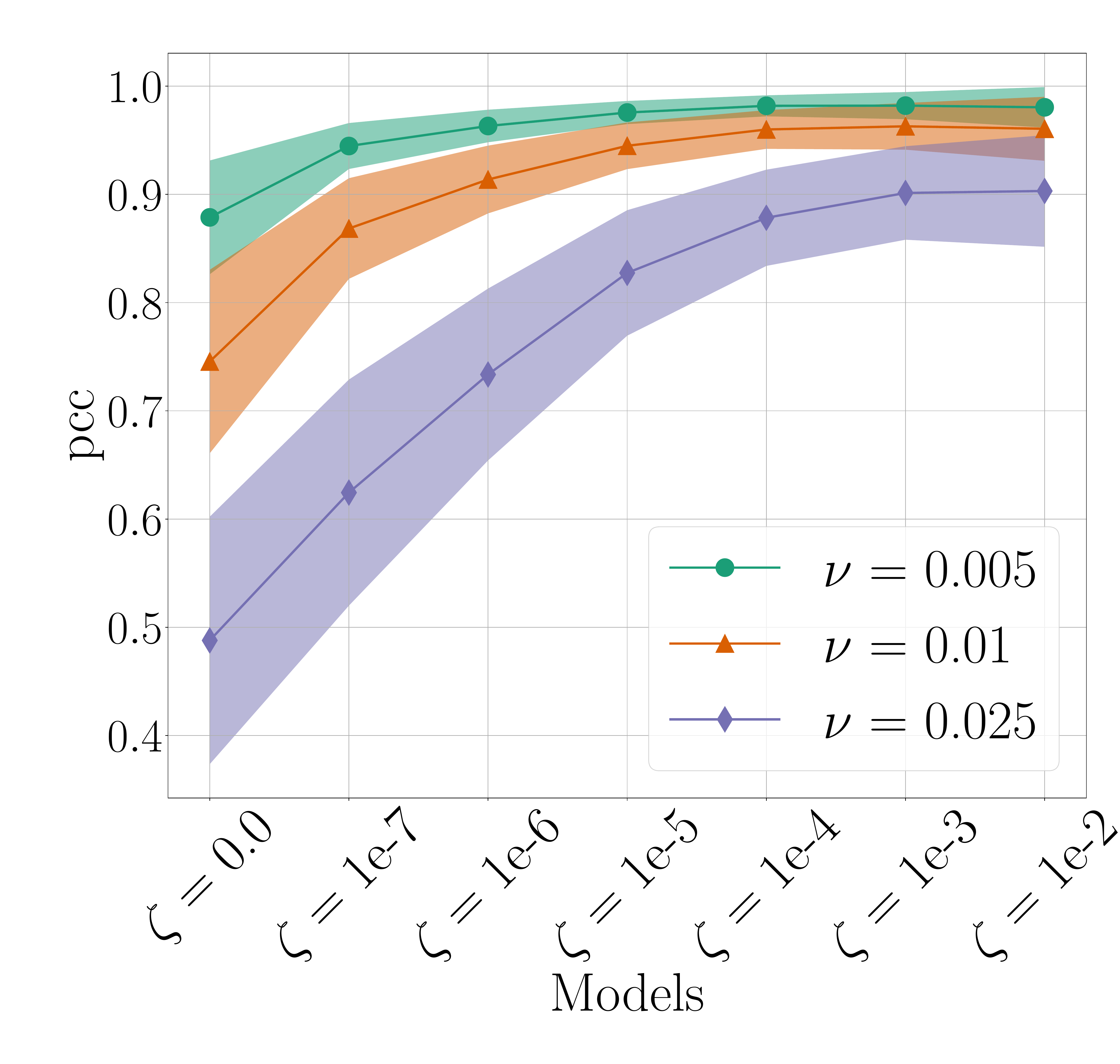}
\endminipage\hfill
\minipage{0.49\textwidth}
\includegraphics[width=.9\linewidth]{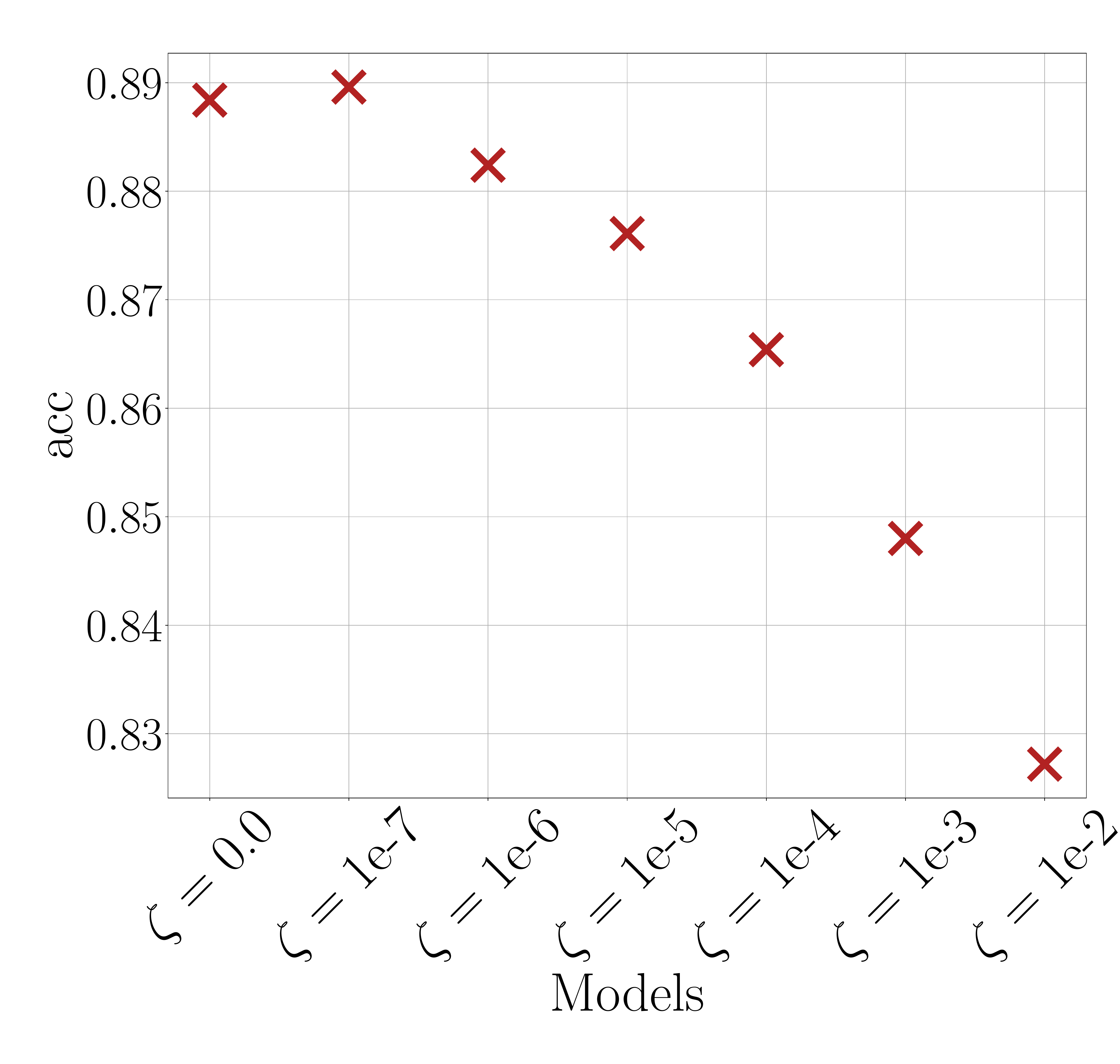}
\endminipage
  \caption{Left: PCC is larger for networks trained with strong minimization of the Hessian norm $\norm{H}$ (larger $\zeta$ values). Therefore, minimizing $\norm{H}$ improves robustness of explanations. We show mean +/- std for three different noise levels $\nu$. Right: accuracy decreases when $\zeta$ gets large. All networks were trained with weight decay ($\lambda=5$e-4) and Softplus activations ($\beta=10$).}
\label{fig:curv_min_gradient_beta_10_wd_5e-4}
\end{figure}

Figure~\ref{fig:curv_min_gradient_beta_10_wd_5e-4} shows how curvature minimization affects the robustness against random perturbations, when using weight decay with $\lambda=$~5e-4 and Softplus activations with $\beta=10$. Even a small value for $\zeta$ results in significant improvement. For larger $\zeta$ values, the PCC value slowly converges to one.  We list results for all combinations of the weight decay parameter $\lambda$ and the Softplus parameter $\beta$ in Appendix~\ref{appndx:network}. 

\begin{figure}[htp!]
\centering
\includegraphics[width=.7\linewidth]{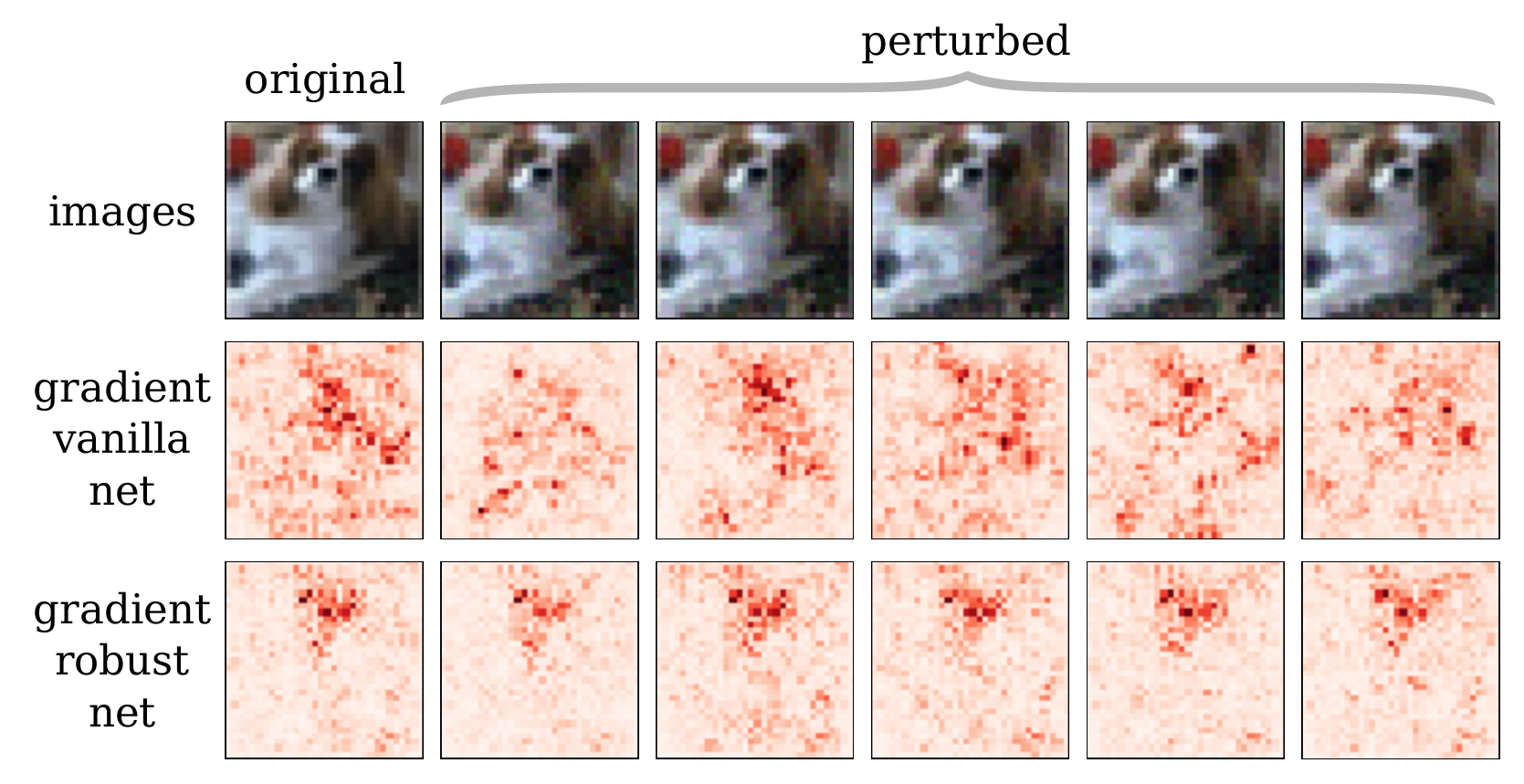}
  \caption{Top row: original image and image with added noise ($\nu=0.025$). Middle row: Gradient explanations for a network trained with $\lambda=0$, ReLU activations and $\zeta=0$. Bottom row: Gradient explanations for a network trained with weight decay ($\lambda=$~5e-4), Softplus activations ($\beta=10$) and Hessian minimization ($\zeta=$~1e-7). The explanations of the robust network in the bottom row are clearly more resilient to random input perturbations.}
\label{fig:noisy_expl_18_gradient_text}
\end{figure}

Figure~\ref{fig:noisy_expl_18_gradient_text} shows a concrete example. In the top row, we show an image and several samples with added Gaussian noise (with noise level $\nu=0.025$). Below we show the Gradient explanation maps of two different networks. For the first network (middle row) the explanations appear noisy and vary strongly. This network was trained without any techniques to enhance robustness (no weight decay, ReLU activations, no Hessian minimization). For the second network (bottom row) the explanations stay relatively steady. This network was trained with measures that enhance robustness (weight decay with $\lambda=$~5e-4, Softplus activations with $\beta=10$, Hessian minimization with $\zeta=$~1e-7).

\subsection{Other explanation methods}
So far we have focused on Gradient explanation maps. But we can apply any other suitable explanation method to our networks.
In Table~\ref{table:statistics_other_explanations}, we show results for different explanation methods. Specifically, PCC values (averaged over the complete test set) between original and manipulated explanations when perturbing images with a noise level of $\nu=0.025$ are listed. In the first row, we show how the respective explanations change when using the original network ($\lambda=0$, ReLU activations, $\zeta=0$) and in the second row we show the values for a network trained with all our robustness measures ($\lambda=5$e-4, $\beta=10$, $\zeta=1$e-6).
While the Gradient explanation is most vulnerable to random perturbations, the results for Gradient$\times$Input, and Integrated Gradients look qualitatively similar to the Gradient explanation. When using all our robustifying measures the PCC similarity between these explanations improves by around 30 to 40 percentage points. Guided Backpropagation (GBP) and Layerwise Relevance Propagation (LRP) are noticeably more resilient to random perturbations. However, our robust network still achieves significantly higher PCC similarities, demonstrating that even more robust explanation methods can profit. We refer to Appendix~\ref{app:otherexplanations} for a more detailed discussion.
\begin{table}[htp!]
    \centering 
    \begin{tabular}{|| c | c | c | c | c | c ||} 
 \hline
Network  & Gradient & \makecell{Grad$\times$Input}  & \makecell{IntGrad} & GBP & \makecell{LRP} \\ [0.5ex] 
 \hline\hline
original & 0.32$\pm$0.12 & 0.44$\pm$0.13 & 0.53$\pm$0.12 & 0.78$\pm$0.10 & 0.91$\pm$0.06\\
robust & 0.73$\pm$0.08 & 0.76$\pm$0.08 & 0.82$\pm$0.06 & 0.94$\pm$0.03 & 0.98$\pm$0.01  \\
 \hline
\end{tabular}
    \caption{PCC (mean $\pm$ std) between original explanations and explanations of perturbed inputs (noise level $\nu=0.025$) for explanation maps: Gradient, Gradient$\times$Input, Integrated Gradients, Guided Backpropagation (GBP), and Layerwise Relevance Propagation (LRP). High PCC values indicate high robustness.}
    \label{table:statistics_other_explanations}
\end{table}

\subsection{Comparison of proposed methods}\label{sec:comparison}
All our proposed methods can improve robustness of explanations against input manipulations. We observe this trend for all considered explanation methods, similarity measures and noise levels.

We note that each method appears to improve robustness in a different manner.
As evident from our theoretically-derived upper bound~Eq.~\eqref{eq:masterbound}, both weight decay and small $\beta$ values for the Softplus activations affect the Hessian norm.
Weight decay leads to smaller Hessian norms by minimizing the weight norms. Replacing ReLU by Softplus with comparatively small $\beta$ parameter also leads to smaller Hessian norms but the weight norms stay approximately constant for different $\beta$ values.
When minimizing the Hessian norm directly during training, the Hessian norms decrease significantly while the weight norms decrease only minimally. This shows that Hessian norm minimization does not just improve robustness by indirectly minimizing the weight norms. 

While we showed that each method separately improves robustness\textemdash we keep the weight-decay hyperparameter $\lambda$ constant when evaluating different smoothing parameters $\beta$ for the Softplus activations and we keep the weight-decay hyperparameter $\lambda$ and the smoothing parameter $\beta$ constant when evaluating different values for hyperparameter $\zeta$ for the Hessian norm minimization\textemdash we get most benefits when combining them.
Besides enhancing robustness, weight decay plays an essential role for the accuracy\textemdash as expected and well-known in the literature \cite{WeightDecay_krogh}; all networks trained without weight decay stay at an accuracy below 86.5\%.

\section{Conclusion}
Explanation methods have gained significant popularity among practitioners in science and engineering recently. With increased attention to explainable AI, questions about manipulability and thus trustworthiness of explanations have been raised. In this contribution, we have addressed the need for robustness of explanation methods against manipulation of the input data. Rather than introducing a new explanation method, we focused on enhancing the robustness of the networks themselves and, as a result, any applied explanation method was shown to profit. 

We could derive bounds for the maximal change in explanation. Based on this theoretical analysis, we proposed three approaches to increase the robustness of explanations. Specifically, we show that weight decay can efficiently boost robustness of explanations. We furthermore propose to use networks with smoothed activation functions and to include a regularizer for the network's curvature in the training process, which leads to significantly enhanced resilience against manipulated inputs.

An interesting direction for future research will be to relate the established limits of robust explanation methods to techniques for uncertainty quantification respectively in relation to methods studying the relevant structural parts in learning models \cite{JMLR:v9:braun08a, JMLR:v12:montavon11a}. Furthermore it will be helpful to discuss resilience to  manipulation of explanation methods also in the context of unsupervised learning \cite{kauffmann2019clustering,kauffmann2020towards,ruff2020unifying} and multi-modal data/similarity streams \cite{eberle2020building}.  

\section*{Declaration of Competing Interest}
The authors declare that they have no known competing financial interests or personal relationships that influence the work reported in this paper.

\section*{Acknowledgments}
This work was supported in part by the German Ministry for Education and Research (BMBF) under Grants 01IS14013A-E, 01GQ1115, 01GQ0850, 031L0207D, 01IS18025A and 01IS18037A.
This work was also partly supported by the Institute of Information \& Communications Technology Planning \& Evaluation (IITP) grants funded by the Korea Government (No. 2017-0-00451, Development of BCI based Brain and Cognitive Computing Technology for Recognizing User’s Intentions using Deep Learning and No. 2019-0-00079,  Artificial Intelligence Graduate School Program, Korea University),
as well as by the Research Training Group ``Differential Equation- and Data-driven Models in Life Sciences and Fluid Dynamics (DAEDALUS)'' (GRK 2433) and Grant Math+, EXC 2046/1, Project ID 390685689 both funded by the German Research Foundation (DFG).
We gratefully acknowledge helpful comments on the ms by Rodolphe Jenatton. 
Correspondence to KRM and PK. 

\FloatBarrier
\bibliography{refs}

\begin{thebibliography}{10}

\bibitem{Baehrens2010ExplainIndividual}
David Baehrens, Timon Schroeter, Stefan Harmeling, Motoaki Kawanabe, Katja
  Hansen, and Klaus-Robert M\"uller.
\newblock {How to Explain Individual Classification Decisions}.
\newblock {\em Journal of Machine Learning Research}, 11(61):1803--1831, 2010.

\bibitem{Simonyan2014DeepInsideCNN}
Karen Simonyan, Andrea Vedaldi, and Andrew Zisserman.
\newblock {Deep Inside Convolutional Networks: Visualising Image Classification
  Models and Saliency Maps}.
\newblock In {\em 2nd International Conference on Learning Representations,
  {ICLR} 2014, Banff, AB, Canada, April 14-16, 2014, Workshop Track
  Proceedings}, 2014.

\bibitem{Zeiler2014Deconvnet}
Matthew~D. Zeiler and Rob Fergus.
\newblock {Visualizing and Understanding Convolutional Networks}.
\newblock In {\em Computer Vision - {ECCV} 2014 - 13th European Conference,
  Zurich, Switzerland, September 6-12, 2014, Proceedings, Part {I}}, pages
  818--833, 2014.

\bibitem{gbp}
Jost~Tobias Springenberg, Alexey Dosovitskiy, Thomas Brox, and Martin~A.
  Riedmiller.
\newblock {Striving for Simplicity: The All Convolutional Net}.
\newblock In {\em 3rd International Conference on Learning Representations,
  {ICLR} 2015, San Diego, CA, USA, May 7-9, 2015, Workshop Track Proceedings},
  2015.

\bibitem{Bach2015OnPixelwiseExplanations}
Sebastian Bach, Alexander Binder, Grégoire Montavon, Frederick Klauschen,
  Klaus-Robert Müller, and Wojciech Samek.
\newblock {On Pixel-Wise Explanations for Non-Linear Classifier Decisions by
  Layer-Wise Relevance Propagation}.
\newblock {\em PLoS ONE}, 10(7):1--46, 07 2015.

\bibitem{Selvaraju2017GradCAM}
Ramprasaath~R. Selvaraju, Abhishek Das, Ramakrishna Vedantam, Michael Cogswell,
  Devi Parikh, and Dhruv Batra.
\newblock {Grad-CAM: Visual Explanations from Deep Networks via Gradient-Based
  Localization}.
\newblock In {\em 2017 IEEE International Conference on Computer Vision
  (ICCV)}, pages 618--626, 2017.

\bibitem{Ribeiro2016LIME}
Marco~Tulio Ribeiro, Sameer Singh, and Carlos Guestrin.
\newblock {``Why Should I Trust You?'': Explaining the Predictions of Any
  Classifier}.
\newblock In {\em Proceedings of the 22nd ACM SIGKDD International Conference
  on Knowledge Discovery and Data Mining}, page 1135–1144, New York, NY, USA,
  2016. Association for Computing Machinery.

\bibitem{Zintgraf2017VisualizingDeepNNDecisions}
Luisa~M. Zintgraf, Taco~S. Cohen, Tameem Adel, and Max Welling.
\newblock {Visualizing Deep Neural Network Decisions: Prediction Difference
  Analysis}.
\newblock In {\em 5th International Conference on Learning Representations
  (ICLR), Toulon, France, April 24-26, 2017, Conference Track Proceedings},
  2017.

\bibitem{Shrikumar2017LearningImportantFeatures}
Avanti Shrikumar, Peyton Greenside, and Anshul Kundaje.
\newblock {Learning Important Features Through Propagating Activation
  Differences}.
\newblock In {\em Proceedings of the 34th International Conference on Machine
  Learning, {ICML} 2017, Sydney, NSW, Australia, 6-11 August 2017}, pages
  3145--3153, 2017.

\bibitem{Lundberg2017SHAP}
Scott~M Lundberg and Su-In Lee.
\newblock {A Unified Approach to Interpreting Model Predictions}.
\newblock In {\em Advances in Neural Information Processing Systems 30}, pages
  4765--4774. Curran Associates, Inc., 2017.

\bibitem{Dabkowski2017RealTimeImageSaliency}
Piotr Dabkowski and Yarin Gal.
\newblock {Real Time Image Saliency for Black Box Classifiers}.
\newblock In {\em Advances in Neural Information Processing Systems 30}, pages
  6967--6976. Curran Associates, Inc., 2017.

\bibitem{Smilkov2017SmoothGrad}
Daniel Smilkov, Nikhil Thorat, Been Kim, Fernanda~B. Viégas, and Martin
  Wattenberg.
\newblock {SmoothGrad: removing noise by adding noise}.
\newblock In {\em {Workshop on Visualization for Deep Learning, International
  Conference on Machine Learning (ICML) 2017, Sydney, Australia, Aug 10}},
  2017.

\bibitem{Sundararajan2017AxiomaticAttribution_IntGrad}
Mukund Sundararajan, Ankur Taly, and Qiqi Yan.
\newblock {Axiomatic Attribution for Deep Networks}.
\newblock In {\em Proceedings of the 34th International Conference on Machine
  Learning, {ICML} 2017, Sydney, NSW, Australia, 6-11 August 2017}, pages
  3319--3328, 2017.

\bibitem{fong2017interpretable}
Ruth~C Fong and Andrea Vedaldi.
\newblock {Interpretable explanations of black boxes by meaningful
  perturbation}.
\newblock In {\em IEEE International Conference on Computer Vision (ICCV)},
  pages 3449--3457. IEEE, 2017.

\bibitem{Montavon2017DeepTaylorDecomposition}
Gr{\'e}goire Montavon, Sebastian Lapuschkin, Alexander Binder, Wojciech Samek,
  and Klaus-Robert M{\"u}ller.
\newblock {Explaining nonlinear classification decisions with deep Taylor
  decomposition}.
\newblock {\em Pattern Recognition}, 65:211--222, 2017.

\bibitem{Kindermans2018PatternAttrNet}
Pieter-Jan Kindermans, Kristof~T Sch{\"u}tt, Maximilian Alber, Klaus-Robert
  M{\"u}ller, Dumitru Erhan, Been Kim, and Sven D{\"a}hne.
\newblock {Learning how to explain neural networks: PatternNet and
  PatternAttribution}.
\newblock In {\em International Conference on Learning Representations (ICLR),
  Vancouver Convention Center, Vancouver, BC, Canada April 30 -- May 3}, 2018.

\bibitem{kim2017interpretability}
Been Kim, Martin Wattenberg, Justin Gilmer, Carrie~J. Cai, James Wexler,
  Fernanda~B. Vi{\'{e}}gas, and Rory Sayres.
\newblock {Interpretability Beyond Feature Attribution: Quantitative Testing
  with Concept Activation Vectors {(TCAV)}}.
\newblock In {\em Proceedings of the 35th International Conference on Machine
  Learning, {ICML} 2018, Stockholmsm{\"{a}}ssan, Stockholm, Sweden, July
  10-15}, pages 2673--2682, 2018.

\bibitem{montavon2018methods}
Gr{\'{e}}goire Montavon, Wojciech Samek, and Klaus-Robert M\"uller.
\newblock {Methods for Interpreting and Understanding Deep Neural Networks}.
\newblock {\em Digital Signal Processing}, 73:1--15, 2018.

\bibitem{samek2019ExplainableAI}
Wojciech Samek, Gr{\'e}goire Montavon, Andrea Vedaldi, Lars~Kai Hansen, and
  Klaus-Robert M{\"u}ller.
\newblock {\em {Explainable AI: Interpreting, Explaining and Visualizing Deep
  Learning}}, volume 11700.
\newblock Springer Nature, 2019.

\bibitem{samek2020toward}
Wojciech Samek, Gr{\'e}goire Montavon, Sebastian Lapuschkin, Christopher~J
  Anders, and Klaus-Robert M{\"u}ller.
\newblock {Toward Interpretable Machine Learning: Transparent Deep Neural
  Networks and Beyond}.
\newblock {\em arXiv preprint arXiv:2003.07631}, 2020.

\bibitem{sturm2016InterpretableEEG}
Irene Sturm, Sebastian Lapuschkin, Wojciech Samek, and Klaus-Robert M{\"u}ller.
\newblock {Interpretable deep neural networks for single-trial EEG
  classification}.
\newblock {\em Journal of neuroscience methods}, 274:141--145, 2016.

\bibitem{arbabzadah2016IdentifyingFacialExpressions}
Farhad Arbabzadah, Gr{\'e}goire Montavon, Klaus-Robert M{\"u}ller, and Wojciech
  Samek.
\newblock {Identifying Individual Facial Expressions by Deconstructing a Neural
  Network}.
\newblock In {\em German Conference on Pattern Recognition}, pages 344--354.
  Springer, 2016.

\bibitem{schutt2017QuantumInsights}
Kristof~T Sch{\"u}tt, Farhad Arbabzadah, Stefan Chmiela, Klaus~R M{\"u}ller,
  and Alexandre Tkatchenko.
\newblock {Quantum-chemical insights from deep tensor neural networks}.
\newblock {\em Nature communications}, 8(1):13890, 2017.

\bibitem{hagele2020resolving}
Miriam H{\"a}gele, Philipp Seegerer, Sebastian Lapuschkin, Michael Bockmayr,
  Wojciech Samek, Frederick Klauschen, Klaus-Robert M{\"u}ller, and Alexander
  Binder.
\newblock Resolving challenges in deep learning-based analyses of
  histopathological images using explanation methods.
\newblock {\em Scientific reports}, 10(1):6423, 2020.

\bibitem{zahavy2016Graying}
Tom Zahavy, Nir~Ben Zrihem, and Shie Mannor.
\newblock {Graying the Black Box: Understanding DQNs}.
\newblock In {\em Proceedings of the 33rd International Conference on
  International Conference on Machine Learning (ICML) - Volume 48}, page
  1899–1908, 2016.

\bibitem{arras2017relevant}
Leila Arras, Franziska Horn, Gr{\'e}goire Montavon, Klaus-Robert M{\"u}ller,
  and Wojciech Samek.
\newblock {``What is relevant in a text document?'': An interpretable machine
  learning approach}.
\newblock {\em PloS one}, 12(8):e0181142, 2017.

\bibitem{greydanus2018VisualizingAtari}
Samuel Greydanus, Anurag Koul, Jonathan Dodge, and Alan Fern.
\newblock {Visualizing and Understanding Atari Agents}.
\newblock In {\em International Conference on Machine Learning}, pages
  1792--1801. PMLR, 2018.

\bibitem{lapuschkin2016analyzing}
Sebastian Lapuschkin, Alexander Binder, Gr{\'e}goire Montavon, Klaus-Robert
  Muller, and Wojciech Samek.
\newblock {Analyzing classifiers: Fisher vectors and deep neural networks}.
\newblock In {\em Proceedings of the IEEE Conference on Computer Vision and
  Pattern Recognition}, pages 2912--2920, 2016.

\bibitem{lapuschkin2019unmasking}
Sebastian Lapuschkin, Stephan W{\"a}ldchen, Alexander Binder, Gr{\'e}goire
  Montavon, Wojciech Samek, and Klaus-Robert M{\"u}ller.
\newblock {Unmasking Clever Hans predictors and assessing what machines really
  learn}.
\newblock {\em Nature communications}, 10(1):1096, 2019.

\bibitem{anders2019SPRAY}
Christopher~J Anders, Talmaj Marin{\v{c}}, David Neumann, Wojciech Samek,
  Klaus-Robert M{\"u}ller, and Sebastian Lapuschkin.
\newblock {Analyzing ImageNet with Spectral Relevance Analysis: Towards
  ImageNet un-Hans'ed}.
\newblock {\em arXiv preprint arXiv:1912.11425}, 2019.

\bibitem{fragile}
Amirata Ghorbani, Abubakar Abid, and James~Y. Zou.
\newblock {Interpretation of Neural Networks Is Fragile}.
\newblock In {\em The 33rd Conference on Artificial Intelligence, {AAAI}},
  pages 3681--3688, 2019.

\bibitem{blamingGeometry}
Ann-Kathrin Dombrowski, Maximillian Alber, Christopher Anders, Marcel
  Ackermann, Klaus-Robert M{\"u}ller, and Pan Kessel.
\newblock {Explanations can be manipulated and geometry is to blame}.
\newblock In {\em Advances in Neural Information Processing Systems}, pages
  13567--13578, 2019.

\bibitem{heo2019fooling}
Juyeon Heo, Sunghwan Joo, and Taesup Moon.
\newblock {Fooling Neural Network Interpretations via Adversarial Model
  Manipulation}.
\newblock In {\em Advances in Neural Information Processing Systems}, pages
  2921--2932, 2019.

\bibitem{fairwashing}
Christopher Anders, Plamen Pasliev, Ann-Kathrin Dombrowski, Klaus-Robert
  Müller, and Pan Kessel.
\newblock {Fairwashing Explanations with Off-Manifold Detergent}.
\newblock In {\em Proceedings of the 37th International Conference on Machine
  Learning, {ICML}, Vienna, Austria, PMLR 119}, 2020.

\bibitem{rightToExplanation}
Parliament and Council of~the European~Union.
\newblock {\em Article 22. Automated individual decision making, including
  profiling}.
\newblock Official Journal of the European Union, 2016.

\bibitem{evaluating}
Wojciech Samek, Alexander Binder, Gregoire Montavon, Sebastian Lapuschkin, and
  Klaus-Robert Müller.
\newblock {Evaluating the visualization of what a Deep Neural Network has
  learned}.
\newblock {\em IEEE Transactions on Neural Networks and Learning Systems},
  28:2660--2673, 11 2017.

\bibitem{arras2017ExplainingRNN}
Leila Arras, Gr{\'e}goire Montavon, Klaus-Robert M{\"u}ller, and Wojciech
  Samek.
\newblock Explaining recurrent neural network predictions in sentiment
  analysis.
\newblock In {\em Proceedings of the 8th Workshop on Computational Approaches
  to Subjectivity, Sentiment and Social Media Analysis}, pages 159--168,
  Copenhagen, Denmark, September 2017. Association for Computational
  Linguistics.

\bibitem{sanityChecksMetrics}
Richard Tomsett, Dan Harborne, Supriyo Chakraborty, Prudhvi Gurram, and Alun
  Preece.
\newblock {Sanity Checks for Saliency Metrics}.
\newblock {\em Proceedings of the AAAI Conference on Artificial Intelligence},
  (04):6021--6029, Apr. 2020.

\bibitem{roar}
Sara Hooker, Dumitru Erhan, Pieter-Jan Kindermans, and Been Kim.
\newblock {A Benchmark for Interpretability Methods in Deep Neural Networks}.
\newblock In {\em Advances in Neural Information Processing Systems}, pages
  9734--9745, 2019.

\bibitem{sanity}
Julius Adebayo, Justin Gilmer, Michael Muelly, Ian~J. Goodfellow, Moritz Hardt,
  and Been Kim.
\newblock {Sanity Checks for Saliency Maps}.
\newblock In {\em Advances in Neural Information Processing Systems 31: Annual
  Conference on Neural Information Processing Systems, NeurIPS 2018, 3-8
  December, Montr{\'{e}}al, Canada.}, pages 9525--9536, 2018.

\bibitem{Sixt2020WhenExplLie}
Leon Sixt, Maximilian Granz, and Tim Landgraf.
\newblock {When Explanations Lie: Why Many Modified BP Attributions Fail}.
\newblock In {\em Proceedings of the 37th International Conference on Machine
  Learning, {ICML}, Vienna, Austria, PMLR 119}, 2020.

\bibitem{cat}
Pieter{-}Jan Kindermans, Sara Hooker, Julius Adebayo, Maximilian Alber,
  Kristof~T. Sch{\"{u}}tt, Sven D{\"{a}}hne, Dumitru Erhan, and Been Kim.
\newblock {The (Un)reliability of Saliency Methods}.
\newblock In {\em Explainable {AI:} Interpreting, Explaining and Visualizing
  Deep Learning}, pages 267--280. Springer, 2019.

\bibitem{wang2020GradientNLPManipulable}
Junlin Wang, Jens Tuyls, Eric Wallace, and Sameer Singh.
\newblock {Gradient-based Analysis of NLP Models is Manipulable}.
\newblock {\em arXiv preprint arXiv:2010.05419}, 2020.

\bibitem{FoolingLimeAndShap}
Dylan Slack, Sophie Hilgard, Emily Jia, Sameer Singh, and Himabindu Lakkaraju.
\newblock {Fooling LIME and SHAP: Adversarial Attacks on Post hoc Explanation
  Methods}.
\newblock In {\em Proceedings of the AAAI/ACM Conference on AI, Ethics, and
  Society}, pages 180--186, 2020.

\bibitem{Wang2020SmoothedGeometry}
Zifan Wang, Haofan Wang, Shakul Ramkumar, Piotr Mardziel, Matt Fredrikson, and
  Anupam Datta.
\newblock {Smoothed Geometry for Robust Attribution}.
\newblock {\em Advances in Neural Information Processing Systems}, 33, 2020.

\bibitem{rieger2020simple}
Laura Rieger and Lars~Kai Hansen.
\newblock {A simple defense against adversarial attacks on heatmap
  explanations}.
\newblock {\em 2020 Workshop on Human Interpretability in Machine Learning
  (WHI)}, 2020.

\bibitem{lakkaraju2020RobustBlackBoxExpl}
Himabindu Lakkaraju, Nino Arsov, and Osbert Bastani.
\newblock {Robust and stable black box explanations}.
\newblock In {\em Proceedings of the 37th International Conference on Machine
  Learning, {ICML}, Vienna, Austria, PMLR 119}, 2020.

\bibitem{RobustnessViaCurvReg}
Seyed-Mohsen Moosavi-Dezfooli, Alhussein Fawzi, Jonathan Uesato, and
  P.~Frossard.
\newblock {Robustness via Curvature Regularization, and Vice Versa}.
\newblock {\em 2019 IEEE/CVF Conference on Computer Vision and Pattern
  Recognition (CVPR)}, pages 9070--9078, 2019.

\bibitem{Pearlmutter1994FastExactHessian}
Barak~A. Pearlmutter.
\newblock {Fast Exact Multiplication by the Hessian}.
\newblock {\em Neural Computation}, 6(1):147–160, January 1994.

\bibitem{WeightDecay_krogh}
Anders Krogh and John~A Hertz.
\newblock {A Simple Weight Decay Can Improve Generalization}.
\newblock In {\em Advances in neural information processing systems}, pages
  950--957, 1992.

\bibitem{WeightDecay_hanson}
Stephen Hanson and Lorien Pratt.
\newblock {Comparing biases for minimal network construction with
  back-propagation}.
\newblock {\em Advances in neural information processing systems}, 1:177--185,
  1988.

\bibitem{WeightDecay_weigend}
Andreas~S Weigend, David~E Rumelhart, and Bernardo~A Huberman.
\newblock {Generalization by Weight-Elimination with Application to
  Forecasting}.
\newblock In {\em Advances in neural information processing systems}, pages
  875--882, 1991.

\bibitem{GoodfellowBook}
Ian Goodfellow, Yoshua Bengio, and Aaron Courville.
\newblock {\em {Deep Learning}}.
\newblock MIT Press, 2016.
\newblock \url{http://www.deeplearningbook.org}.

\bibitem{cifar}
Alex Krizhevsky.
\newblock {Learning Multiple Layers of Features from Tiny Images, CIFAR10
  dataset}.
\newblock Technical report, 2009.

\bibitem{montavon2019LRPOverview}
Gr{\'e}goire Montavon, Alexander Binder, Sebastian Lapuschkin, Wojciech Samek,
  and Klaus-Robert M{\"u}ller.
\newblock {Layer-Wise Relevance Rropagation: An Overview}.
\newblock In {\em Explainable AI: interpreting, explaining and visualizing deep
  learning}, pages 193--209. Springer, 2019.

\bibitem{JMLR:v9:braun08a}
Mikio~L. Braun, Joachim~M. Buhmann, and Klaus-Robert M{{\"u}}ller.
\newblock {On Relevant Dimensions in Kernel Feature Spaces}.
\newblock {\em Journal of Machine Learning Research}, 9(62):1875--1908, 2008.

\bibitem{JMLR:v12:montavon11a}
Gr{{\'e}}goire Montavon, Mikio~L. Braun, and Klaus-Robert M{{\"u}}ller.
\newblock {Kernel Analysis of Deep Networks}.
\newblock {\em Journal of Machine Learning Research}, 12(78):2563--2581, 2011.

\bibitem{kauffmann2019clustering}
Jacob Kauffmann, Malte Esders, Gr{\'e}goire Montavon, Wojciech Samek, and
  Klaus-Robert M{\"u}ller.
\newblock {From clustering to cluster explanations via neural networks}.
\newblock {\em arXiv preprint arXiv:1906.07633}, 2019.

\bibitem{kauffmann2020towards}
Jacob Kauffmann, Klaus-Robert M{\"u}ller, and Gr{\'e}goire Montavon.
\newblock {Towards explaining anomalies: a deep Taylor decomposition of
  one-class models}.
\newblock {\em Pattern Recognition}, 101:107198, 2020.

\bibitem{ruff2020unifying}
Lukas Ruff, Jacob~R Kauffmann, Robert~A Vandermeulen, Gr{\'e}goire Montavon,
  Wojciech Samek, Marius Kloft, Thomas~G Dietterich, and Klaus-Robert
  M{\"u}ller.
\newblock {A Unifying Review of Deep and Shallow Anomaly Detection}.
\newblock {\em arXiv preprint arXiv:2009.11732}, 2020.

\bibitem{eberle2020building}
Oliver Eberle, Jochen B{\"u}ttner, Florian Kr{\"a}utli, Klaus-Robert
  M{\"u}ller, Matteo Valleriani, and Gr{\'e}goire Montavon.
\newblock {Building and Interpreting Deep Similarity Models}.
\newblock {\em arXiv preprint arXiv:2003.05431}, 2020.

\end{thebibliography}
\bibliographystyle{unsrt}%

\clearpage
\appendix

\begin{center}
\huge{Appendix}
\end{center}
\vspace{1cm}
\tableofcontents
\unhidefromtoc
\section{Proof of Theorem 1}\label{appndx:proofTheorem1}
Let $\sigma(x)$ denote the non-linearity of the network. We also use the notation $\sigma^{(l)}(x) = \sigma (W^{(l)} x)$ where $W^{(l)}$ are the weights of layer $l$. By assumption, the activation functions $\sigma$ are twice-differentiable and bounded
\begin{align}
    |\sigma'(x)| \le \Sigma_1 \,, && |\sigma''(x)| \le \Sigma_2 \,.
\end{align}

The activation at layer $L$ is then given by
\begin{align}
    a^{(L)}(x) = (\sigma^{(L)} \circ \dots \circ \sigma^{(1)})(x)
\end{align}
Its derivative $\partial_k a^{(l)}_i$ is equal to
\begin{align*}
     \sum_{s_2 \dots s_l} W^{(l)}_{i s_l} \sigma'\left(\sum_j W^{(l)}_{ij} a^{(l-1)}_j\right) W^{(l-1)}_{s_{l} s_{l-1}} \sigma'\left(\sum_j W^{(l-1)}_{s_{l}j} a^{(l-2)}_j\right)\\
     \hfill\dots W^{(1)}_{s_{2}k} \sigma'\left(\sum_j W^{(1)}_{s_2 j} x_j\right) \,.
\end{align*}
We therefore obtain
\begin{align}
    \left \| \nabla a^{(l)} \right\|_F \le \left(\Sigma_1\right)^l \, \prod_{i=1}^l \left\| W^{(i)} \right\|_F
\end{align}
From the expression for $\partial_k a^{(l)}_i$, we can straightforwardly derive that
\begin{align*}
    \partial_l \partial_k a^{(L)}_i = &\sum_m \sum_{s_2 \dots s_L} \big\{ \\
    &W^{(L)}_{i s_L} \sigma'\left(\sum_j W^{(L)}_{ij} a^{(L-1)}_j\right) W^{(L-1)}_{s_{L} s_{L-1}} \sigma'\left(\sum_j W^{(L-1)}_{s_L j} a^{(L-2)}_j\right) \\
    & \dots \sum_{p} W^{(m)}_{s_{m+1} p} W^{(m)}_{s_{m+1}s_{m}} \sigma''\left(\sum_j W^{(m)}_{s_{m+1} j} a^{(m-1)}_j(x)\right) \partial_{l} a^{(m-1)}_{p}(x) \\
    & \dots W^{(1)}_{s_2 k} \sigma'\left(\sum_j W^{(1)}_{s_2 j} x_j\right) \big\} \,.
\end{align*}
Restrict to the case for which the index $i$ only takes a single value, the Hessian $H_{ij}(g)=\partial_i \partial_j a^L(x)$ is then bounded by
\begin{align}
    \left \| H (g) \right\|_F  \le  \sum_{m=1}^L \left( \prod_{l=1}^m ||W^{(l)}||_F^2 \prod_{l=m+1}^L ||W^{(l)}||_F \right) \, \Sigma_1^{L+m-2} \, \Sigma_2 \,.
\end{align}

\section{Relu networks}\label{app:relu}

As was discussed in the main text, for Softplus non-linearites the bound \eqref{eq:masterbound} diverges for ReLU non-linearities. This is because ReLU can be obtained from Softplus by taking the limit $\beta \to \infty$ and the constant $\Sigma_2$ in \eqref{eq:masterbound} diverges in this limit, see \eqref{eq:Softplusbound}. The underlying fundamental difficulty is that the Hessian of ReLU networks is not well-defined. 

In the following, we will discuss how to generalize the analysis to networks with ReLU activations. We will establish that a distributional generalization of the Hessian can be derived for ReLU networks. A distributional form of the Hessian is sufficient for our purposes because in deriving a bound for the maximal change in explanation we only need to consider the Hessian under an integral, see \eqref{eq:gradienttrick}. Since integrals over distributions are well-defined, the resulting expression will be well-defined as well.

In the following, we will first illustrate this for a simple toy model before considering the general case.

\subsection{Toy example}

\begin{figure}[htp!]
\centering
\begin{minipage}{.45\textwidth}
  \centering
  \includegraphics[width=1\textwidth]{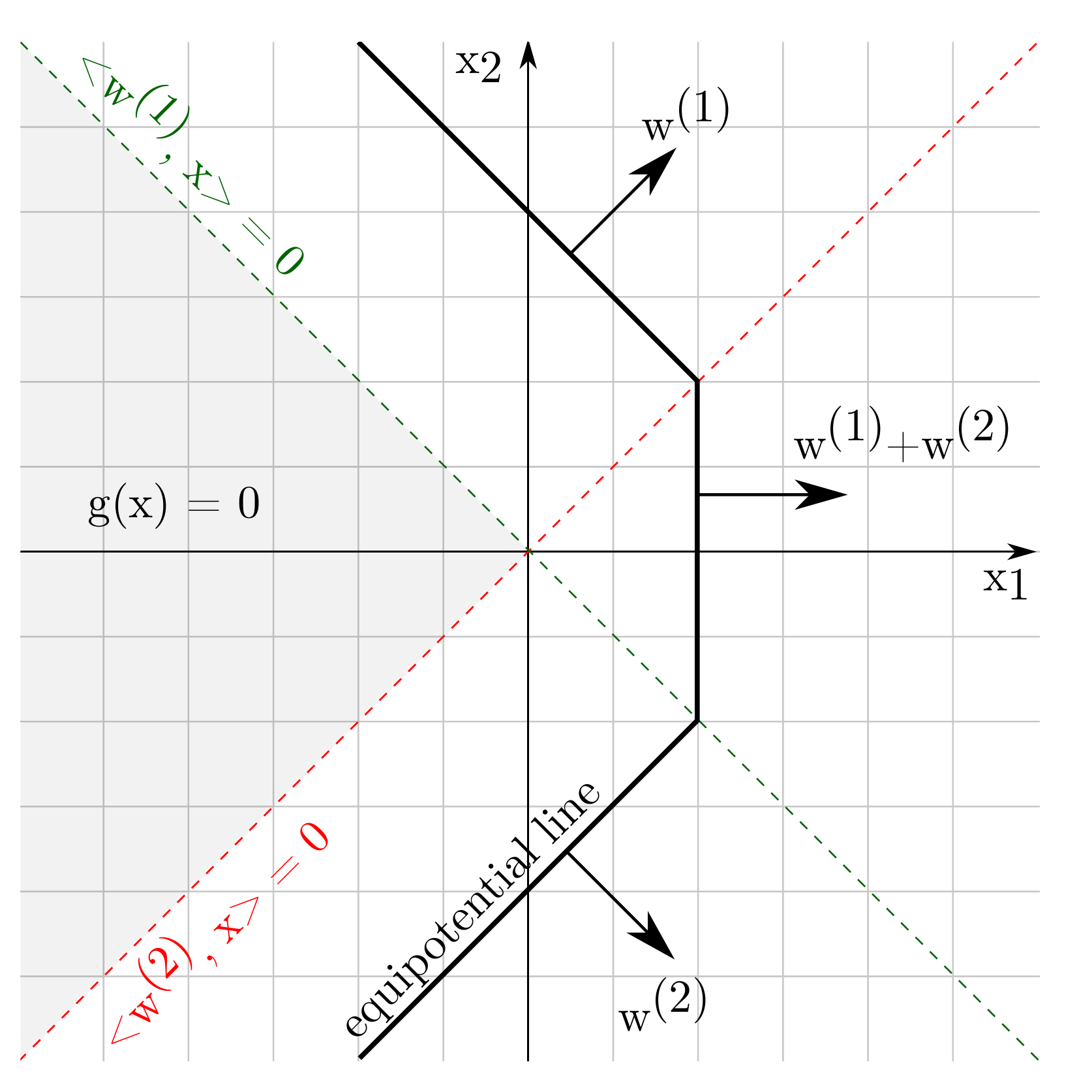}
  \caption{Toy function. Lines of rootpoints are marked in green and red. The grey area shows where $g(x) = 0$}
  \label{fig:toy_function}
\end{minipage}\hfill
\begin{minipage}{.45\textwidth}
  \centering
  \includegraphics[width=1\textwidth]{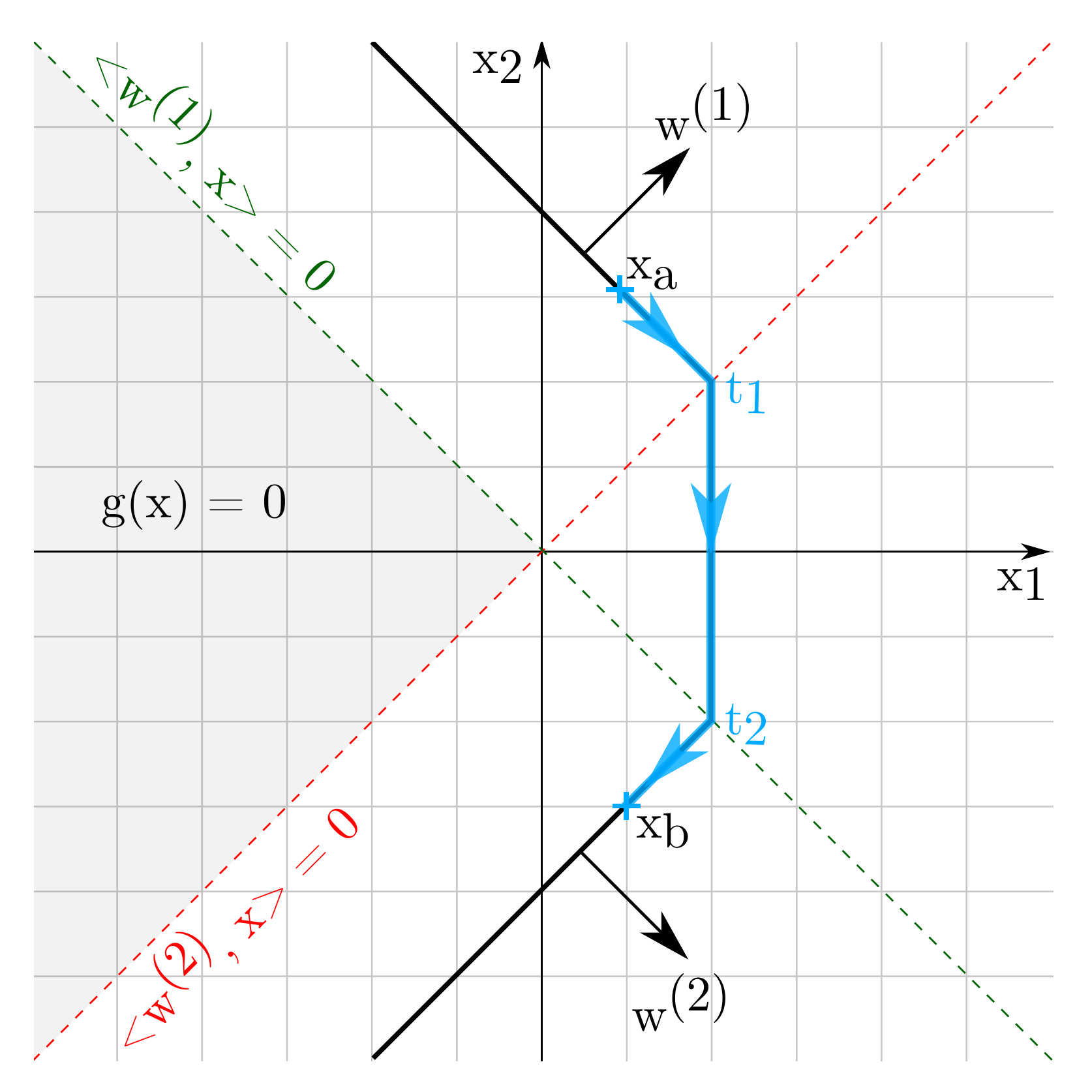}
  \caption{Path (in blue) along an equipotential line (constant network output $g(x)=\text{const}$)}
  \label{fig:blue_path}
  \end{minipage}
\end{figure}

Consider the network (depicted in Figure~\ref{fig:toy_function})
\begin{equation}
g(x) = \relu(w ^{(1)T}x) + \relu(w ^{(2)T}x)
\end{equation}

with input vector $x\in \R^2$ and weight vectors $w^{(1)} = \frac{1}{\sqrt{2}}[1, 1]^T$ and $w^{(2)} = \frac{1}{\sqrt{2}}[1, -1]^T$. 
The first derivative with respect to $x_j$ is
\begin{equation}
\partial_j g(x) = w^{(1)}_j \theta(w ^{(1)T}x) + w^{(2)}_j \theta(w ^{(2)T}x) \,,
\end{equation}
where we have defined the Heaviside step function
\begin{align}
    \theta(x) = \begin{cases} 
      1 & x\ge 0 \,,\\
      0 & x < 0 \,. 
   \end{cases}
\end{align}
We note that the derivative of the step function is not well-defined at zero. However, a distributional generalization thereof can be defined, i.e.
\begin{align}
    \theta'(x) = \delta(x) \,, 
\end{align}
where $\delta$ denotes the Dirac delta distribution. 

With this definition, the $ij$-th entry of (the distributional generalization of) the Hessian matrix can formally be written as
\begin{equation}
\partial_i\partial_j g(x) = w^{(1)}_i w^{(1)}_j \delta(w ^{(1)T}x) + w^{(2)}_i w^{(2)}_j \delta(w ^{(2)T}x) \,.
\end{equation}
By \eqref{eq:gradienttrick}, the change in (gradient) explanation when moving from point $x$ to $x_{\textrm{adv}}$ is then given by 
\begin{equation}
(h(x)-h(x_{\textrm{adv}}))_j =  \int_{-\infty}^\infty \sum_i  \left(w^{(1)}_i w^{(1)}_j \delta(w ^{(1)T}x) + w^{(2)}_i w^{(2)}_j \delta(w ^{(2)T}x)\right) \dot{x}_i  \, \d t\,,
\end{equation}
where we have used the notation $x(t)$ for the curve connecting the unperturbed and perturbed data points. 

For integrating over the delta distribution in composition with a (scalar-valued) function, we use
\begin{equation}
\int_{-\infty}^\infty f(t)\delta(y(t))\d t = \sum_{t_N} \frac{f(t_N)}{\abs{y'(t_N)}}
\label{eq:integral_to_sum_over_roots}
\end{equation}
with $t_N$ being the roots of $y(t)$. Using this expression, we then obtain the following change in saliency map
\begin{align*}
(h(x)-h(x_{\textrm{adv}}))_j &= \sum_{t_N} \frac{\sum_i w^{(1)}_i w^{(1)}_j \dot{x}_i}{\abs{\sum_i w^{(1)}_i \dot{x}_i}} + \sum_{t_N} \frac{\sum_i w^{(2)}_i w^{(2)}_j \dot{x}_i}{\abs{\sum_i w^{(2)}_i \dot{x}_i}}\\
&=\sum_{t_N} \sgn\left(\sum_i w^{(1)}_i \dot{x}_i\right) w^{(1)}_j + \sum_{t_N} \sgn\left(\sum_i w^{(2)}_i \dot{x}_i\right) w^{(2)}_j \,.
\end{align*}
Consider the blue path in Figure~\ref{fig:blue_path} whose root points are denoted by $t_1$ and $t_2$. We note that these root points correspond to kinks in the curve $x(t)$ connecting the unperturbed and perturbed data point. Their corresponding normalized velocity vectors are given by $\dot{x}(t_1) = w^{(2)}$ and $\dot{x}(t_2) = (0, -1)^T$ respectively. We therefore obtain
\begin{align*}
h(x)-h(x_{\textrm{adv}}) &= \sgn\left( \langle w^{(1)}, \dot{x}(t_2) \rangle \right ) w^{(1)} + \sgn\left(\langle w^{(2)}, \dot{x}(t_1) \rangle \right) w^{(2)}\\
&= \sgn\left(-\frac{1}{\sqrt{2}}\right) w^{(1)} + \sgn\left(1\right) w^{(2)}\\
&=w^{(2)}-w^{(1)}
\end{align*}
which is correct as $h(x) = w^{(2)}$ and $h(x_{\textrm{adv}}) = w^{(1)}$. It is important to stress that we have obtain this result despite the fact that the Hessian of the neural network $g$ is only given in generalized distributional form. 

\subsection{General case}\label{sec:UpperBound}
The argument of the previous section can be generalized to arbitrary fully-connected networks with weights $W^l$ of layer $l \in \{1,\dots, L\}$. The general logic follows closely the toy model discussed in the previous section, i.e. a distributional generalization of the Hessian is derived and since on the right-hand-side of \eqref{eq:gradienttrick} the Hessian only appears under an integral, this distributional form is sufficient to obtain a bound on the maximal change in explanation due to a perturbation of the input. Using this technique, we derive the following theorem: 
\begin{theorem}
Let $x$ and $x_{\textrm{adv}}=x+\delta x$ denote the unperturbed and perturbed data points respectively. We denote by $x(t)$ the curve connecting the unperturbed and perturbed points, i.e. $x(t=-\infty)=x$ and $x(t=+\infty)=x_{\textrm{adv}}$. Furthermore, we assume that all points on the curve have the same network output, i.e. $g(x(t_1))=g(x(t_2))$ for all $t_1, t_2 \in \mathbb{R}$. The maximal change of explanation is then given by
\begin{align}
    \left || h(x) - h(x_{\textrm{adv}})\right ||^2 \le  \sum_{\textrm{kinks}(x(t))} \left( \prod_{l=1}^L \left \| W^{(l)} \right\|_F^2 \right)   \,,
\end{align}
where the sum runs over all kinks of the curve $x(t)$.
\end{theorem}
We can give an intuition for the theorem by considering the blue curve of Figure~\ref{fig:blue_path} for the toy model of the previous section. In this case, the sum over the kinks would run over $x(t_1)$ and $x(t_2)$, see Figure~\ref{fig:blue_path}. Only at these kinks, the gradient of the network will change. In the theorem, we then estimate this change by its maximal value, i.e. the change is equal to the product of all weights. 

As a practical consequence of the theorem, we can make explanations more robust by weight decay also in the case of ReLU non-linearities.  

\paragraph{Proof:}
Let $W^{(l)}$ be the weights of layer $l$. We denote the $l$-th layer by $\relu^{(l)}(x)=\relu(W^{(l)}x)$. It then follows that 
\begin{align}
    \partial_k \, \relu\left(\sum_j W_{ij} x_j\right) = W_{ik} \, \theta\left(\sum_j W_{ij} x_j\right) \\
    \partial_l \, \theta\left(\sum_j W_{ij} x_j\right) = W_{il} \, \delta\left(\sum_j W_{ij} x_j\right)
\end{align}
where $\theta$ and $\delta$ are the Heaviside step function and the delta distribution respectively.
The activation at layer $L$ is then given by
\begin{align}
    a^{(L)}(x) = (\relu^{(L)} \circ \dots \circ \relu^{(1)})(x)
\end{align}
Its derivative $\partial_k a^{(L)}_i$ is equal to
\begin{align*}
     \sum_{s_2 \dots s_L} W^{(L)}_{i s_L} \theta\left(\sum_j W^{(L)}_{ij} a^{(L-1)}_j\right) W^{(L-1)}_{s_{L} s_{L-1}} \theta\left(\sum_j W^{(L-1)}_{s_{L}j} a^{(L-2)}_j\right) \\
     \hfill\dots W^{(1)}_{s_{2}k} \theta\left(\sum_j W^{(1)}_{s_2 j} x_j\right)
\end{align*}
Deriving this expression for $\partial_k a^{(L)}_i$ again, we obtain
\begin{align*}
    \partial_l \partial_k a^{(L)}_i = &\sum_m \sum_{s_2 \dots s_L} \big\{ \\
    &W^{(L)}_{i s_L} \theta\left(\sum_j W^{(L)}_{ij} a^{(L-1)}_j\right) W^{(L-1)}_{s_{L} s_{L-1}} \theta\left(\sum_j W^{(L-1)}_{s_L j} a^{(L-2)}_j\right) \\
    & \dots \sum_{p} W^{(m)}_{s_{m+1} p} W^{(m)}_{s_{m+1}s_{m}} \delta\left(\sum_j W^{(m)}_{s_{m+1} j} a^{(m-1)}_j(x)\right) \partial_{l} a^{(m-1)}_p(x) \\
    & \dots W^{(1)}_{s_2 k} \theta\left(\sum_j W^{(1)}_{s_2 j} x_j\right) \big\}
\end{align*}
We now restrict to the case that $a^{(L)}$ has only a single output value. As a result, the index $i$ in the expression above only takes one value, i.e. $i=1$. We define $g(x)=a^{(L)}_1(x)$ to ease notation.
We then substitute this expression for $\partial_l \partial_k g = \partial_l \partial_k a^{(L)}_1$ in \eqref{eq:gradienttrick} and obtain
\begin{align*}
\left( h(x) - h(x_{\textrm{adv}})\right)_k &=\sum_m \sum_{s_2 \dots s_L}  \int_{-\infty}^\infty \d t \, \Big\{ \\
    &W^{(L)}_{1 s_L} \theta\left(\sum_j W^{(L)}_{ij} a^{(L-1)}_j\right) W^{(L-1)}_{s_{L} s_{L-1}} \theta\left(\sum_j W^{(L-1)}_{s_L j} a^{(L-2)}_j\right) \\
    & \dots \sum_{\hat{s}_m} W^{(m)}_{s_{m+1} \hat{s}_m} W^{(m)}_{s_{m+1}s_{m}} \delta\left(\sum_j W^{(m)}_{s_{m+1} j} a^{(m-1)}_j(x)\right) \dot{a}^{(m-1)}_{\hat{s}_m}(x) \\
    & \dots W^{(1)}_{s_2 k} \theta\left(\sum_j W^{(1)}_{s_2 j} x_j\right) \Big\}\,,
\end{align*}
where we have used the notation $\partial_t a^{(m-1)} = \dot{a}^{(m-1)}$ for notational simplicity. Using the identity \eqref{eq:integral_to_sum_over_roots}, we then obtain
\begin{align*}
\left( h(x) - h(x_{\textrm{adv}})\right)_k &= \sum_m \sum_{x_N^m}  \sum_{s_2 \dots s_L}  \, \Big\{ \\
    &W^{(L)}_{1 s_L} \theta\left(\sum_j W^{(L)}_{ij} a^{(L-1)}_j\right) W^{(L-1)}_{s_{L} s_{L-1}} \theta\left(\sum_j W^{(L-1)}_{s_L j} a^{(L-2)}_j\right) \\
    & \dots W^{(m)}_{s_{m+1}s_{m}} \text{sgn}\left(\sum_j W^{(m)}_{s_{m+1} j} \dot{a}^{(m-1)}_j(x^m_N)\right)  \\
    & \dots W^{(1)}_{s_2 k} \theta\left(\sum_j W^{(1)}_{s_2 j} (x^m_N)_j\right) \Big\}\,,
\end{align*}
where the sum over $x^m_N$ runs over all zeropoints of $\sum_j W^{(m)}_{s_{m+1} j} a^{(m-1)}$ along the trajectory connecting $x$ with $x_{\textrm{adv}}$. Using the fact that  $|\theta(\bullet)| \le 1$ and  $|\text{sgn}(\bullet)| \le 1$, we obtain
\begin{align}
    \left || h(x) - h(x_{\textrm{adv}})\right ||^2 \le  \sum_m \sum_{x_N^m} \left \| W^{(L)} \right\|_F^2 \left \| W^{(L-1)} \right\|_F^2 \dots \left \| W^{(m)} \right\|_F^2 \dots \left \| W^{(1)} \right\|_F^2 \,.
\end{align}
As in the case of the toymodel, the summands run over all kinks of the trajectory.
This bound for ReLU networks depends purely on the network weights and the number of kinks passed when moving from $x$ to $x_{\textrm{adv}}$. If we reduce the Frobenius norms of the weights, we also reduce the maximal possible change in explanation. 

\section{Interchangeability of Softplus \texorpdfstring{$\beta$}{beta}}\label{appndx:InterchangeabilitySoftplus}
\subsection{Interchangeability of Softplus \texorpdfstring{$\beta$}{beta}}\label{sec:InterschangeabilityBeta}

When training Softplus networks with different $\beta$ values it is interesting to consider how they differ as the beta values partially cancel out or can be absorbed into the weights and biases.

The Softplus function is defined as:

\begin{equation}
\text{sp}_\beta(x) = \frac{1}{\beta}\ln{(1+\text{e}^{\beta x})}    
\end{equation}

Therefore, we can relate two Softplus functions with different $\beta$ values $\beta_1$ and $\beta_2$ as follows:

\begin{equation}
    \text{sp}_{\beta_1}(x)=\frac{\beta_2}{\beta_1} \text{sp}_{\beta_2}(\frac{\beta_1}{\beta_2} x)
\end{equation}

A network consisting of linear layers and Softplus activations with $\beta=\beta_1$ has weights $W^{(i)}$ and biases $b^{(i)}$.
We can define a network with the same structure but a different Softplus $\beta=\beta_2$ and weights $\tilde{W}^{(i)}$ and biases $\tilde{b}^{(i)}$. The networks give identical outputs for all inputs if we define the weights and biases of the second network in the following way:
\begin{align*}
    \tilde{W}^{(1)} &=\frac{\beta_1}{\beta_2} W^{(1)}\\ 
    \tilde{W}^{(i)} &=  W^{(i)},\quad \forall i: 1<i<n\\
    \tilde{W}^{(n)} &=\frac{\beta_2}{\beta_1} W^{(n)}\\ 
    \tilde{b}^{(i)} &=  \frac{\beta_1}{\beta_2} b^{(i)},\quad \forall i: i<n\\
    \tilde{b}^{(n)} &=  b^{(n)}
\end{align*}
However, this mapping is not learned when training networks with different $\beta$ values from scratch as the distribution over weight norms stays very similar while the distribution changes drastically when artificially changing the $\beta$ value as demonstrated above. We show this effect for a few examples in~\ref{sec:InterschangeabilityBetaExamples}. 

\subsection{Examples}\label{sec:InterschangeabilityBetaExamples}
In Section~\ref{sec:InterschangeabilityBeta}, we show that, by adjusting weights and biases of a Softplus network, it can be functionally equivalent to a network with the same structure but different $\beta$ value for the Softplus activation. Artificially constructing networks in this way leads to a larger variance in the weight norms. Even when no weight decay is used during training the weight norms of the different network layers in one network tend to vary within one order of magnitude. 

\begin{figure}[htp!]
  \centering
  \includegraphics[width=.8\linewidth]{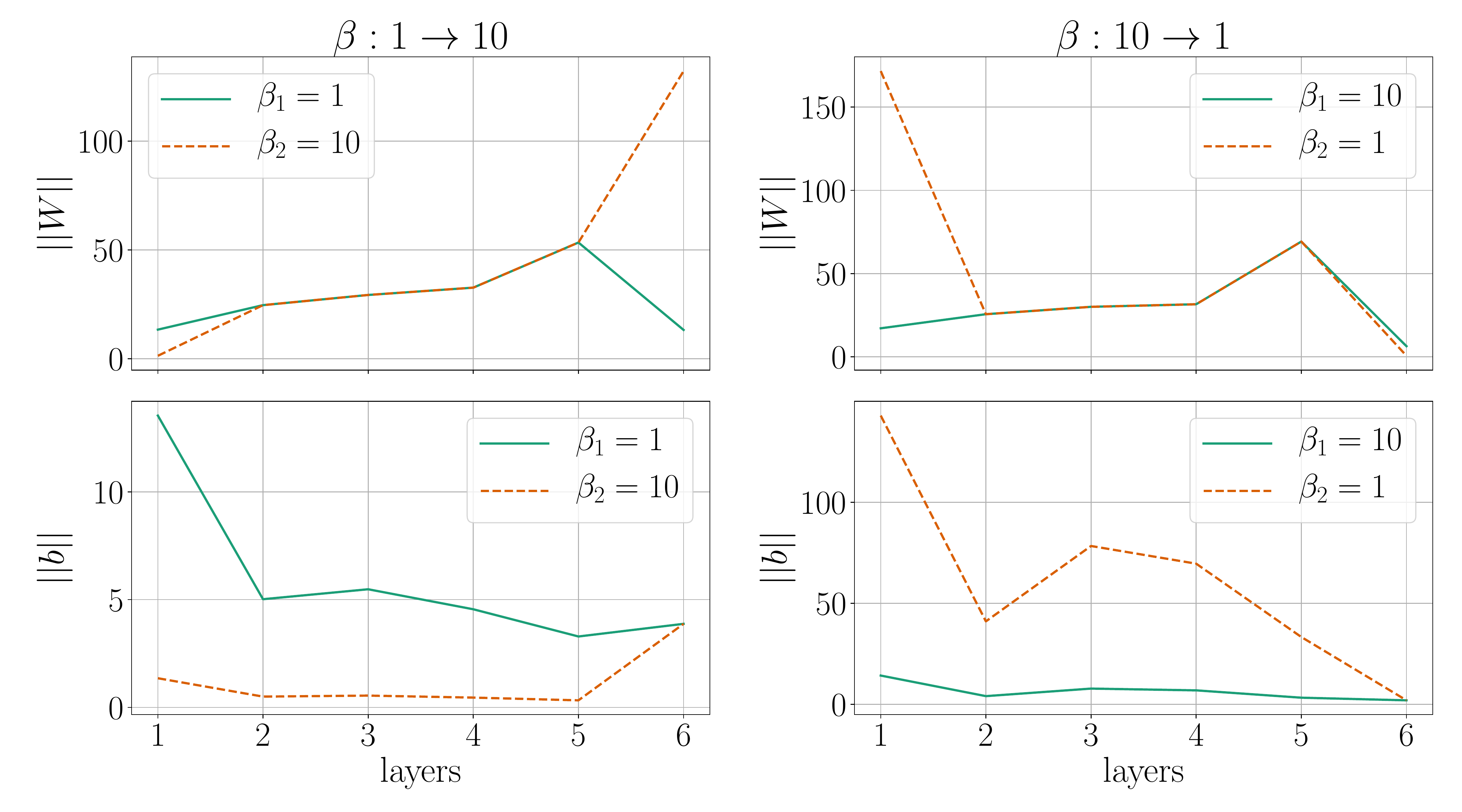}
  \caption{Weights and biases for networks with identical outputs but different $\beta$ value for the Softplus activation. Left: $\beta$ was changed from 1 to 10. Right: $\beta$ was changed from 10 to 1}
  \label{fig:weights_and_biases}
\end{figure}

Figure~\ref{fig:weights_and_biases} shows the weights and biases for two networks from Table~\ref{tab:statiscticsAllNets} with $\beta=1$ and $\beta=10$ and the respective weights and biases for two networks that produce identical output but have changed $\beta$ values. In both cases the average weight norm of the constructed network is higher than of the original.

\begin{figure}[htp!]
  \centering
  \includegraphics[width=1\linewidth]{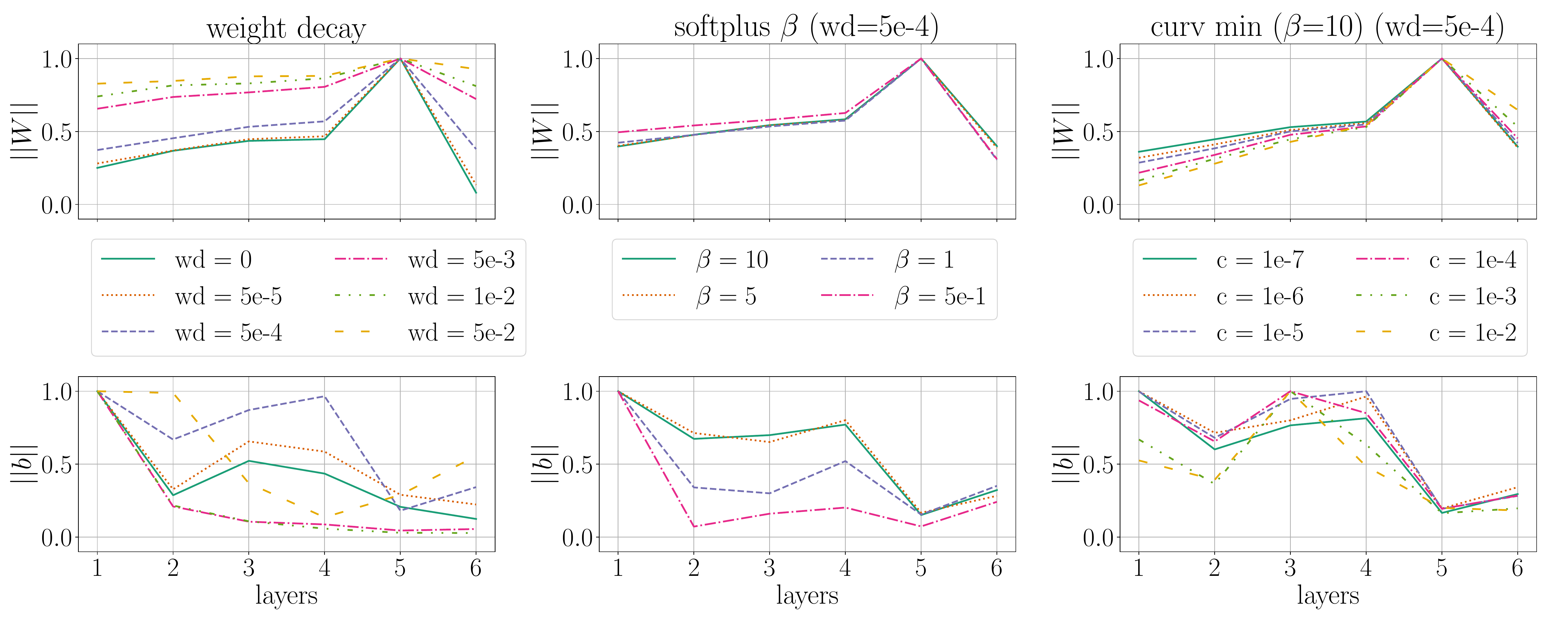}
  \caption{Weights and biases plotted over layers for different networks (left column: networks with weight decay, middle column: networks with different $\beta$ values, right column: networks trained with curvature minimization}
  \label{fig:weights_and_biases_distributions}
\end{figure}

Figure~\ref{fig:weights_and_biases_distributions} shows weights and biases for some of our networks from Table~\ref{tab:statiscticsAllNets}. The weight and bias norms for each network are normalized with the respective maximum value over all layers. Without exception the highest weight norm is found in layer 5 in contrast to the maximum weight norm when we do the artificial $\beta$ value switch. Thus training Softplus networks from scratch does produce fundamentally different networks that cannot be obtained with a mere rescaling of weights and biases.

\section{Experimental Analysis}\label{appndx:experiments}

\subsection{Error measures}

In this section we define the error measures we use to quantify our results. To ease notation we refer to the explanation of our original image as $h(x)=u$ and to the explanation of our perturbed image as $h(x_{\textrm{adv}})=v$. both explanations can be expressed as a vector of length $n$.

\begin{itemize}
\item The \textbf{Mean squared error (MSE)} between two explanations is defined as
\begin{equation*}
    \text{MSE}(u,v) = \frac{1}{n} \sum_i^n(u_i-v_i)^2
\end{equation*}
\item The \textbf{Pearson correlation coefficient (PCC)} between two explanations is given by
    \begin{equation*}
        \text{PCC}(u,v) = \frac{\sum_i^n(u_i-\mu_{u})(v_i-\mu_{v})}{\sqrt{\sum_i^n (u_i-\mu_{u})^2}\sqrt{\sum_i^n (v_i-\mu_{v})^2}}
    \end{equation*}
    where $\mu_{u} = \frac{1}{n}\sum_i^n u_i$ is the mean value of explanation $u$. Analogously for $\mu_{v}$.
 \item The \textbf{Structural similarity index (SSIM)} between two explanations is given by calculating
 \begin{equation*}
     \text{SSIM}(u,v) = \frac{(2\mu_u\mu_v + (k_1 L)^2)(2\sigma_{uv}+(k_2 L)^2)}{(\mu_u^2 + \mu_v^2 + (k1 L)^2)(\sigma_u^2 + \sigma_v^2 + (k_2 L)^2)}
 \end{equation*}
 for a $7\times7$ window centered at each pixel and then averaging over all windows.
 Mean values $\mu_u$ and $\mu_v$ and variances $\sigma_{uv}$, $\sigma_u$ and $\sigma_v$ are calculated separately for each $7\times7$ window. $L$ is the range between the largest and smallest value that a pixel in the explanation can have and $k_1=0.01$ and $k_2=0.03$ are constants to stabilize the division.
 \item The \textbf{accuracy (acc)} of a network is the ratio of correctly classified images of the test set:
 \begin{equation*}
     \text{acc} = \frac{correct\:predictions}{all\:predictions}
 \end{equation*}
\end{itemize}

\subsection{Network structure}\label{appndx:network}

\begin{figure}[htp!]
\begin{verbatim}
CNN_CIFAR(
  (features): Sequential(
    (conv0): Conv2d(3, 32, kernel_size=(3, 3), 
                    stride=(1, 1), padding=(1, 1))
    (acti0): ActivationFunction()
    (conv1): Conv2d(32, 32, kernel_size=(3, 3), 
                    stride=(1, 1), padding=(1, 1))
    (acti1): ActivationFunction()
    (pool2): MaxPool2d(kernel_size=2, stride=2,
                    padding=0, dilation=1, 
                    ceil_mode=False)
    (conv3): Conv2d(32, 64, kernel_size=(3, 3),
                    stride=(1, 1), padding=(1, 1))
    (acti3): ActivationFunction()
    (conv4): Conv2d(64, 64, kernel_size=(3, 3),
                    stride=(1, 1), padding=(1, 1))
    (acti4): ActivationFunction()
    (pool5): MaxPool2d(kernel_size=2, stride=2,
                    padding=0, dilation=1, 
                    ceil_mode=False)
  )
  (classifier): Sequential(
    (view0): Reshape()
    (dens0): Linear(in_features=4096, 
                    out_features=256, 
                    bias=True)
    (acti0): ActivationFunction()
    (dens1): Linear(in_features=256,
                    out_features=10, 
                    bias=True)
  )
)
\end{verbatim}
\caption{Setup of simple CNN for CIFAR-10}
\label{code:cnn_cifar}
\end{figure}

The structure of all networks trained within the scope of this work is depicted in Figure~\ref{code:cnn_cifar}. The activation function is either ReLU or Softplus (for the networks trained with $\beta$ smoothing or Hessian minimization). In order to focus on the robustness we aimed to train the different networks to similar accuracy (albeit no longer than 200 epochs). We use Stochastic Gradient Descent with momentum and learning rate decay. We do not perform any further hyperparameter optimization. Statistics for all trained networks are summarized in Table~\ref{tab:statiscticsAllNets}.

\begin{longtable}{||c | c | c | c | c | c | c | c | c ||} 
 \hline 
 $\lambda$ & $\beta$ & $\zeta$ & acc & \multicolumn{3}{c|}{pcc for different $\nu$} &  $||W||$ &  $||H||$ \\ 
  \hline 
 & & & & 0.005 & 0.01 & 0.025 & & \\[0.5ex] 
 \hline \hline 
 \endhead
0 & ReLU & 0.0 & 85.75 & 0.73 & 0.57 & 0.32 & 30.79 & - \\ 
5e-5 & ReLU & 0.0 & 86.38 & 0.77 & 0.62 & 0.36 & 23.28 & - \\ 
5e-4 & ReLU & 0.0 & 88.63 & 0.84 & 0.70 & 0.45 & 11.37 & - \\ 
5e-3 & ReLU & 0.0 & 86.10 & 0.89 & 0.80 & 0.62 & 4.80 & - \\ 
1e-2 & ReLU & 0.0 & 81.41 & 0.89 & 0.80 & 0.64 & 3.61 & - \\ 
0 & 10 & 0.0 & 85.61 & 0.81 & 0.63 & 0.34 & 30.01 & 503.61 \\ 
0 & 5 & 0.0 & 85.60 & 0.88 & 0.73 & 0.39 & 28.70 & 280.71 \\ 
0 & 1 & 0.0 & 85.60 & 0.93 & 0.85 & 0.61 & 27.76 & 59.21 \\ 
0 & 5e-1 & 0.0 & 84.51 & 0.94 & 0.88 & 0.67 & 28.84 & 39.06 \\ 
5e-5 & 10 & 0.0 & 86.36 & 0.86 & 0.70 & 0.39 & 22.91 & 298.79 \\ 
5e-5 & 5 & 0.0 & 86.33 & 0.91 & 0.78 & 0.46 & 22.97 & 154.94 \\ 
5e-5 & 1 & 0.0 & 86.03 & 0.94 & 0.86 & 0.62 & 22.73 & 51.63 \\ 
5e-5 & 5e-1 & 0.0 & 85.34 & 0.94 & 0.88 & 0.67 & 23.76 & 36.43 \\ 
5e-4 & 10 & 0.0 & 88.84 & 0.88 & 0.75 & 0.49 & 11.24 & 91.68 \\ 
5e-4 & 5 & 0.0 & 88.76 & 0.91 & 0.79 & 0.53 & 11.36 & 59.57 \\ 
5e-4 & 1 & 0.0 & 86.80 & 0.93 & 0.86 & 0.63 & 11.93 & 28.45 \\ 
5e-4 & 5e-1 & 0.0 & 85.36 & 0.93 & 0.86 & 0.67 & 10.62 & 16.78 \\ 
5e-3 & 10 & 0.0 & 86.13 & 0.91 & 0.82 & 0.64 & 4.81 & 9.54 \\ 
5e-3 & 5 & 0.0 & 85.44 & 0.91 & 0.83 & 0.64 & 4.76 & 7.49 \\ 
5e-3 & 1 & 0.0 & 83.35 & 0.92 & 0.86 & 0.71 & 4.86 & 3.79 \\ 
5e-3 & 5e-1 & 0.0 & 77.60 & 0.96 & 0.93 & 0.85 & 4.66 & 2.02 \\ 
1e-2 & 10 & 0.0 & 80.44 & 0.90 & 0.82 & 0.66 & 3.48 & 5.42 \\ 
1e-2 & 5 & 0.0 & 77.57 & 0.89 & 0.82 & 0.65 & 3.33 & 5.27 \\ 
1e-2 & 1 & 0.0 & 71.74 & 0.97 & 0.94 & 0.86 & 3.21 & 1.26 \\ 
1e-2 & 5e-1 & 0.0 & 72.03 & 0.98 & 0.95 & 0.89 & 3.35 & 0.97 \\ 
0 & 10 & 1e-7 & 85.65 & 0.95 & 0.87 & 0.60 & 24.72 & 56.77 \\ 
0 & 10 & 1e-6 & 85.74 & 0.97 & 0.92 & 0.73 & 22.45 & 21.47 \\ 
0 & 10 & 1e-5 & 85.56 & 0.98 & 0.95 & 0.84 & 20.59 & 8.49 \\ 
0 & 10 & 1e-4 & 84.12 & 0.99 & 0.97 & 0.90 & 19.14 & 3.23 \\ 
0 & 10 & 1e-3 & 82.40 & 0.99 & 0.98 & 0.94 & 17.91 & 1.19 \\ 
0 & 10 & 1e-2 & 80.07 & 0.99 & 0.98 & 0.94 & 17.08 & 0.43 \\ 
0 & 5 & 1e-7 & 86.26 & 0.95 & 0.88 & 0.64 & 25.44 & 49.58 \\ 
0 & 5 & 1e-6 & 85.94 & 0.97 & 0.92 & 0.74 & 23.41 & 20.85 \\ 
0 & 5 & 1e-5 & 85.87 & 0.98 & 0.95 & 0.83 & 21.88 & 7.91 \\ 
0 & 5 & 1e-4 & 84.81 & 0.98 & 0.97 & 0.90 & 20.74 & 2.96 \\ 
0 & 5 & 1e-3 & 83.12 & 0.99 & 0.98 & 0.93 & 19.72 & 1.29 \\ 
0 & 5 & 1e-2 & 80.95 & 0.99 & 0.98 & 0.94 & 19.02 & 0.41 \\ 
0 & 1 & 1e-7 & 85.24 & 0.94 & 0.88 & 0.69 & 27.69 & 31.61 \\ 
0 & 1 & 1e-6 & 85.17 & 0.95 & 0.90 & 0.75 & 26.29 & 17.67 \\ 
0 & 1 & 1e-5 & 84.85 & 0.97 & 0.94 & 0.83 & 25.11 & 7.62 \\ 
0 & 1 & 1e-4 & 84.70 & 0.98 & 0.95 & 0.87 & 24.25 & 3.03 \\ 
0 & 1 & 1e-3 & 82.68 & 0.98 & 0.96 & 0.90 & 23.23 & 1.16 \\ 
0 & 1 & 1e-2 & 81.57 & 0.98 & 0.95 & 0.89 & 22.20 & 0.42 \\ 
0 & 5e-1 & 1e-7 & 81.90 & 0.95 & 0.90 & 0.77 & 12.05 & 9.28 \\ 
0 & 5e-1 & 1e-6 & 85.46 & 0.96 & 0.91 & 0.77 & 28.07 & 15.20 \\ 
0 & 5e-1 & 1e-5 & 84.41 & 0.97 & 0.93 & 0.82 & 26.81 & 6.79 \\ 
0 & 5e-1 & 1e-4 & 84.08 & 0.98 & 0.96 & 0.89 & 25.59 & 2.86 \\ 
0 & 5e-1 & 1e-3 & 82.84 & 0.98 & 0.96 & 0.89 & 22.21 & 1.12 \\ 
0 & 5e-1 & 1e-2 & 81.04 & 0.98 & 0.96 & 0.90 & 23.36 & 0.41 \\ 
5e-5 & 10 & 1e-7 & 86.68 & 0.95 & 0.87 & 0.62 & 20.24 & 52.22 \\ 
5e-5 & 10 & 1e-6 & 86.47 & 0.97 & 0.92 & 0.74 & 18.75 & 21.36 \\ 
5e-5 & 10 & 1e-5 & 85.87 & 0.98 & 0.95 & 0.84 & 17.28 & 8.18 \\ 
5e-5 & 10 & 1e-4 & 84.55 & 0.99 & 0.97 & 0.90 & 16.05 & 3.29 \\ 
5e-5 & 10 & 1e-3 & 83.03 & 0.99 & 0.98 & 0.93 & 15.01 & 1.14 \\ 
5e-5 & 10 & 1e-2 & 80.47 & 0.99 & 0.98 & 0.94 & 13.88 & 0.43 \\ 
5e-5 & 5 & 1e-7 & 86.76 & 0.95 & 0.87 & 0.62 & 21.05 & 45.57 \\ 
5e-5 & 5 & 1e-6 & 86.41 & 0.97 & 0.92 & 0.74 & 19.57 & 19.16 \\ 
5e-5 & 5 & 1e-5 & 86.16 & 0.98 & 0.95 & 0.84 & 18.29 & 8.00 \\ 
5e-5 & 5 & 1e-4 & 85.21 & 0.99 & 0.97 & 0.90 & 17.20 & 3.24 \\ 
5e-5 & 5 & 1e-3 & 82.69 & 0.99 & 0.98 & 0.93 & 16.41 & 1.17 \\ 
5e-5 & 5 & 1e-2 & 80.91 & 0.99 & 0.97 & 0.93 & 15.35 & 0.45 \\ 
5e-5 & 1 & 1e-7 & 85.78 & 0.95 & 0.89 & 0.70 & 22.42 & 29.93 \\ 
5e-5 & 1 & 1e-6 & 86.31 & 0.96 & 0.92 & 0.77 & 21.76 & 16.98 \\ 
5e-5 & 1 & 1e-5 & 85.54 & 0.97 & 0.93 & 0.82 & 20.53 & 7.38 \\ 
5e-5 & 1 & 1e-4 & 84.62 & 0.98 & 0.95 & 0.88 & 18.94 & 2.99 \\ 
5e-5 & 1 & 1e-3 & 83.82 & 0.98 & 0.97 & 0.91 & 18.00 & 1.17 \\ 
5e-5 & 1 & 1e-2 & 80.49 & 0.98 & 0.97 & 0.92 & 17.10 & 0.43 \\ 
5e-5 & 5e-1 & 1e-7 & 84.84 & 0.95 & 0.89 & 0.70 & 23.02 & 26.39 \\ 
5e-5 & 5e-1 & 1e-6 & 85.21 & 0.96 & 0.91 & 0.75 & 20.38 & 16.81 \\ 
5e-5 & 5e-1 & 1e-5 & 82.98 & 0.96 & 0.92 & 0.81 & 12.51 & 6.07 \\ 
5e-5 & 5e-1 & 1e-4 & 84.29 & 0.97 & 0.95 & 0.87 & 15.54 & 3.02 \\ 
5e-5 & 5e-1 & 1e-3 & 82.67 & 0.98 & 0.96 & 0.89 & 14.95 & 1.12 \\ 
5e-5 & 5e-1 & 1e-2 & 80.35 & 0.98 & 0.96 & 0.89 & 11.56 & 0.42 \\ 
5e-4 & 10 & 1e-7 & 88.96 & 0.94 & 0.87 & 0.62 & 10.95 & 35.47 \\ 
5e-4 & 10 & 1e-6 & 88.24 & 0.96 & 0.91 & 0.73 & 10.76 & 17.68 \\ 
5e-4 & 10 & 1e-5 & 87.61 & 0.98 & 0.94 & 0.83 & 9.98 & 7.27 \\ 
5e-4 & 10 & 1e-4 & 86.54 & 0.98 & 0.96 & 0.88 & 9.20 & 3.08 \\ 
5e-4 & 10 & 1e-3 & 84.80 & 0.98 & 0.96 & 0.90 & 8.42 & 1.10 \\ 
5e-4 & 10 & 1e-2 & 82.72 & 0.98 & 0.96 & 0.90 & 7.67 & 0.41 \\ 
5e-4 & 5 & 1e-7 & 88.68 & 0.94 & 0.87 & 0.63 & 11.07 & 32.80 \\ 
5e-4 & 5 & 1e-6 & 88.31 & 0.96 & 0.91 & 0.73 & 10.78 & 16.68 \\ 
5e-4 & 5 & 1e-5 & 87.75 & 0.97 & 0.94 & 0.83 & 10.03 & 6.97 \\ 
5e-4 & 5 & 1e-4 & 86.35 & 0.98 & 0.96 & 0.88 & 9.70 & 2.96 \\ 
5e-4 & 5 & 1e-3 & 84.74 & 0.98 & 0.96 & 0.91 & 8.97 & 1.16 \\ 
5e-4 & 5 & 1e-2 & 82.12 & 0.98 & 0.96 & 0.91 & 8.04 & 0.44 \\ 
5e-4 & 1 & 1e-7 & 87.11 & 0.94 & 0.86 & 0.64 & 11.67 & 25.09 \\ 
5e-4 & 1 & 1e-6 & 87.08 & 0.96 & 0.91 & 0.75 & 11.34 & 14.06 \\ 
5e-4 & 1 & 1e-5 & 86.72 & 0.97 & 0.94 & 0.82 & 11.13 & 7.31 \\ 
5e-4 & 1 & 1e-4 & 85.64 & 0.98 & 0.96 & 0.88 & 9.81 & 2.92 \\ 
5e-4 & 1 & 1e-3 & 83.48 & 0.98 & 0.97 & 0.91 & 9.49 & 1.20 \\ 
5e-4 & 1 & 1e-2 & 81.87 & 0.98 & 0.96 & 0.91 & 8.11 & 0.44 \\ 
5e-4 & 5e-1 & 1e-7 & 78.10 & 0.93 & 0.87 & 0.73 & 7.07 & 4.57 \\ 
5e-4 & 5e-1 & 1e-6 & 85.76 & 0.95 & 0.90 & 0.75 & 11.03 & 12.23 \\ 
5e-4 & 5e-1 & 1e-5 & 85.56 & 0.97 & 0.93 & 0.82 & 10.82 & 6.20 \\ 
5e-4 & 5e-1 & 1e-4 & 85.21 & 0.98 & 0.96 & 0.88 & 10.21 & 2.86 \\ 
5e-4 & 5e-1 & 1e-3 & 81.05 & 0.96 & 0.93 & 0.84 & 6.99 & 0.94 \\ 
5e-4 & 5e-1 & 1e-2 & 81.98 & 0.98 & 0.97 & 0.91 & 8.12 & 0.46 \\ 
5e-3 & 10 & 1e-7 & 86.09 & 0.91 & 0.83 & 0.65 & 4.79 & 8.23 \\ 
5e-3 & 10 & 1e-6 & 85.92 & 0.91 & 0.83 & 0.66 & 4.74 & 6.31 \\ 
5e-3 & 10 & 1e-5 & 85.59 & 0.93 & 0.87 & 0.72 & 4.66 & 3.73 \\ 
5e-3 & 10 & 1e-4 & 84.53 & 0.95 & 0.91 & 0.79 & 4.50 & 1.88 \\ 
5e-3 & 10 & 1e-3 & 83.11 & 0.96 & 0.93 & 0.84 & 4.30 & 0.86 \\ 
5e-3 & 10 & 1e-2 & 80.16 & 0.97 & 0.95 & 0.88 & 4.00 & 0.33 \\ 
5e-3 & 5 & 1e-7 & 85.90 & 0.90 & 0.82 & 0.64 & 4.78 & 7.09 \\ 
5e-3 & 5 & 1e-6 & 85.43 & 0.91 & 0.84 & 0.67 & 4.79 & 5.57 \\ 
5e-3 & 5 & 1e-5 & 85.26 & 0.92 & 0.86 & 0.71 & 4.67 & 3.48 \\ 
5e-3 & 5 & 1e-4 & 84.51 & 0.95 & 0.90 & 0.78 & 4.52 & 1.76 \\ 
5e-3 & 5 & 1e-3 & 83.07 & 0.96 & 0.93 & 0.84 & 4.32 & 0.82 \\ 
5e-3 & 5 & 1e-2 & 80.53 & 0.97 & 0.94 & 0.87 & 4.06 & 0.32 \\ 
5e-3 & 1 & 1e-7 & 82.84 & 0.91 & 0.85 & 0.70 & 4.76 & 4.11 \\ 
5e-3 & 1 & 1e-6 & 83.40 & 0.92 & 0.86 & 0.72 & 4.85 & 3.37 \\ 
5e-3 & 1 & 1e-5 & 81.72 & 0.93 & 0.88 & 0.75 & 4.65 & 2.62 \\ 
5e-3 & 1 & 1e-4 & 82.19 & 0.95 & 0.91 & 0.81 & 4.68 & 1.52 \\ 
5e-3 & 1 & 1e-3 & 81.14 & 0.96 & 0.93 & 0.86 & 4.43 & 0.74 \\ 
5e-3 & 1 & 1e-2 & 79.13 & 0.96 & 0.94 & 0.86 & 4.16 & 0.32 \\ 
5e-3 & 5e-1 & 1e-7 & 77.41 & 0.96 & 0.93 & 0.85 & 4.59 & 2.03 \\ 
5e-3 & 5e-1 & 1e-6 & 77.42 & 0.97 & 0.94 & 0.85 & 4.57 & 1.91 \\ 
5e-3 & 5e-1 & 1e-5 & 76.88 & 0.97 & 0.94 & 0.86 & 4.47 & 1.57 \\ 
5e-3 & 5e-1 & 1e-4 & 77.17 & 0.97 & 0.94 & 0.86 & 4.49 & 1.24 \\ 
5e-3 & 5e-1 & 1e-3 & 76.68 & 0.97 & 0.95 & 0.89 & 4.29 & 0.64 \\ 
5e-3 & 5e-1 & 1e-2 & 75.01 & 0.98 & 0.97 & 0.92 & 3.89 & 0.27 \\ 
1e-2 & 10 & 1e-7 & 64.43 & 0.94 & 0.87 & 0.65 & 3.37 & 7.11 \\ 
1e-2 & 10 & 1e-6 & 79.77 & 0.90 & 0.83 & 0.66 & 3.46 & 4.06 \\ 
1e-2 & 10 & 1e-5 & 79.87 & 0.91 & 0.84 & 0.69 & 3.46 & 2.47 \\ 
1e-2 & 10 & 1e-4 & 78.86 & 0.94 & 0.88 & 0.75 & 3.37 & 1.34 \\ 
1e-2 & 10 & 1e-3 & 77.68 & 0.95 & 0.92 & 0.82 & 3.25 & 0.63 \\ 
1e-2 & 10 & 1e-2 & 75.75 & 0.97 & 0.94 & 0.87 & 3.12 & 0.25 \\ 
1e-2 & 5 & 1e-7 & 78.63 & 0.89 & 0.81 & 0.65 & 3.40 & 4.42 \\ 
1e-2 & 5 & 1e-6 & 77.74 & 0.90 & 0.83 & 0.66 & 3.34 & 4.21 \\ 
1e-2 & 5 & 1e-5 & 78.16 & 0.90 & 0.83 & 0.68 & 3.35 & 2.34 \\ 
1e-2 & 5 & 1e-4 & 78.18 & 0.92 & 0.86 & 0.73 & 3.31 & 1.20 \\ 
1e-2 & 5 & 1e-3 & 77.43 & 0.96 & 0.92 & 0.82 & 3.25 & 0.59 \\ 
1e-2 & 5 & 1e-2 & 74.45 & 0.96 & 0.93 & 0.85 & 3.02 & 0.25 \\ 
1e-2 & 1 & 1e-7 & 71.74 & 0.97 & 0.94 & 0.86 & 3.21 & 1.26 \\ 
1e-2 & 1 & 1e-6 & 71.92 & 0.97 & 0.95 & 0.88 & 3.24 & 1.18 \\ 
1e-2 & 1 & 1e-5 & 73.02 & 0.97 & 0.94 & 0.86 & 3.30 & 0.98 \\ 
1e-2 & 1 & 1e-4 & 72.02 & 0.97 & 0.95 & 0.88 & 3.16 & 0.77 \\ 
1e-2 & 1 & 1e-3 & 71.16 & 0.98 & 0.95 & 0.89 & 3.08 & 0.43 \\ 
1e-2 & 1 & 1e-2 & 69.64 & 0.98 & 0.97 & 0.92 & 2.90 & 0.19 \\ 
1e-2 & 5e-1 & 1e-7 & 70.53 & 0.98 & 0.96 & 0.90 & 3.20 & 0.87 \\ 
1e-2 & 5e-1 & 1e-6 & 70.07 & 0.98 & 0.96 & 0.90 & 3.21 & 0.89 \\ 
1e-2 & 5e-1 & 1e-5 & 70.46 & 0.98 & 0.96 & 0.89 & 3.20 & 0.79 \\ 
1e-2 & 5e-1 & 1e-4 & 71.49 & 0.98 & 0.96 & 0.91 & 3.31 & 0.72 \\ 
1e-2 & 5e-1 & 1e-3 & 70.06 & 0.99 & 0.97 & 0.93 & 3.11 & 0.41 \\ 
1e-2 & 5e-1 & 1e-2 & 67.76 & 0.99 & 0.98 & 0.94 & 2.84 & 0.19 \\ 
 \hline 
\caption{Statistics of all network configurations. Columns show weight decay, activation function (ReLU or $\beta$ parameter for Softplus), parameter for curvature minimization $\zeta$, test accuracy (acc), mean Pearson correlation coefficient (pcc) for Gaussian noise with different noise levels $\nu$, average weight norm ($\norm{W}$) and average approximated Hessian norm ($\norm{H}$).}
\label{tab:statiscticsAllNets}

\end{longtable}
\subsection{Gradient Explanation}
\begin{figure}[!htb]
\minipage{0.32\textwidth}
  \includegraphics[width=1\linewidth]{Images/robustness/cifar_cnn/gradient/cifar_cnn_pcc_wd.pdf}
\endminipage\hfill
\minipage{0.32\textwidth}
  \includegraphics[width=1\linewidth]{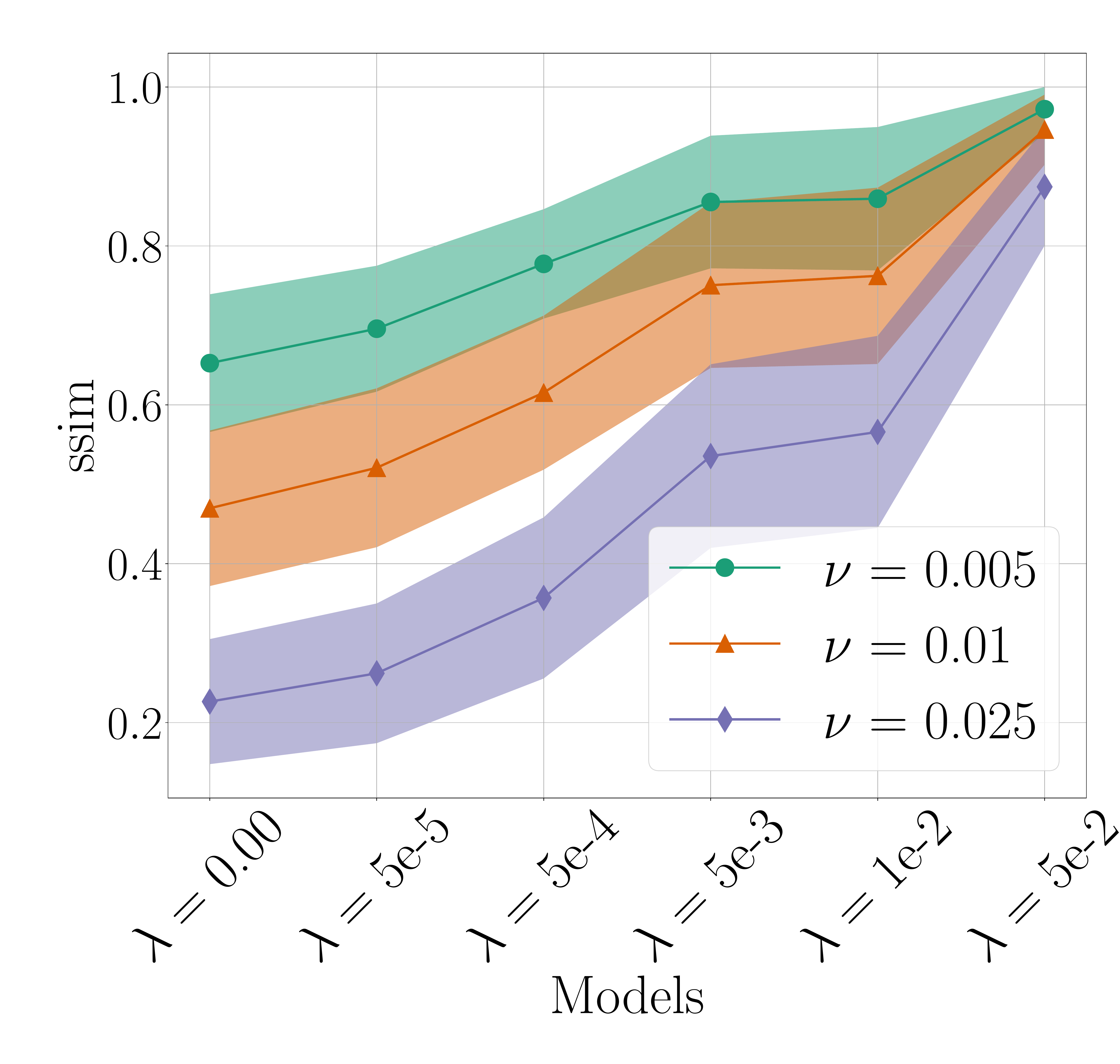}
\endminipage\hfill
\minipage{0.32\textwidth}%
  \includegraphics[width=1\linewidth]{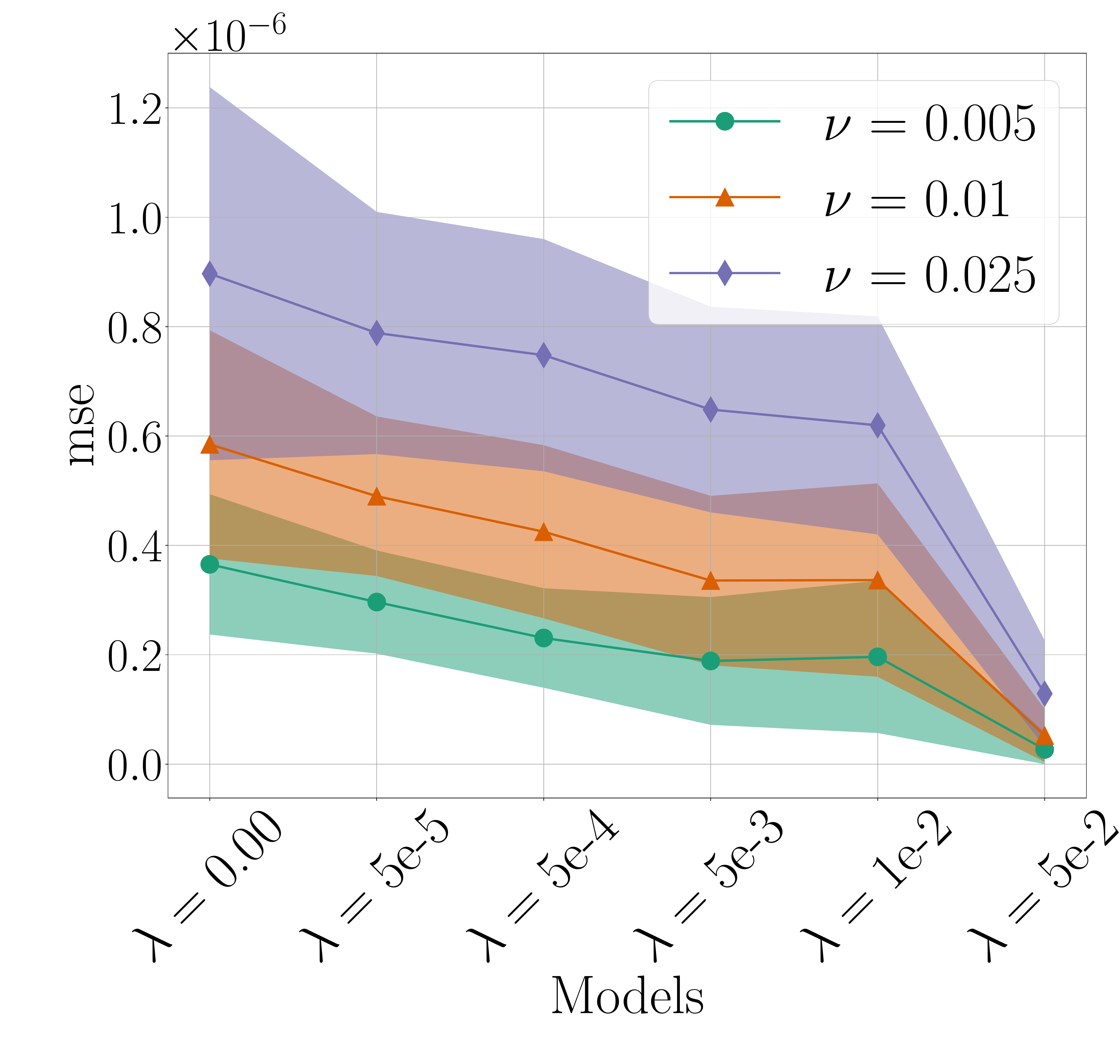}
\endminipage
  \caption{PCC, SSIM and MSE between original Gradient explanation map and explanation after adding random noise to the image. PCC and SSIM are higher and MSE is lower for networks trained with weight decay. That means weight decay improves robustness of explanations. We show mean +/- std.}
  \label{appndx:fig:wd_gradient}
\end{figure}

\begin{figure}[!htb]
\minipage{0.32\textwidth}
  \includegraphics[width=1\linewidth]{Images/robustness/cifar_cnn/gradient/cifar_cnn_pcc_sp_wd_5e-4.pdf}
\endminipage\hfill
\minipage{0.32\textwidth}
  \includegraphics[width=1\linewidth]{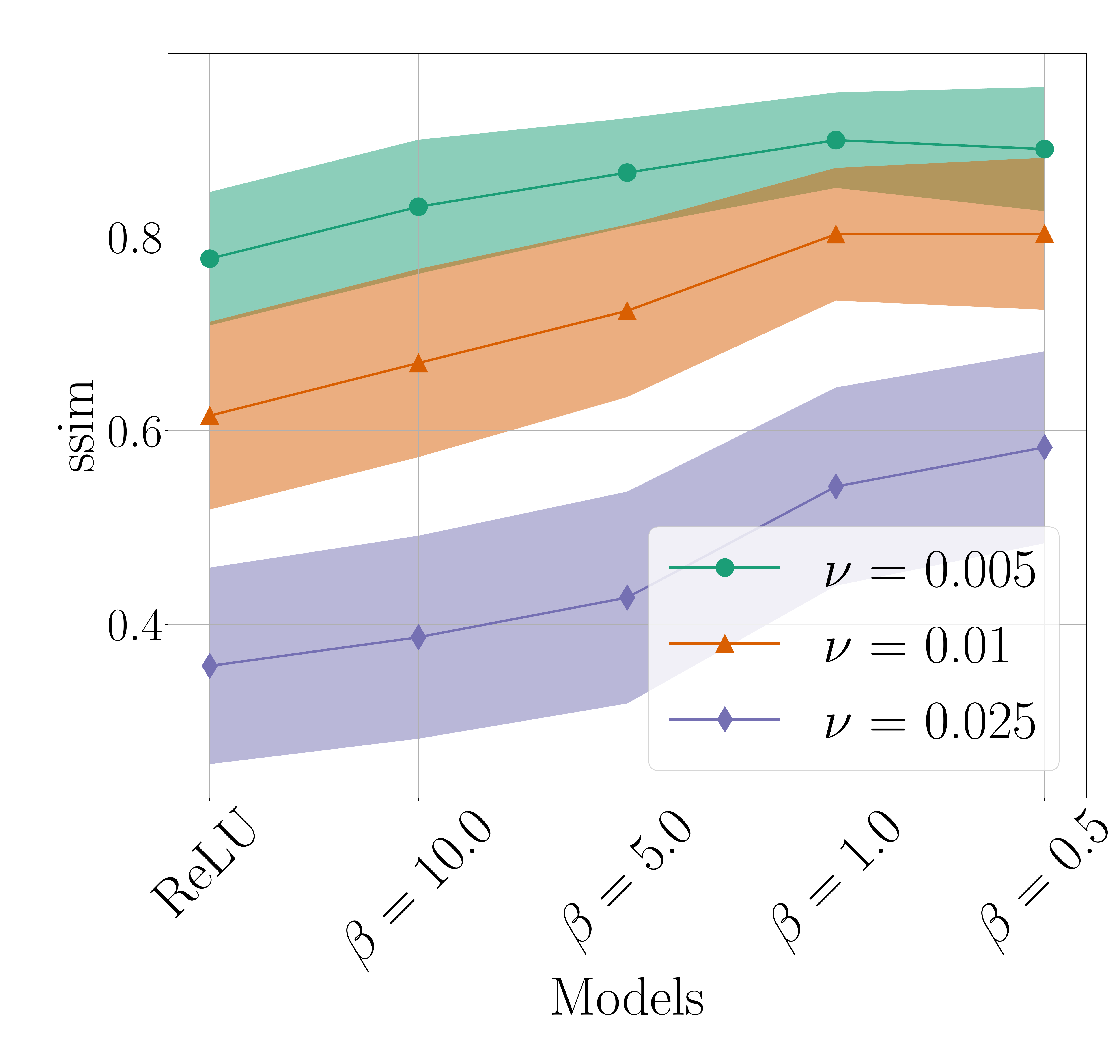}
\endminipage\hfill
\minipage{0.32\textwidth}%
  \includegraphics[width=1\linewidth]{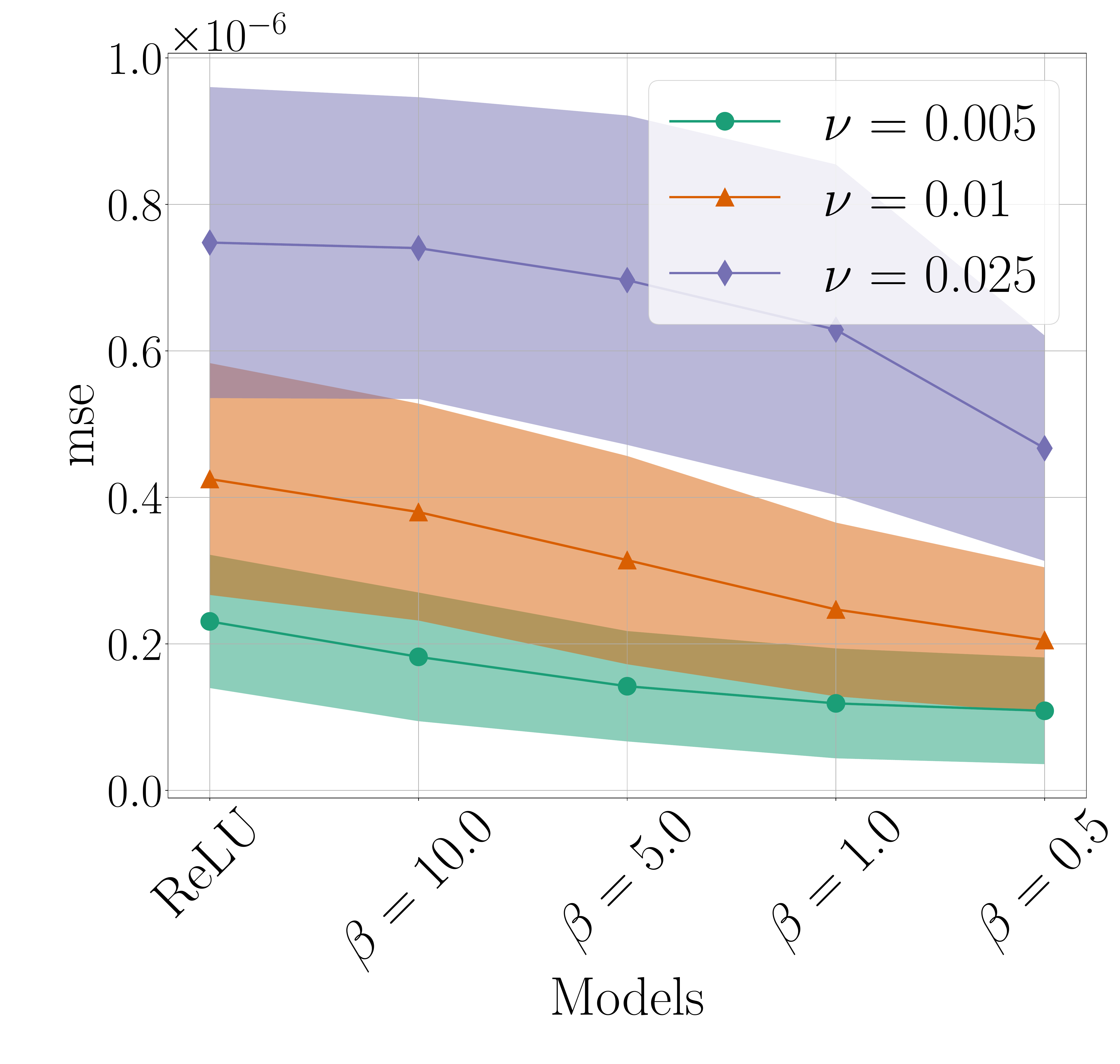}
\endminipage
  \caption{PCC, SSIM and MSE between original Gradient explanation map and explanation after adding random noise to the image. PCC and SSIM are higher and MSE is lower for Softplus networks trained with a small $\beta$ value. That means Softplus activations improve robustness of explanations. All nets were trained with weight decay ($\lambda$=5e-4). We show mean +/- std.}
  \label{appndx:fig:sp_gradient}
\end{figure}

\begin{figure}[!htp]
\minipage{0.32\textwidth}
  \includegraphics[width=1\linewidth]{Images/robustness/cifar_cnn/gradient/cifar_cnn_pcc_curv_sp_10_wd_5e-4.pdf}
\endminipage\hfill
\minipage{0.32\textwidth}
  \includegraphics[width=1\linewidth]{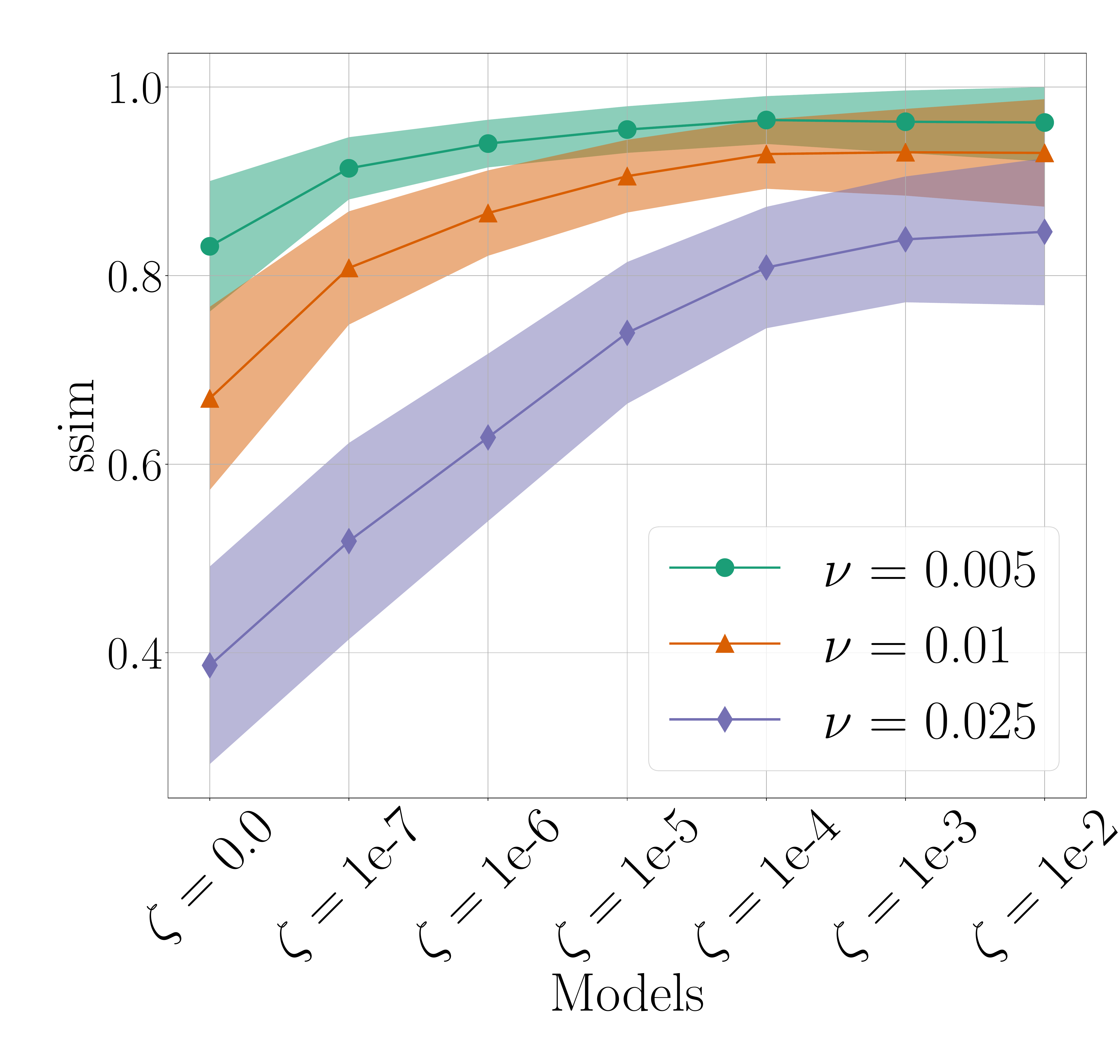}
\endminipage\hfill
\minipage{0.32\textwidth}%
  \includegraphics[width=1\linewidth]{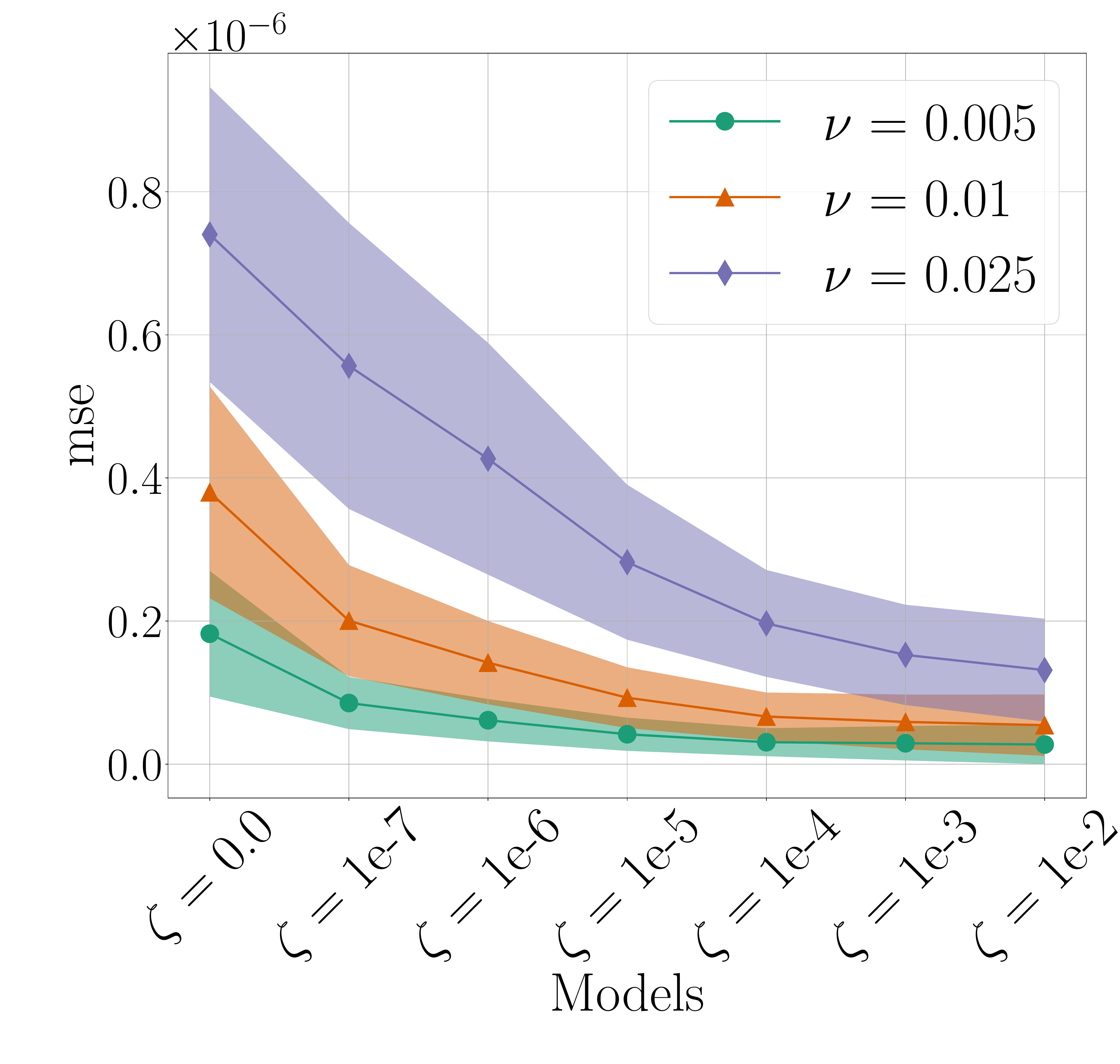}
\endminipage
  \caption{PCC, SSIM and MSE between original Gradient explanation map and explanation after adding random noise to the image. PCC and SSIM are higher and MSE is lower for networks trained with strong curvature minimization. That means minimizing the curvature improves robustness of explanations. All nets were trained with Softplus activations ($\beta=10$) and weight decay ($\lambda$=5e-4). We show mean +/- std.}
  \label{appndx:fig:curv_min_gradient}
\end{figure}

    In Figures~\ref{appndx:fig:wd_gradient},~\ref{appndx:fig:sp_gradient}, and~\ref{appndx:fig:curv_min_gradient}, we show additional error measures for the Gradient explanation. PCC and SSIM increase with robustness while MSE decreases.

\subsection{Other Explanation Methods}\label{app:otherexplanations}

\begin{figure}[!htb]
\minipage{0.32\textwidth}
  \includegraphics[width=1\linewidth]{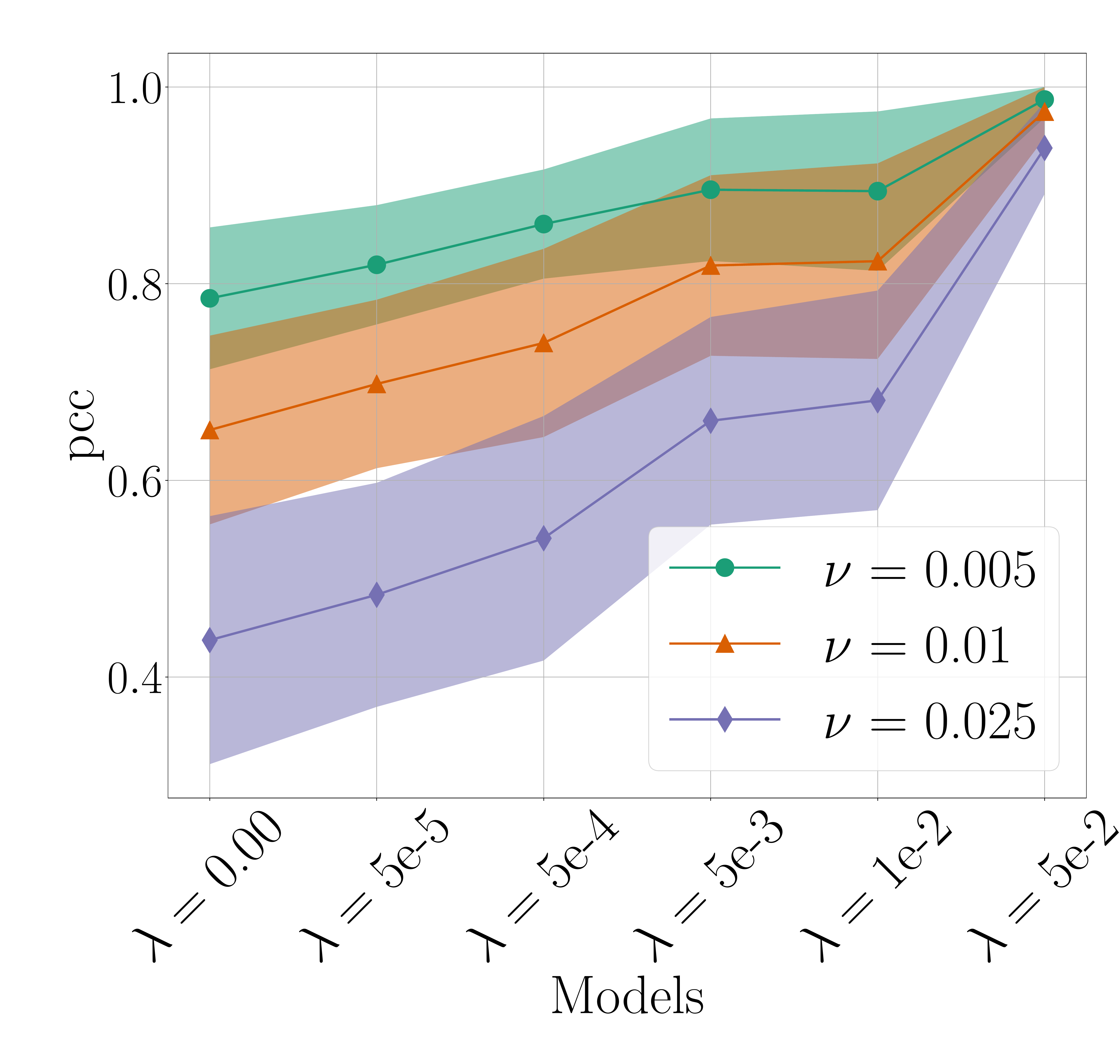}
\endminipage\hfill
\minipage{0.32\textwidth}
  \includegraphics[width=1\linewidth]{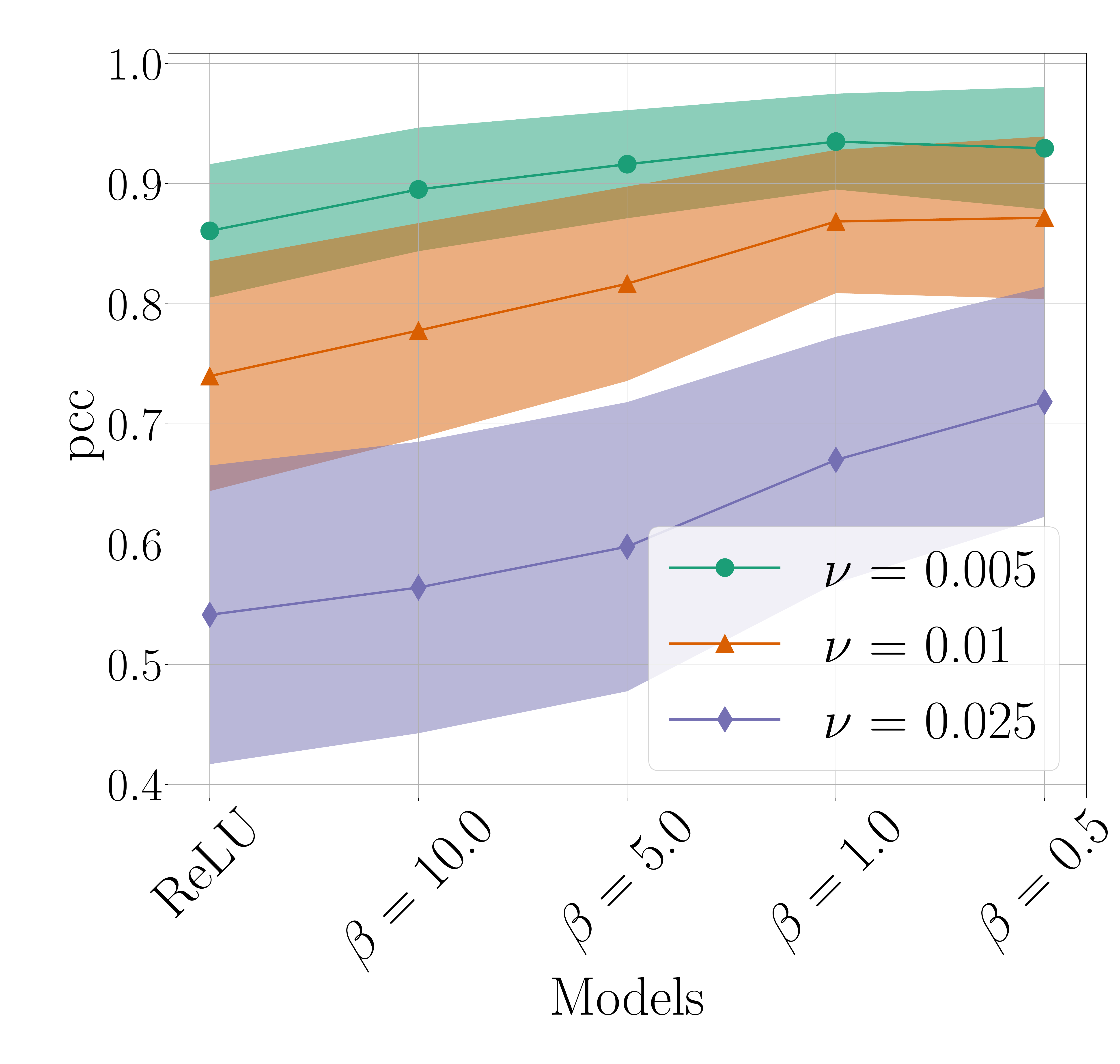}
\endminipage\hfill
\minipage{0.32\textwidth}%
  \includegraphics[width=1\linewidth]{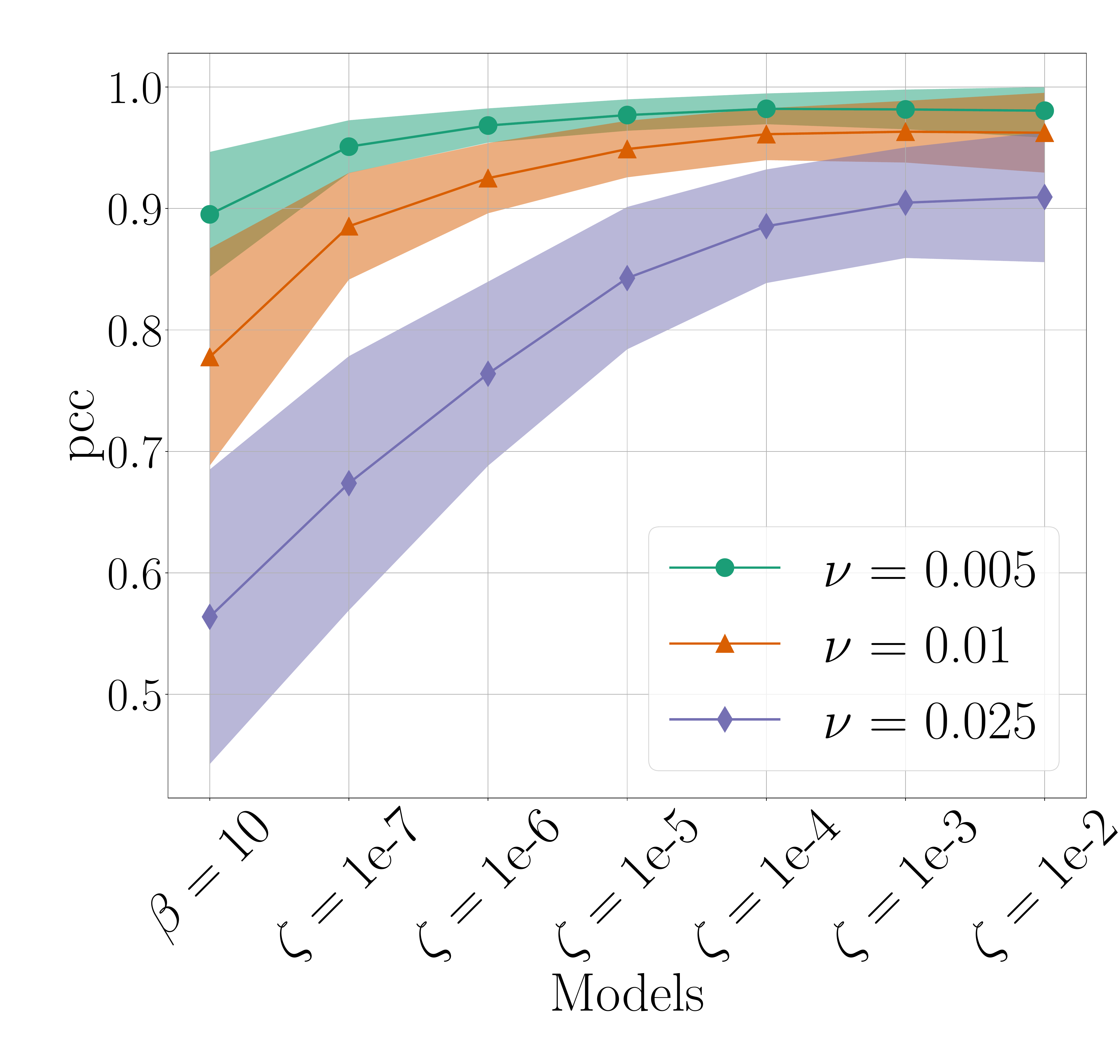}
\endminipage
  \caption{PCCs (mean +/- std) between original Gradient$\times$Input explanation map and explanation after adding random noise to the image. left: effect of weight decay, middle: effect of Softplus $\beta$ (for $\lambda=5$e-4), right: effect of curvature minimization (for $\lambda=5$e-4, $\beta$=10).}
  \label{appndx:fig:wd_sp_curv_min_gradient_times_input}
\end{figure}

\begin{figure}[!htb]
\minipage{0.32\textwidth}
  \includegraphics[width=1\linewidth]{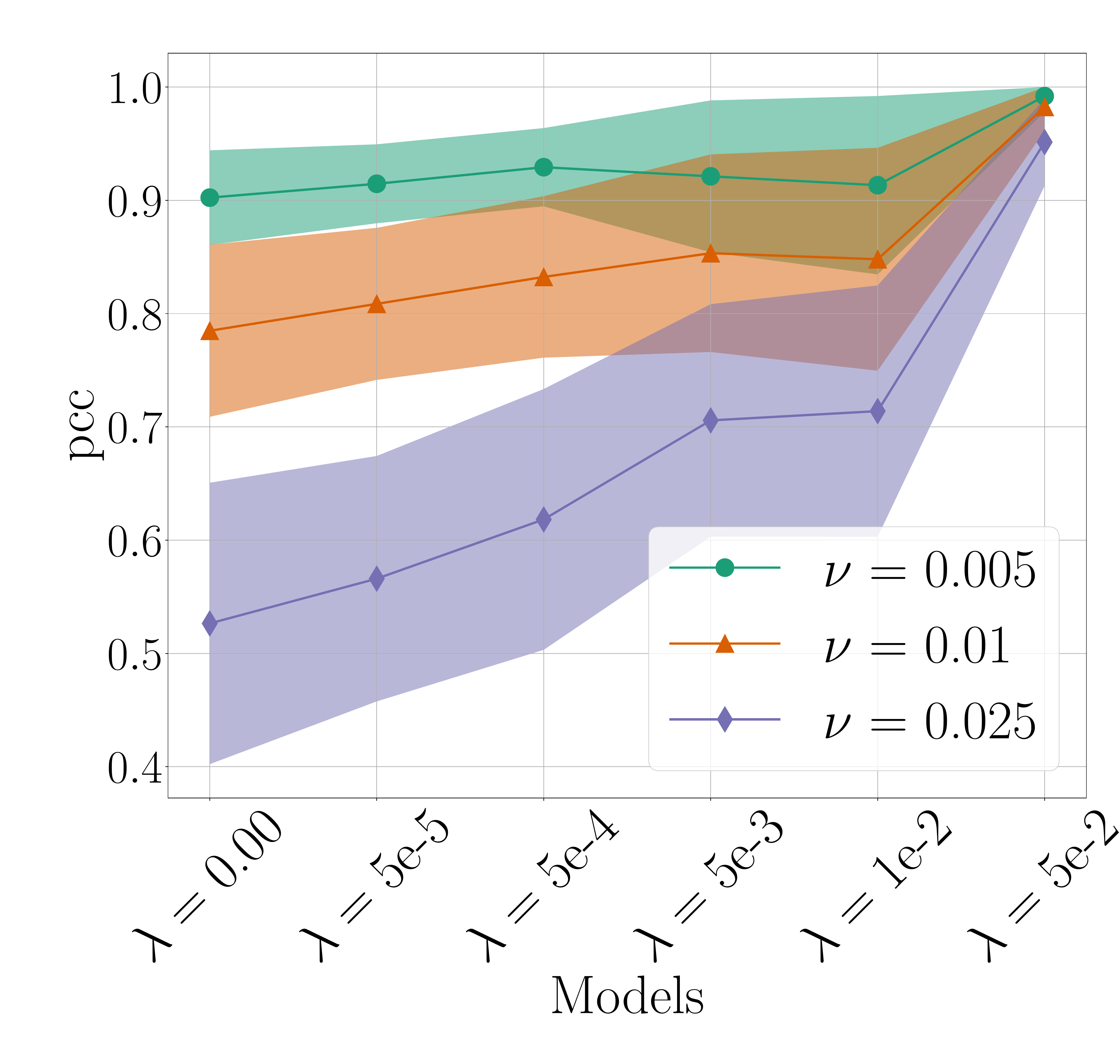}
\endminipage\hfill
\minipage{0.32\textwidth}
  \includegraphics[width=1\linewidth]{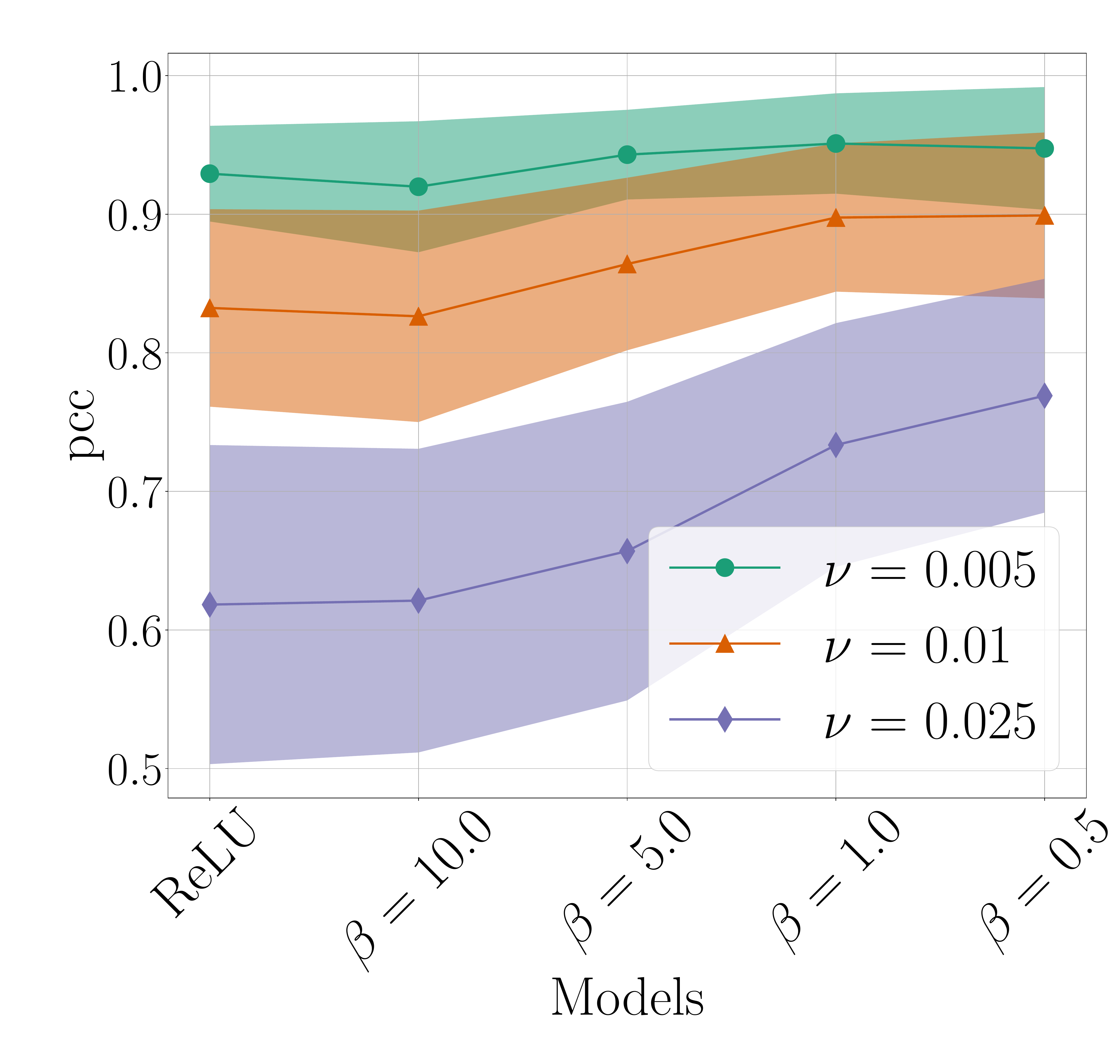}
\endminipage\hfill
\minipage{0.32\textwidth}%
  \includegraphics[width=1\linewidth]{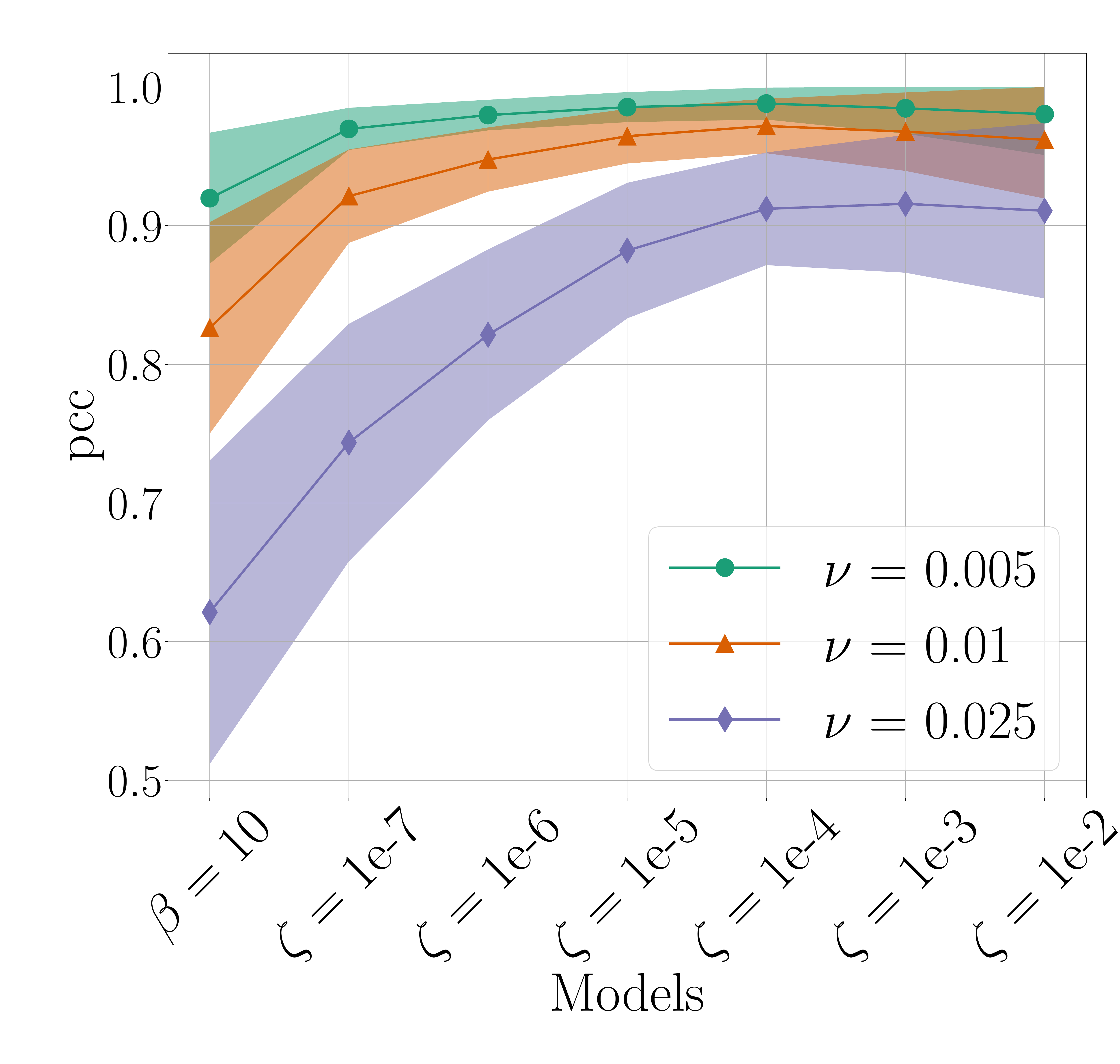}
\endminipage
  \caption{PCCs (mean +/- std) between original Integrated Gradients explanation map and explanation after adding random noise to the image. left: effect of weight decay, middle: effect of Softplus $\beta$ (for $\lambda=5$e-4), right: effect of curvature minimization (for $\lambda=5$e-4, $\beta$=10).}
  \label{appndx:fig:wd_sp_curv_min_integrated_gradients}
\end{figure}

\begin{figure}[!htb]
\minipage{0.32\textwidth}
  \includegraphics[width=1\linewidth]{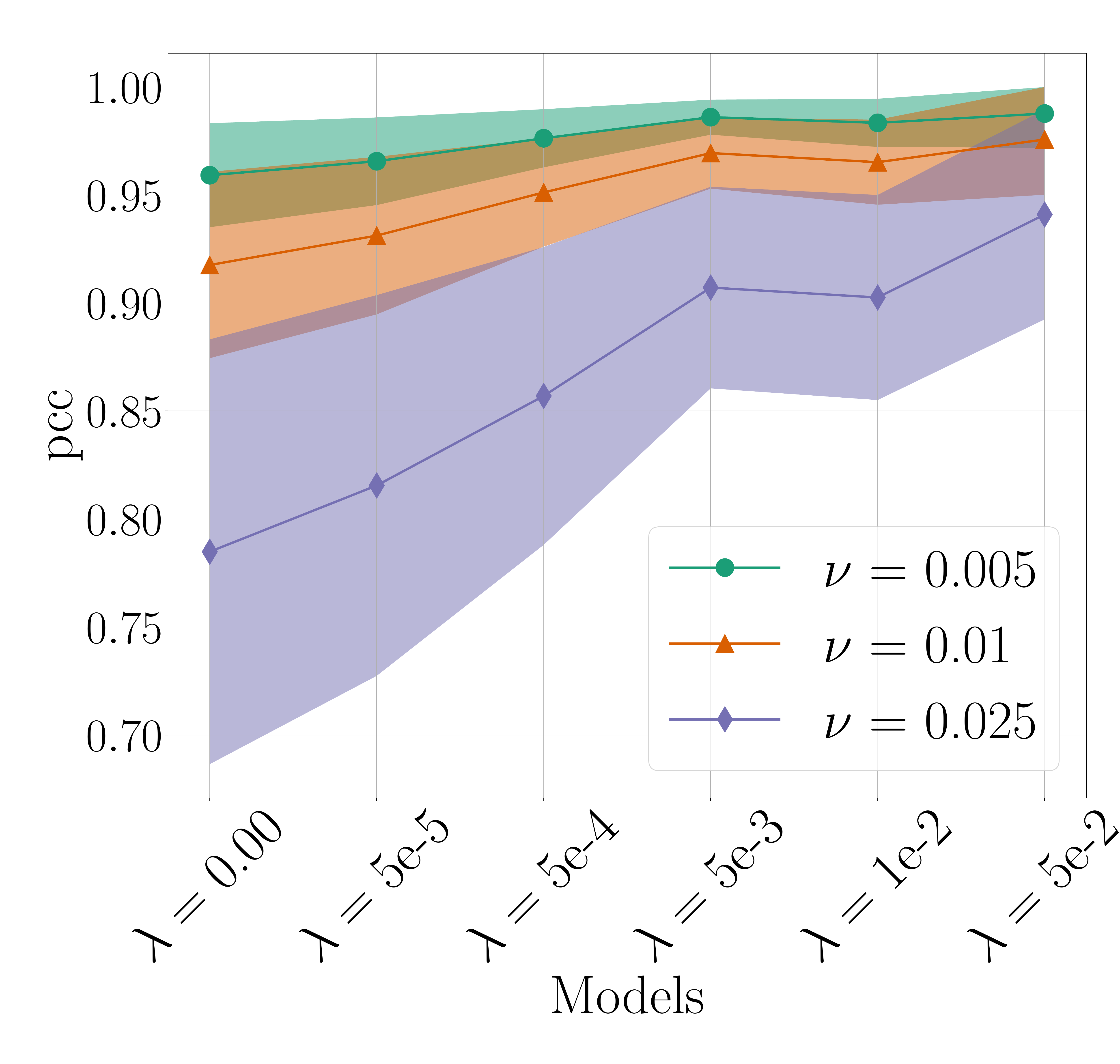}
\endminipage\hfill
\minipage{0.32\textwidth}
  \includegraphics[width=1\linewidth]{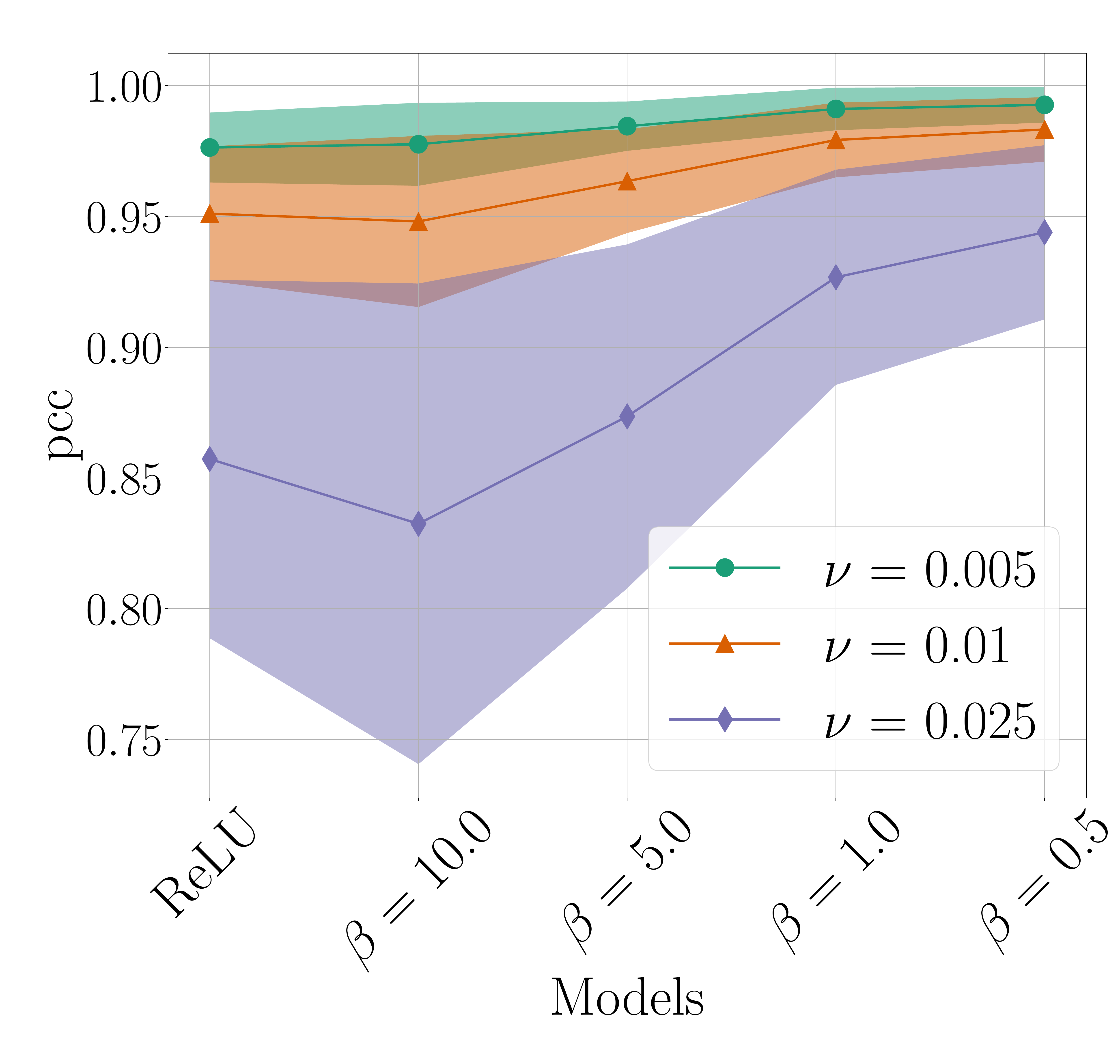}
\endminipage\hfill
\minipage{0.32\textwidth}%
  \includegraphics[width=1\linewidth]{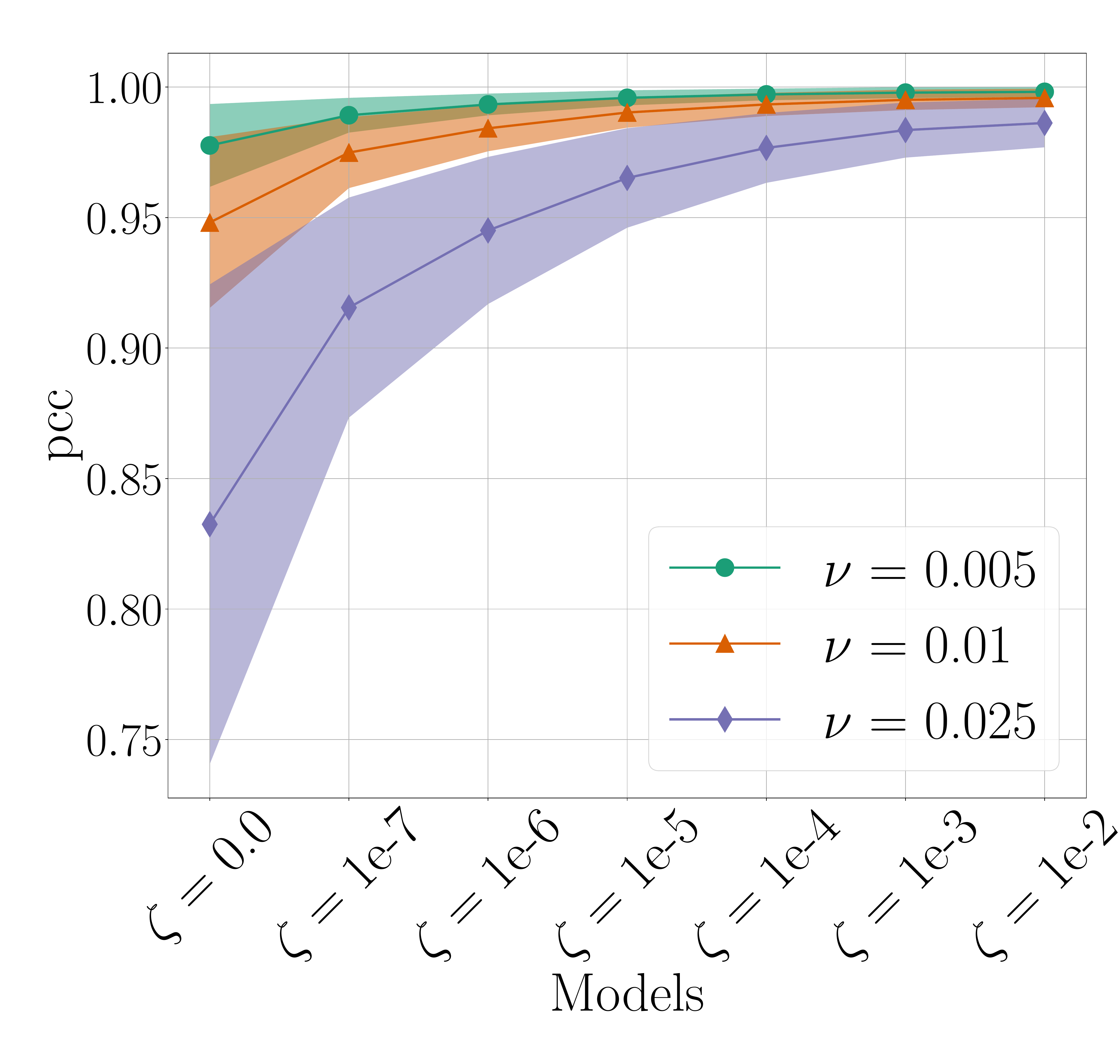}
\endminipage
  \caption{PCCs (mean +/- std) between original Guided Backpropagation explanation map and explanation after adding random noise to the image. left: effect of weight decay, middle: effect of Softplus $\beta$ (for $\lambda=5$e-4), right: effect of curvature minimization (for $\lambda=5$e-4, $\beta$=10).}
  \label{appndx:fig:wd_sp_curv_min_gbp}
\end{figure}

\begin{figure}[!htb]
\minipage{0.32\textwidth}
  \includegraphics[width=1\linewidth]{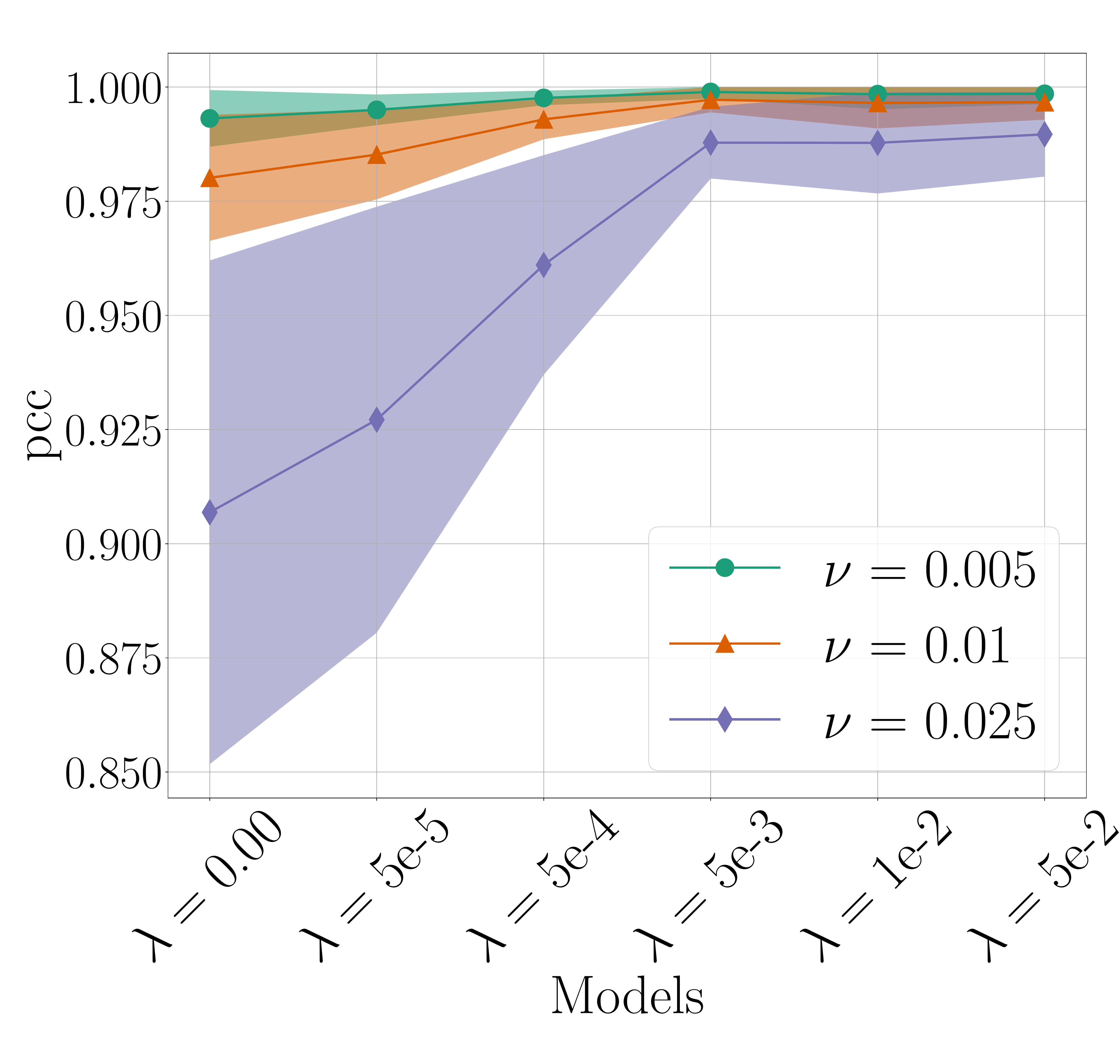}
\endminipage\hfill
\minipage{0.32\textwidth}
  \includegraphics[width=1\linewidth]{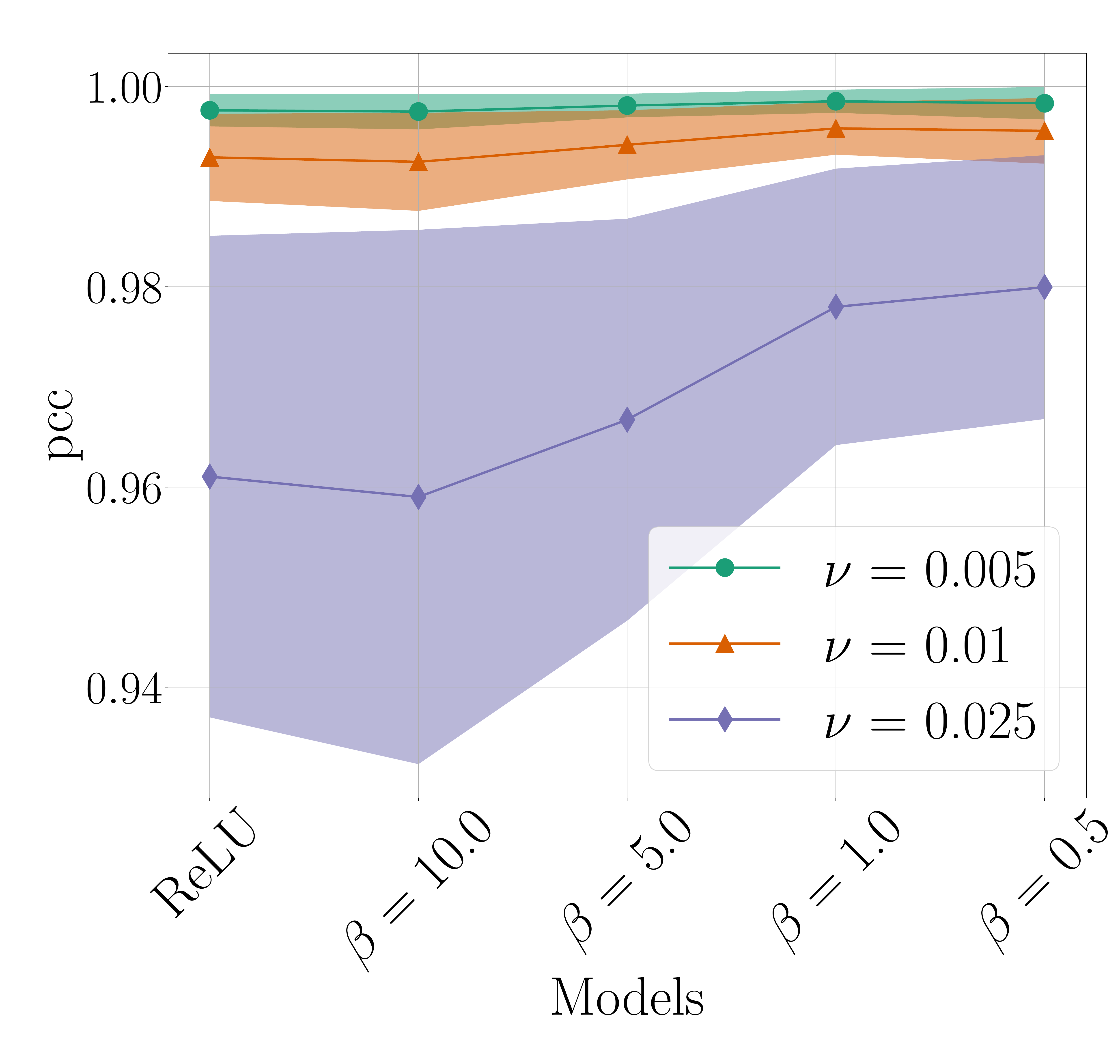}
\endminipage\hfill
\minipage{0.32\textwidth}%
  \includegraphics[width=1\linewidth]{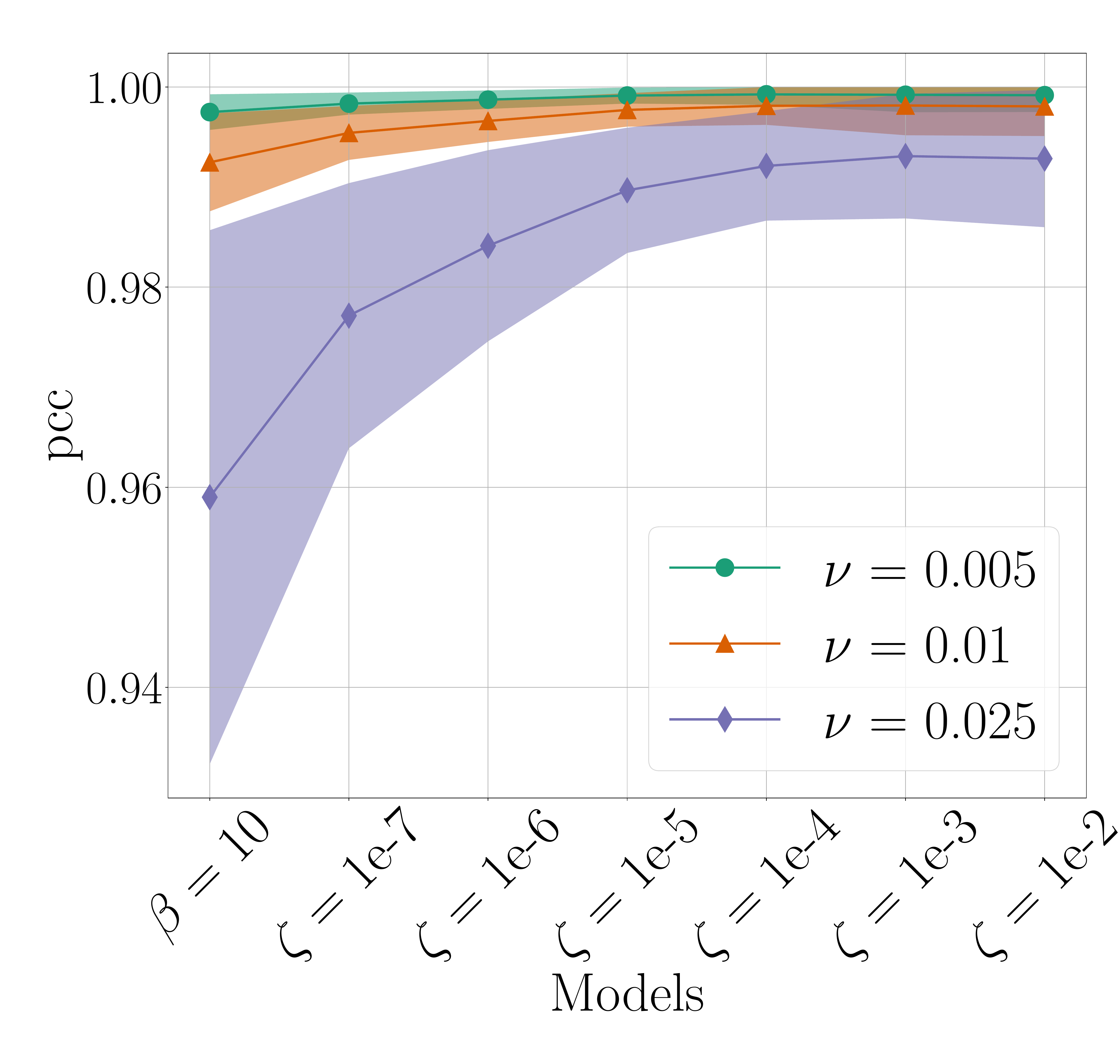}
\endminipage
  \caption{PCCs (mean +/- std) between original Layerwise Relevance Propagation explanation map and explanation after adding random noise to the image. left: effect of weight decay, middle: effect of Softplus $\beta$ (for $\lambda=5$e-4), right: effect of curvature minimization (for $\lambda=5$e-4, $\beta$=10).}
  \label{appndx:fig:wd_sp_curv_min_lrp}
\end{figure}

In Figures~\ref{appndx:fig:wd_sp_curv_min_gradient_times_input},~\ref{appndx:fig:wd_sp_curv_min_integrated_gradients},~\ref{appndx:fig:wd_sp_curv_min_gbp}, and~\ref{appndx:fig:wd_sp_curv_min_lrp}, we show how our proposed measures effect other explanation methods. The trend towards increased robustness is clearly visible for all considered explanation methods.
We note that the explanations start from different levels of robustness but can still profit from our methods. The most resilient method against random input perturbations is Layerwise Relevance Propagation, followed by Guided Backpropagation, Integrated Gradients, Gradient$\times$Input and Gradient in descending order.

\subsection{Other types of noise}\label{app:othernoise}

In the main text we only consider Gaussian noise. We repeat our experiments from~\ref{sec:ExperimentalAnalysis} for the Gradient explanation when we perturb the input images with Laplacian noise and salt-pepper noise.

\subsubsection{Laplace noise}
\begin{figure}[!htb]
\minipage{0.32\textwidth}
  \includegraphics[width=1\linewidth]{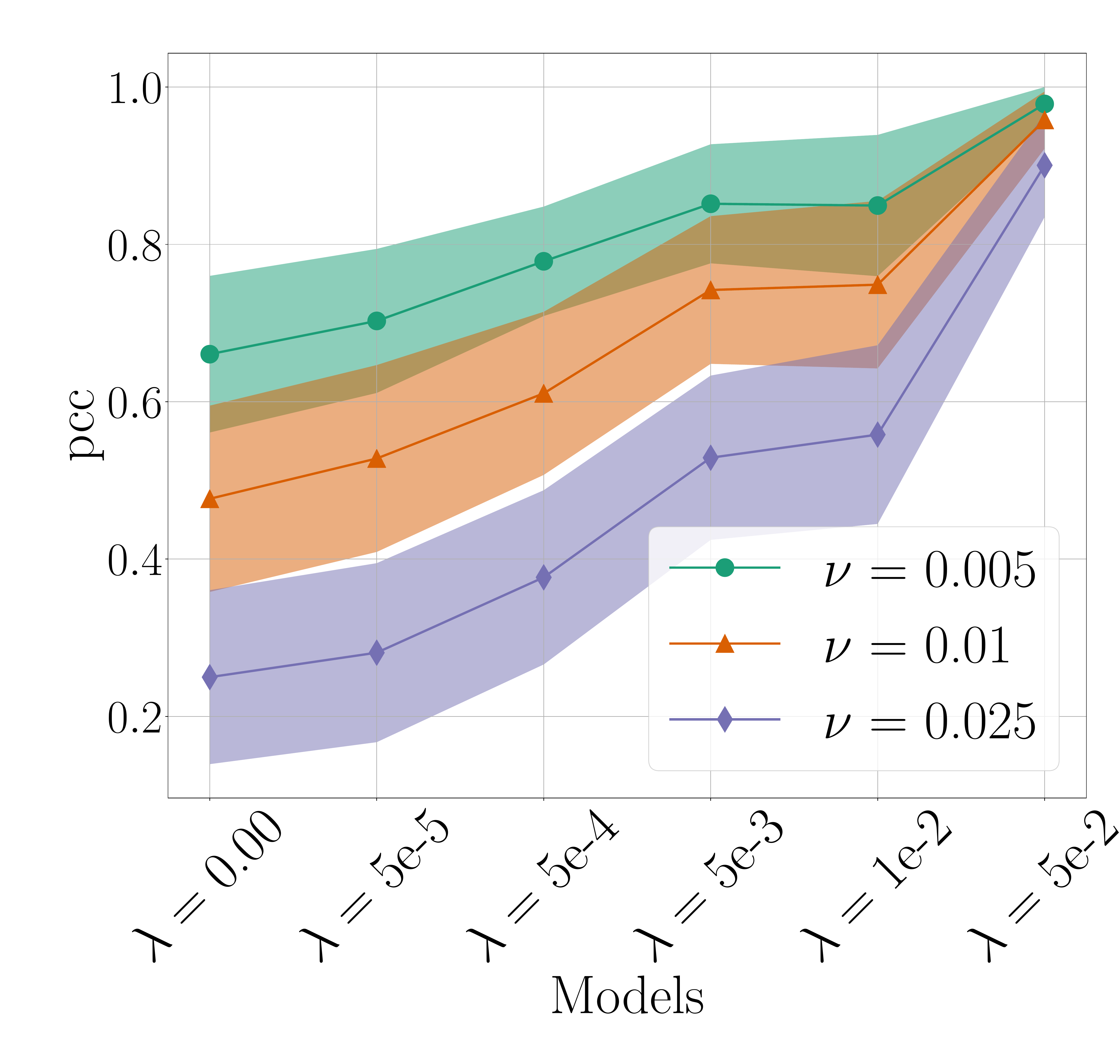}
\endminipage\hfill
\minipage{0.32\textwidth}
  \includegraphics[width=1\linewidth]{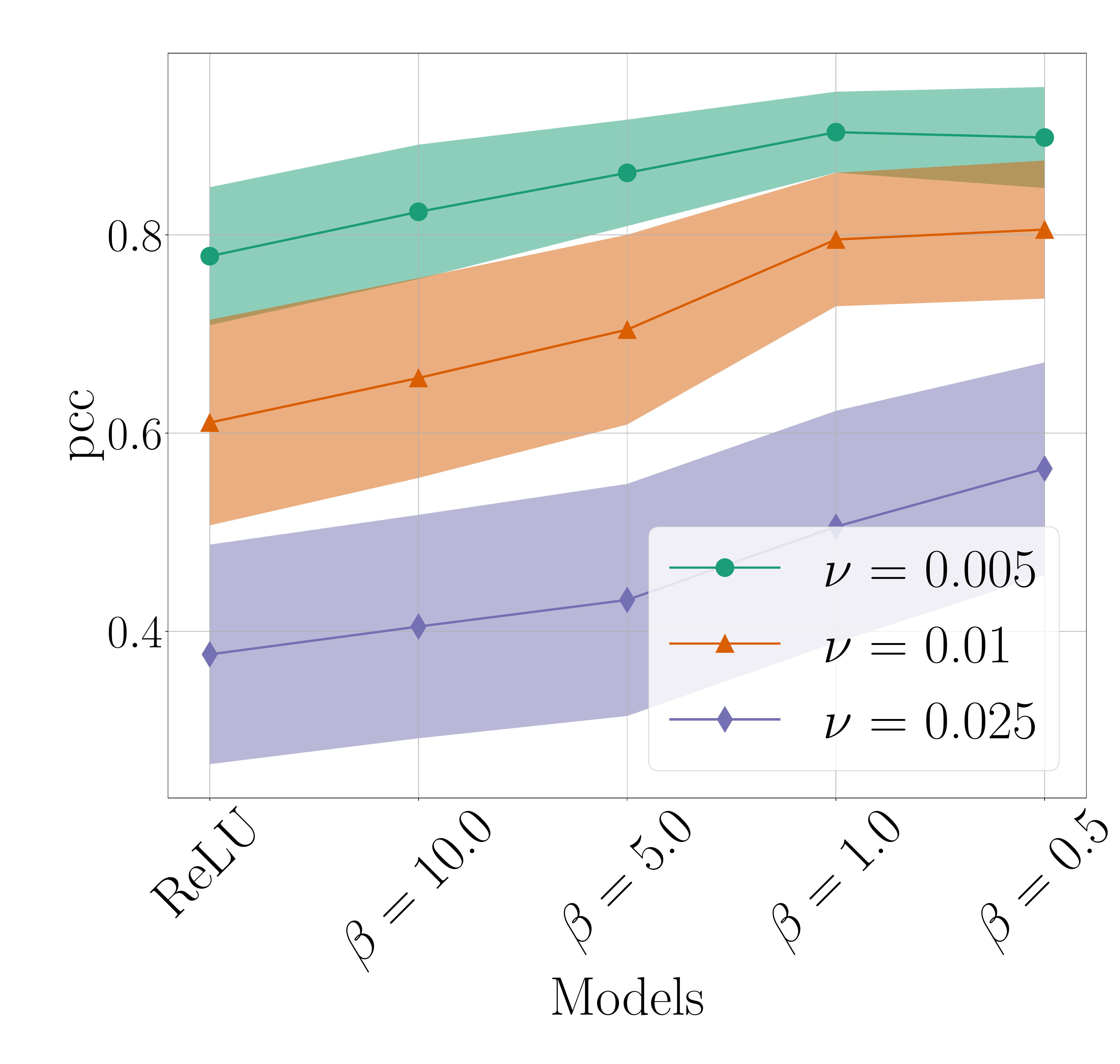}
\endminipage\hfill
\minipage{0.32\textwidth}%
  \includegraphics[width=1\linewidth]{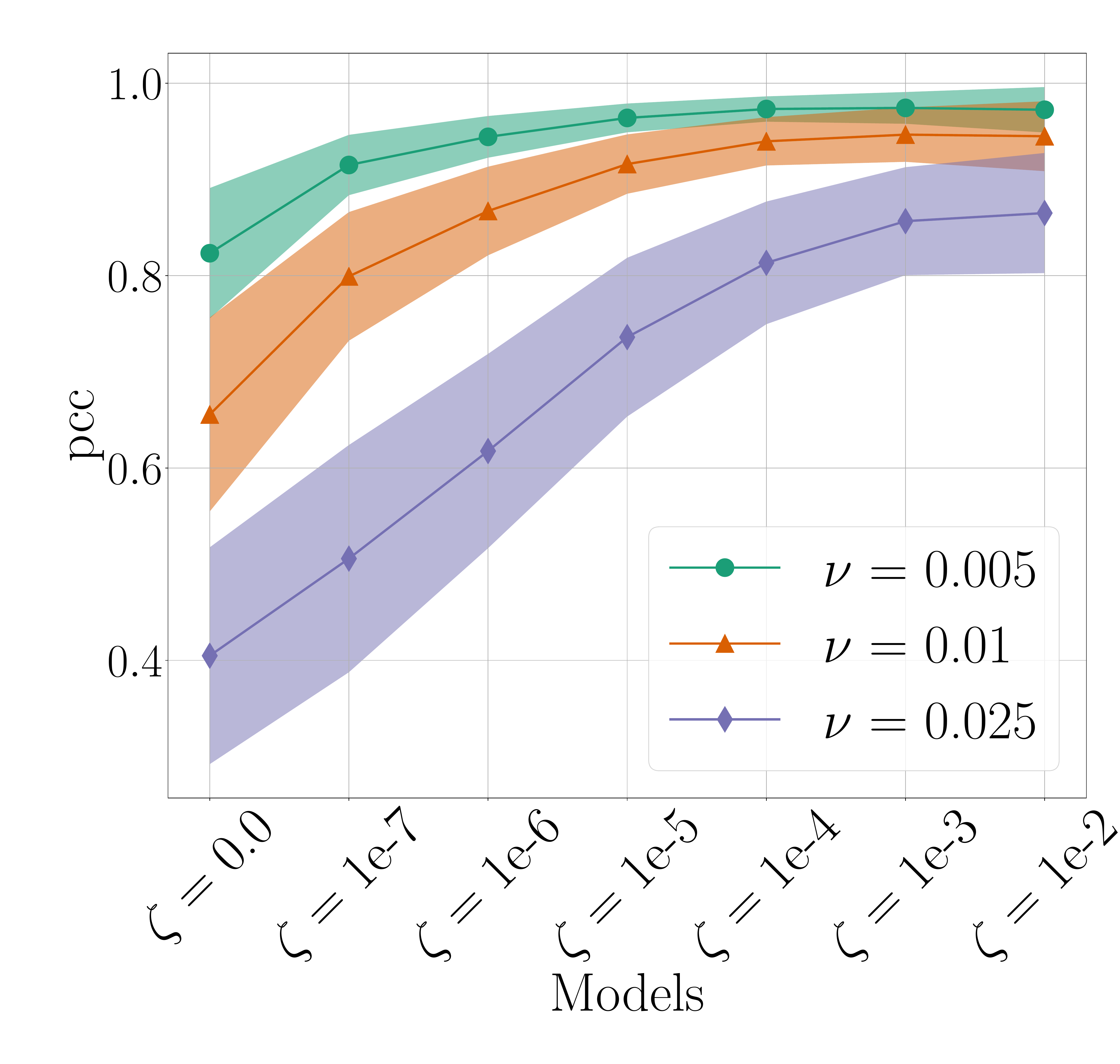}
\endminipage
  \caption{PCCs (mean +/- std) between original Gradient explanation map and explanation after adding Laplace noise to the image. left: effect of weight decay, middle: effect of Softplus $\beta$ (for $\lambda=5$e-4), right: effect of curvature minimization (for $\lambda=5$e-4, $\beta$=10).}
  \label{appndx:fig:wd_sp_curv_min_gradient_laplace}
\end{figure}

We sample random noise from the Laplace distribution

\begin{equation}
    f(x|\mu, b) = \frac{1}{2b}\exp{-\frac{|x-\mu|}{b}}
\end{equation}

where $\mu$ is the data mean and $b$ is a scale parameter which we define as $b= (x_{max} - x_{min}) \nu $, depending on the noise level $\nu$.
Figure~\ref{appndx:fig:wd_sp_curv_min_gradient_laplace} shows effects on the Gradient explanation when adding Laplace noise to the input images. We see that the results look statistically very similar to the results for Gaussian noise.

\subsubsection{Salt-pepper noise}
To perturb an image with salt-pepper noise we randomly select $100*\frac{\nu}{2}$~\% of the pixel in the image and switch them to $x_{max}$ (white) or $x_{min}$ (black) at random.
We select a very small amount of pixel (for noise level $\nu=0.005$ only 3 pixels) to be perturbed as salt-pepper noise has a very strong effect on the classification accuracy which we aim to keep approximately constant.

\begin{figure}[!htb]
\minipage{0.32\textwidth}
  \includegraphics[width=1\linewidth]{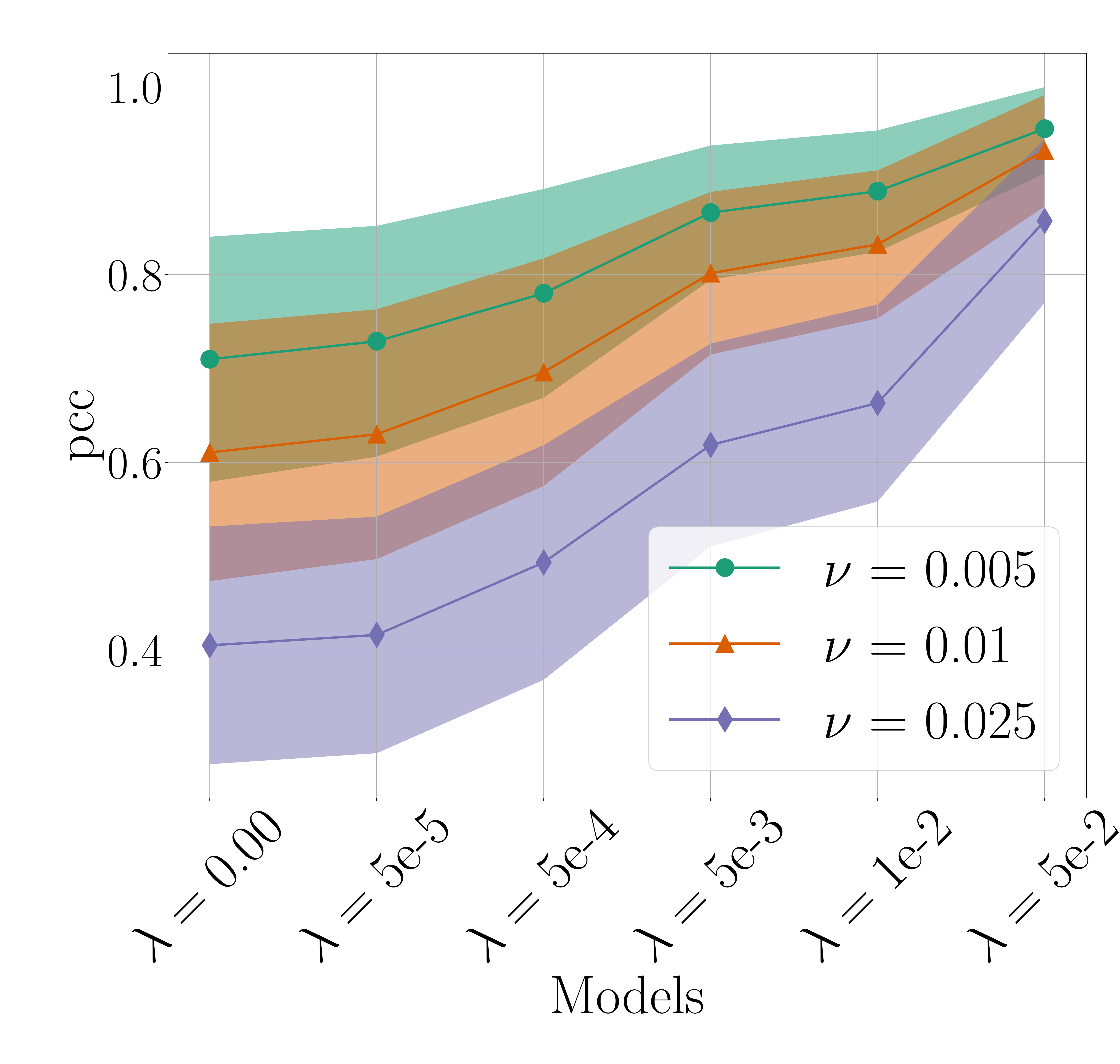}
\endminipage\hfill
\minipage{0.32\textwidth}
  \includegraphics[width=1\linewidth]{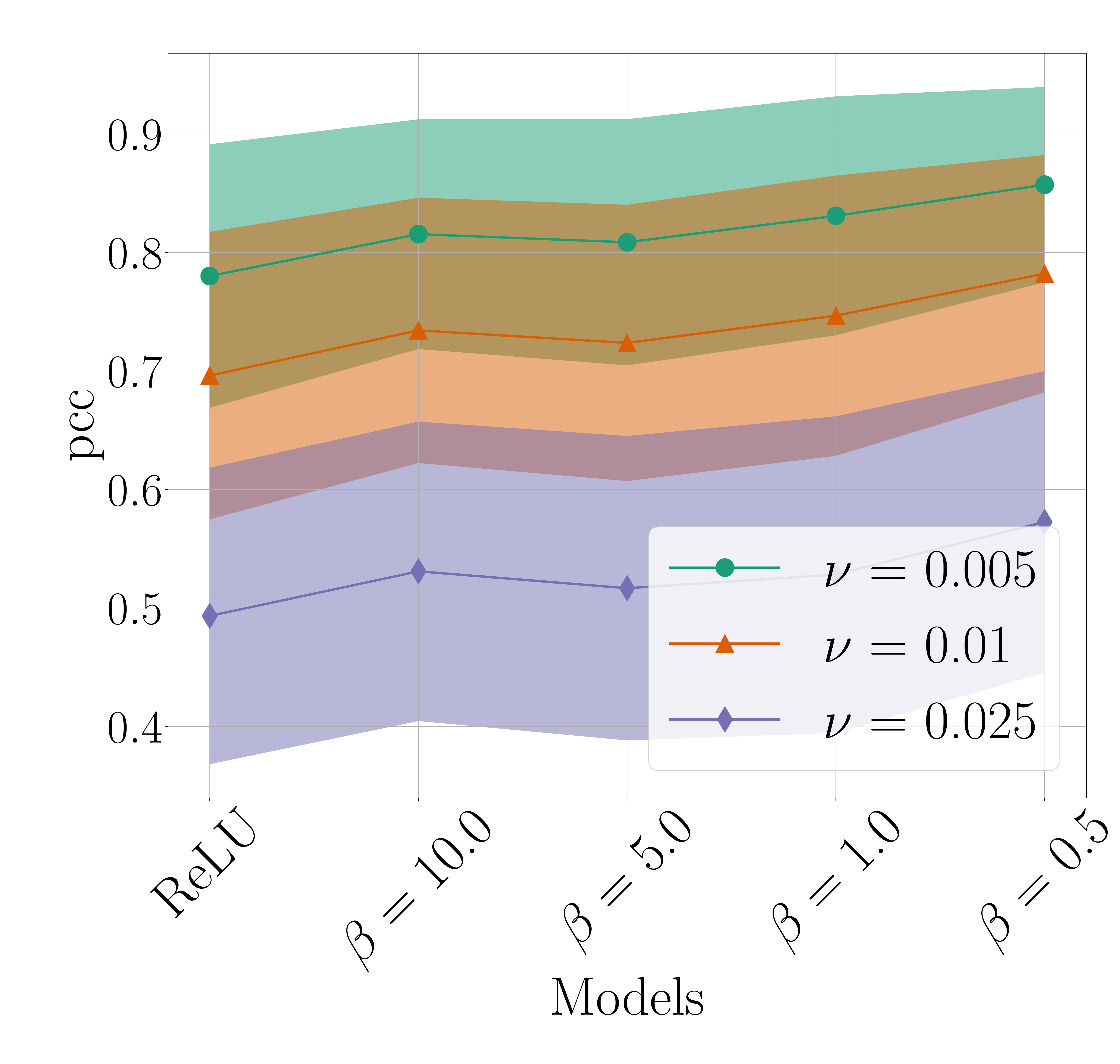}
\endminipage\hfill
\minipage{0.32\textwidth}%
  \includegraphics[width=1\linewidth]{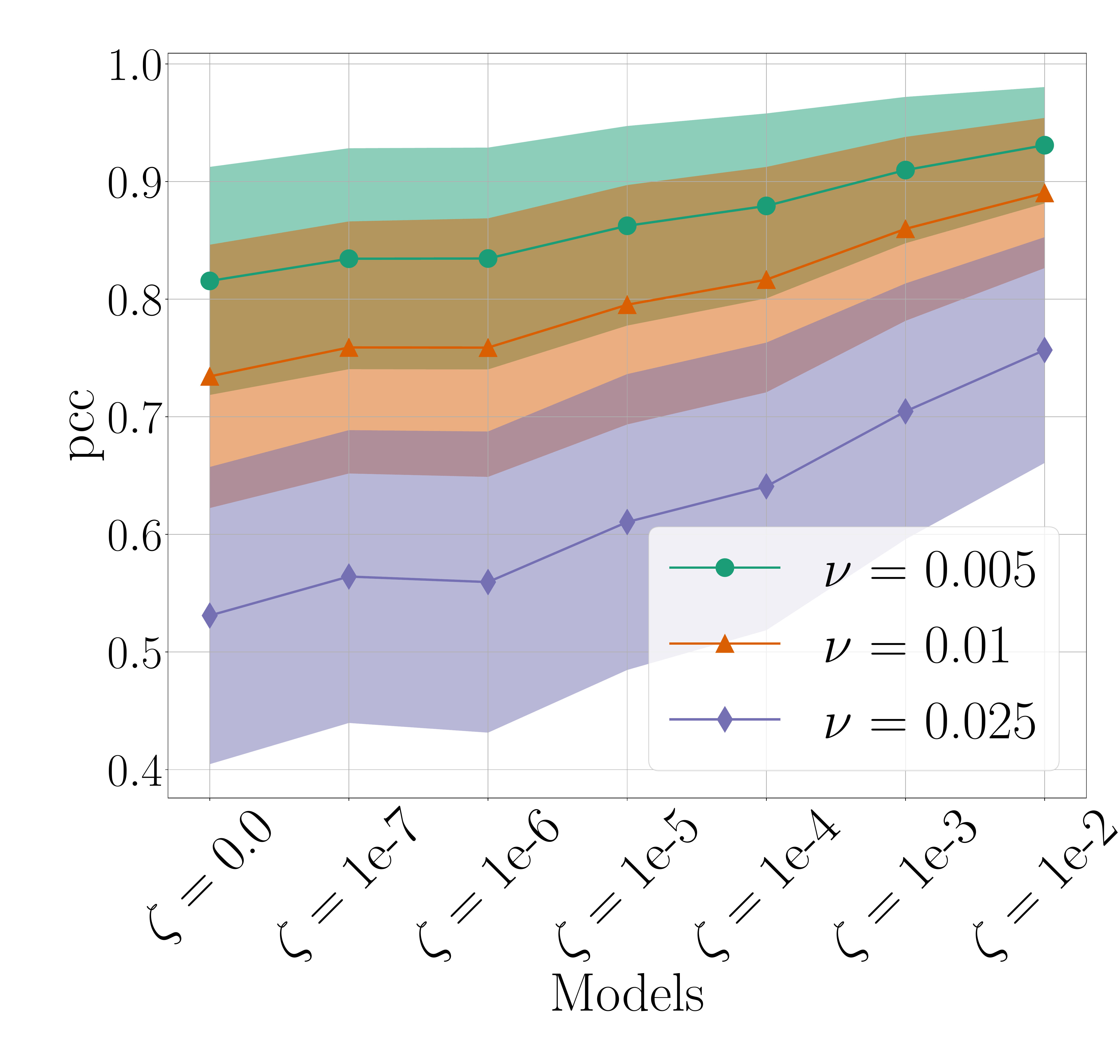}
\endminipage
  \caption{PCCs (mean +/- std) between original Gradient explanation map and explanation after adding salt-pepper noise to the image. left: effect of weight decay, middle: effect of Softplus $\beta$ (for $\lambda=5$e-4), right: effect of curvature minimization (for $\lambda=5$e-4, $\beta$=10).}
  \label{appndx:fig:wd_sp_curv_min_gradient_salt_pepper}
\end{figure}

Figure~\ref{appndx:fig:wd_sp_curv_min_gradient_salt_pepper} shows effects on the Gradient explanation when adding salt-pepper noise to the input images. We can still see a significant improvement in robustness when training networks with our proposed methods, however the effect for Softplus activations and Hessian minimization is less pronounced than for Laplace or Gaussian noise.

\subsection{Connection between Hessian norms and weight norms}
\begin{figure}[htp!]
  \centering
  \includegraphics[width=.9\linewidth]{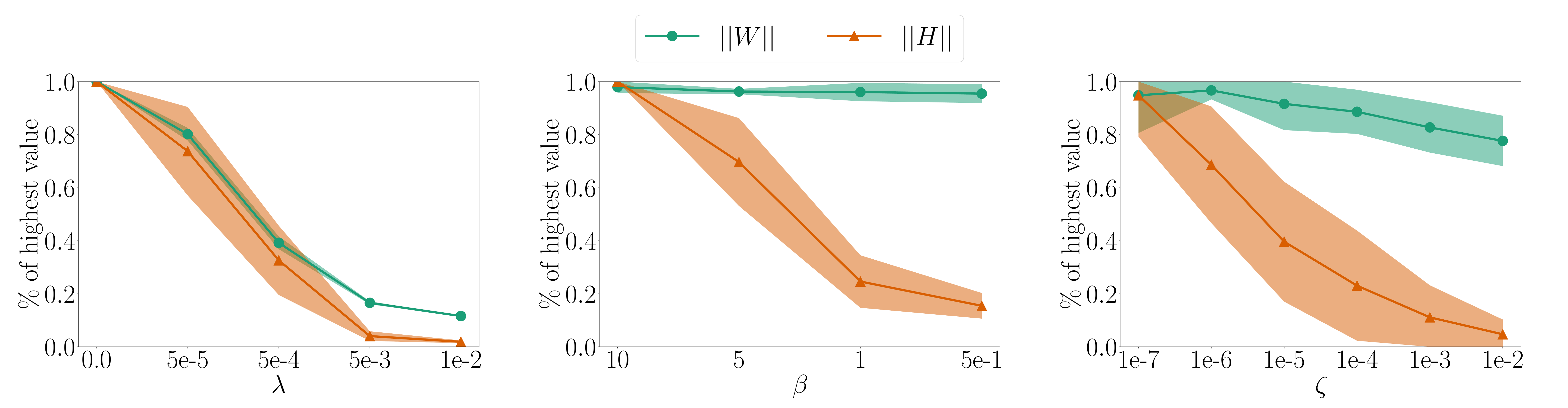}
  \caption{The connection between weight norms $\norm{W}$ and Hessian norms $\norm{H}$ is different for our three methods. Left: $\norm{W}$ and $\norm{H}$ both decrease in a similar manner when applying stronger weight decay (increasing $\lambda$). Middle: $\norm{W}$ stays relatively constant while $\norm{H}$ decreases with increasing Softplus $\beta$. Right: $\norm{W}$ decrease slightly while $\norm{H}$ decrease strongly with stronger Hessian minimization (increasing $\zeta$). We show mean$\pm$std.}
  \label{appndx:fig:W_H_norms}
\end{figure}

In the main text, Section~\ref{sec:comparison}, we mentioned that weight decay leads to decreased weight norms and thus also to decreased Hessian norms. However when we minimize the Hessian norm directly, the weight norms only change minimally.
Figure~\ref{appndx:fig:W_H_norms} shows how weight norms and Hessian norms change when applying weight decay (varying $\lambda$), substituting ReLU with Softplus (varying $\beta$) and minimizing the Hessian norm directly (varying $\zeta$). We average over all Softplus networks trained with $\zeta=0$ for the first two plots and we average over all networks trained with Hessian minimization for the last plot.

\section{Hessian norm approximation}
\begin{figure}[htp!]
  \centering
  \includegraphics[width=.5\linewidth]{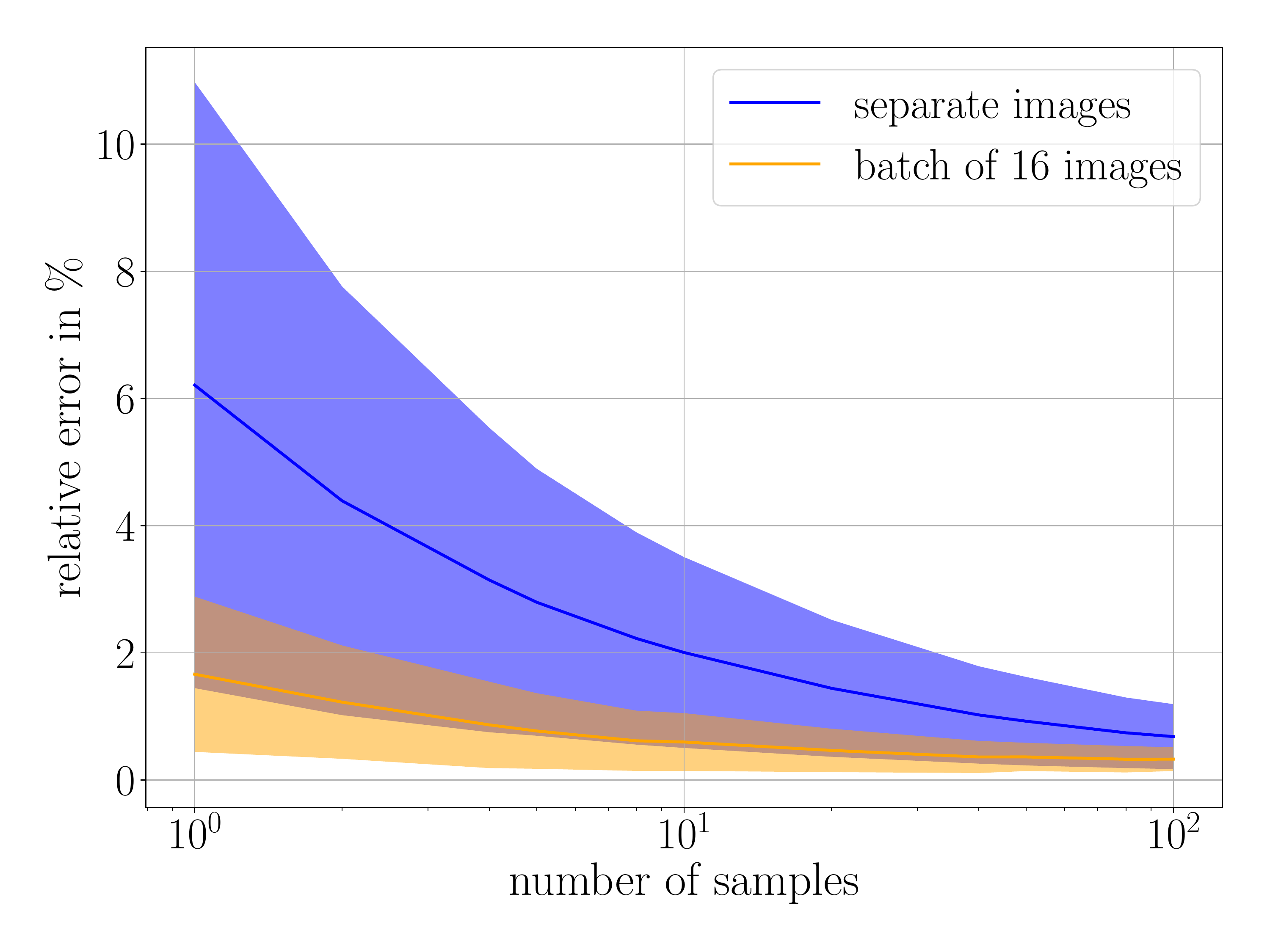}
  \caption{Relative error (mean and standard deviation) between True Hessian norm and approximation via sampling}
  \label{fig:h_sampling_vs_gt}
\end{figure}

In Section~\ref{sec:CurvatureMinimization}, we showed that for the number of samples $N\rightarrow\infty$ the sampling approximation approaches the true Hessian norm. To include the Hessian approximation in our training procedure, we need to fix a certain number of samples and to perform a Monte-Carlo estimate of the expectation value.
Figure~\ref{fig:h_sampling_vs_gt} shows the relative error between sampled hessian norm and true hessian norm. Increasing the sample size noticeably reduces the error, but a sample size of 1 already has a relative error of only 6\%. If we average over a batch of images, the error reduces further. This means that for training and validation sampling once per image is in practice sufficient.

\end{document}